\documentclass[letterpaper]{article}
\usepackage[margin=1in]{geometry}
\usepackage{times}
\usepackage{helvet}
\usepackage{courier}
\usepackage[hyphens]{url}
\usepackage{graphicx}
\usepackage{caption}

\usepackage[sorting=none]{biblatex}
\usepackage{algorithm}
\usepackage{algpseudocode}

\usepackage{hyperref}
\hypersetup{
    colorlinks=true, 
    linktoc=all,     
    linkcolor=blue,  
}

\usepackage{newfloat}
\usepackage{listings}
\DeclareCaptionStyle{ruled}{labelfont=normalfont,labelsep=colon,strut=off}
\floatstyle{ruled}
\newfloat{listing}{tb}{lst}{}
\floatname{listing}{Listing}

\usepackage{url}            
\usepackage{booktabs}       
\usepackage{amsfonts}       
\usepackage{nicefrac}       
\usepackage{microtype}      
\usepackage{xcolor}         
\usepackage{svg}

\usepackage{amsthm}

\usepackage{amsmath}
\usepackage{amssymb}
\usepackage{amsfonts}
\usepackage{mathtools}
\usepackage{svg}
\usepackage{blkarray}
\usepackage{tcolorbox}
\usepackage{cleveref}
\usepackage{arydshln}
\usepackage{subcaption}
\usepackage{cancel}
\usepackage{multirow}

\def\f0{{\mathbf 0}}

\title{Learning Generalized Hamiltonian Dynamics with Stability from Noisy Trajectory Data}

\date{September 6, 2025}

\usepackage{authblk}

\author[1]{Luke McLennan}
\author[1]{Yi Wang}
\author[2]{Ryan Farell}
\author[3]{Minh Nguyen}
\author[1,4]{Chandrajit Bajaj}
\affil[1]{Oden Institute for Computational Engineering and Sciences, The University of Texas at Austin}
\affil[2]{Operations Research and Industrial Engineering, The University of Texas at Austin}
\affil[3]{Department of Mathematics, The University of Texas at Austin}
\affil[4]{Department of Computer Science, The University of Texas at Austin}

\begin{document}

\maketitle

\begin{abstract}
We introduce a robust framework for learning various generalized Hamiltonian dynamics from noisy, sparse phase-space data and in an unsupervised manner based on variational Bayesian inference. Although conservative, dissipative, and port-Hamiltonian systems might share the same initial total energy of a closed system, it is challenging for a single Hamiltonian network model to capture the distinctive and varying motion dynamics and physics of a phase space, from sampled observational phase space trajectories. To address this complicated Hamiltonian manifold learning challenge, we extend sparse symplectic, random Fourier Gaussian processes learning with predictive successive numerical estimations of the Hamiltonian landscape, using generalized form of state and conjugate momentum Hamiltonian dynamics, appropriate to different classes of conservative, dissipative and port-Hamiltonian physical systems. In addition to the kernelized evidence lower bound (ELBO) loss for data fidelity, we additionally incorporate stability and conservation constraints as additional hyper-parameter balanced loss terms to regularize the model’s multi-gradients, enforcing physics correctness for improved prediction accuracy with bounded uncertainty.
\end{abstract}

\tableofcontents

\section{Introduction}
Learning dynamical systems from data is a fundamental problem in physics-informed machine learning, with widespread applications in physics, engineering, and robotics \cite{heinonen2018odegp} \cite{chen2018neuralode} \cite{ayed2019partialobservations} \cite{manek2019stabledeepdynamics}. The aim in these works and our work is to learn operators which map from initial condition to trajectories. This has been considered for deterministic systems, and also stochastic systems \cite{liu2019nsde} \cite{kidger2021nsdegan}. We focus on various forms of generalized Hamiltonian dynamics, conservative \cite{greydanus2019hnn} \cite{rath21sympgpr} \cite{ross2023hgp}, dissipative \cite{sosanya2022dhnn} \cite{zhong2020dissipative} \cite{tanaka2022ssgp}, and port-Hamiltonian \cite{schaft2014portham} \cite{desai2021port}. The different classes of dynamics present different phase spaces which must be learned. While there can be complex phase spaces even for conservative systems, adding dissipation and/or external forcing introduces additional complexity, such as the overlapping of trajectories. We present slightly different methodologies for the different classes of dynamics, but with a commonality of learning the phase space by stochastic process-based training with enforcement of stability.

Simple sparse variational regression techniques in Euclidean space often fail to accurately capture the dynamics with low uncertainty  in prediction as a result of the lack of utilizing the Hamiltonian manifold and missing the underlying physical constitutive and conservation laws in space. The method of choice is to learn dynamics  through optimization on Hamiltonian manifolds, (in this case Riemannian)  and in a principled way that achieves stability throughout the dynamics learning process. In this work, we utilize the differential properties of the Hamiltonian manifold, its symplectic form, while preserving the conservation laws and stability properties as implicit soft regularizers, allowing us to learn dynamics in a manner consistent with inherent physical principles, thereby going beyond what standard Euclidean regression can achieve.

We focus on differentially learning stochastic Hamiltonian systems, which describe the stochastic evolution of states in phase space, comprising generalized positions \(\mathbf{q}(t) \in \mathbb{R}^d\) and generalized conjugate momenta \(\mathbf{p}(t) \in \mathbb{R}^d\) while exploiting both the symplectic energy conservation form and the dissipative forms, caused by motion in dissipative environments. These dynamics are governed by a generalized Hamiltonian function \(\mathcal{H}(\mathbf{x}(t))\), where \(\mathbf{x}(t) = [\mathbf{q}(t); \mathbf{p}(t)] \in \mathbb{R}^{2d}\) represents the state of the system. The principled learning of stochastic Hamiltonians are estimated by path-wise spectrally kernelized (random Fourier feature) stochastic Gaussian processes. We show the advantages of differential dynamical modeling with generalized Hamiltonian systems and with Hamiltonian gradients. 

Our contributions are the following:
\begin{itemize}
    \item We apply and extend a probabilistic framework that handles noisy observations and provides distributions over possible trajectories, enhancing the method's applicability to real-world data that are often noisy and incomplete. 
    \item We extend the optimization problem by incorporating Lyapunov stability and conservation law constraints as soft loss terms, enabling the learned dynamics to adhere closely to physical principles. 
    \item We use improved hyperparameter balancing during training, optimizing the trade-off between model accuracy and stability.
    \item We show experiments on standard and generalized Hamiltonian systems, which demonstrates the improved performance of our method compared to prior work\footnote{Our code is available at \url{https://github.com/CVC-Lab/HamiltonianLearning}.}.
\end{itemize}

\section{Background}
\subsection{Intro to Generalized Hamiltonian Dynamics}
Hamiltonian dynamics describe the dynamics of position $\mathbf{q}$ and conjugate momentum $\mathbf{p}$ using a Hamiltonian $\mathcal H(\mathbf{q},\mathbf{p})$, which defines the energy landscape of the system. In standard conservative Hamiltonian dynamics, the $(\mathbf{q},\mathbf{p})$ dynamics move along constant energy paths defined by the Hamiltonian. There are several generalizations of Hamiltonian dynamics which incorporate terms such as dissipation and external forcing \cite{schaft2014portham}.
We consider the following forms of Hamiltonian dynamics and generalizations:
\begin{itemize}
    \item Conservative Hamiltonian dynamics: 
    \begin{equation}
        (\dot{\mathbf{q}}, \dot{\mathbf{p}}) = J \nabla \mathcal H(\mathbf{q},\mathbf{p}).
        \label{eq:hd:conservative}
    \end{equation}
    \item Dissipative Hamiltonian dynamics: 
    \begin{equation}
        (\dot{\mathbf{q}}, \dot{\mathbf{p}}) = (J+D) \nabla \mathcal H(\mathbf{q},\mathbf{p}).
        \label{eq:hd:dissipative}
    \end{equation}
    \item Port-Hamiltonian dynamics: 
    \begin{equation}
        (\dot{\mathbf{q}}, \dot{\mathbf{p}}) = (J+D) \nabla \mathcal H(\mathbf{q},\mathbf{p}) + F(t).
        \label{eq:hd:port}
    \end{equation}
\end{itemize}
We assume that we are given a dataset of trajectories and in addition the class of Hamiltonian dynamics is specified, as described in the next subsection. Based on the class of dynamics, the structure of the model and the training procedure are slightly different, as described in Section \ref{sec:methods}.

\begin{figure}[t]
    \centering
    \includegraphics[width=0.8\linewidth]{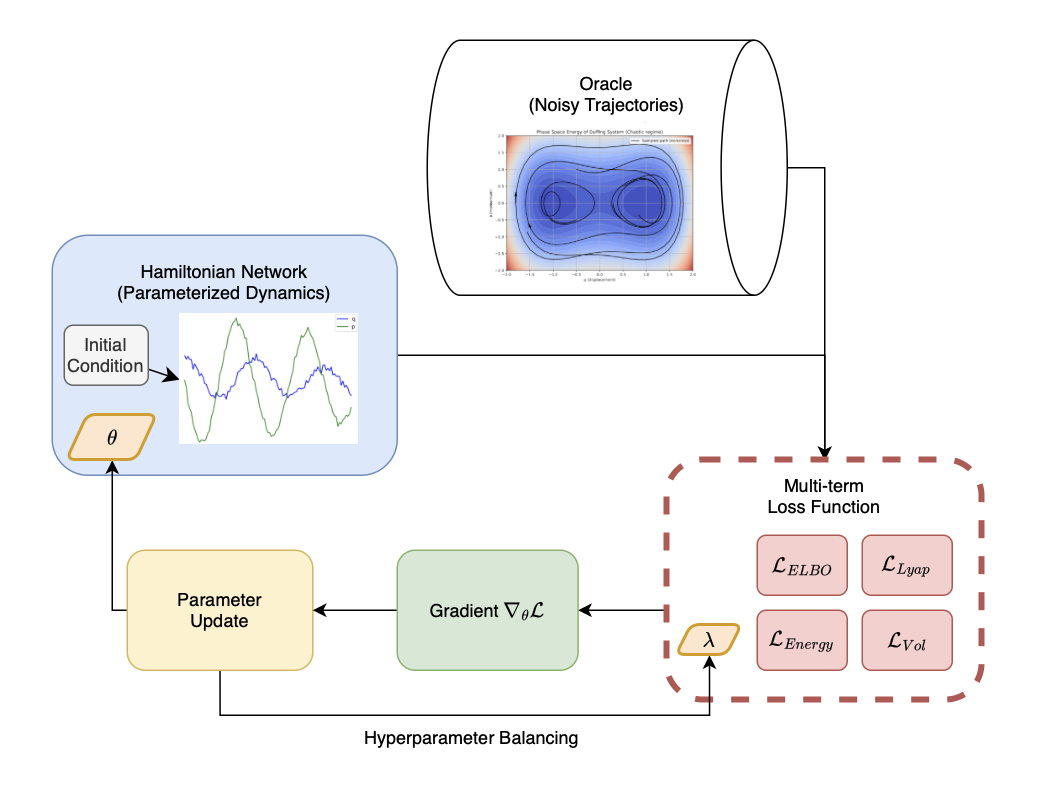}
    \caption{A visualization of our framework for generalized Hamiltonian learning. The Hamiltonian network queries the oracle for data, which is used to evaluate the quality of the model through the multi-term loss function. The gradients of the loss are update the Hamiltonian dynamics parameters and the process is repeated.}
    \label{fig:framework}
\end{figure}

\begin{figure}[t]
    \captionsetup[subfigure]{labelformat=empty}
    \centering
    \begin{subfigure}[t]{0.25\linewidth}
        \includegraphics[width=\linewidth]{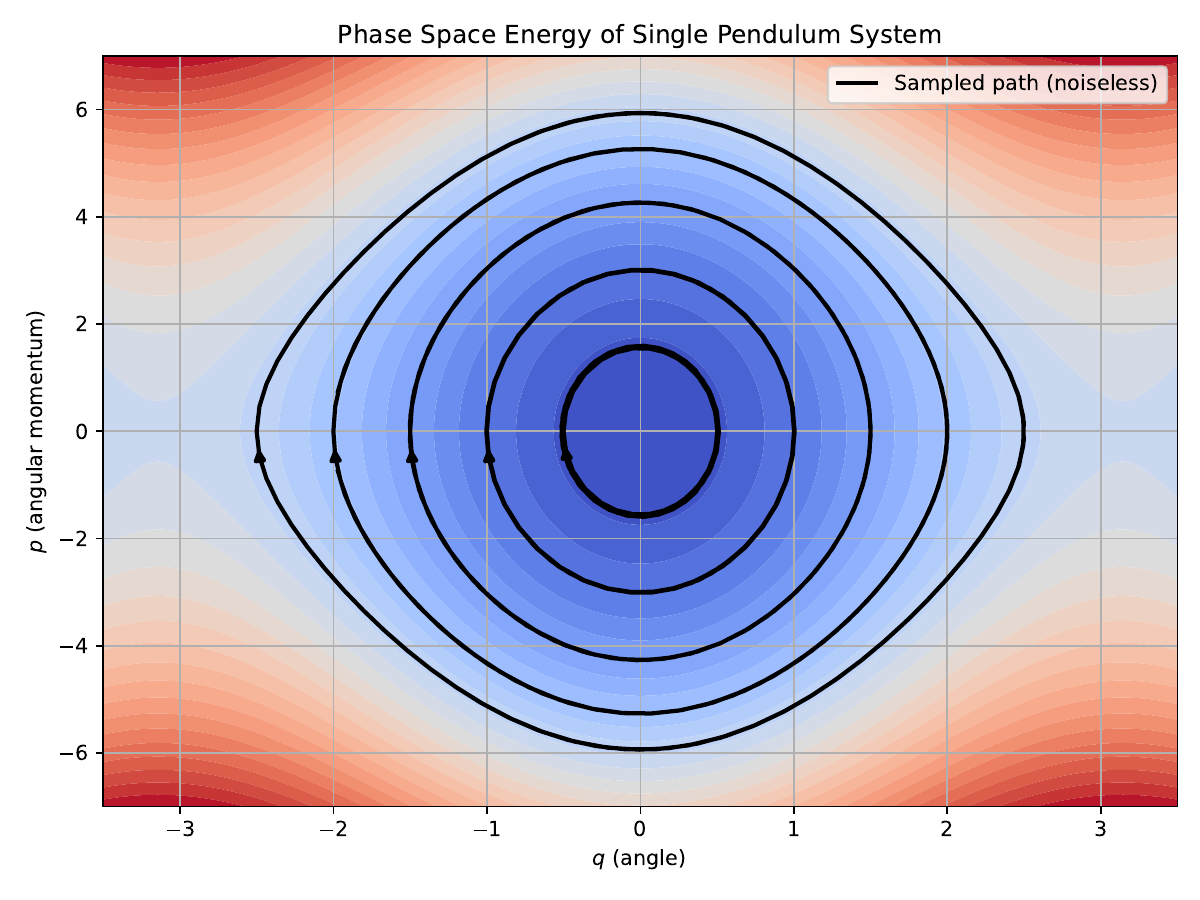}
        \caption{Single Pendulum}
    \end{subfigure}
    \begin{subfigure}[t]{0.25\linewidth}
        \includegraphics[width=\linewidth]{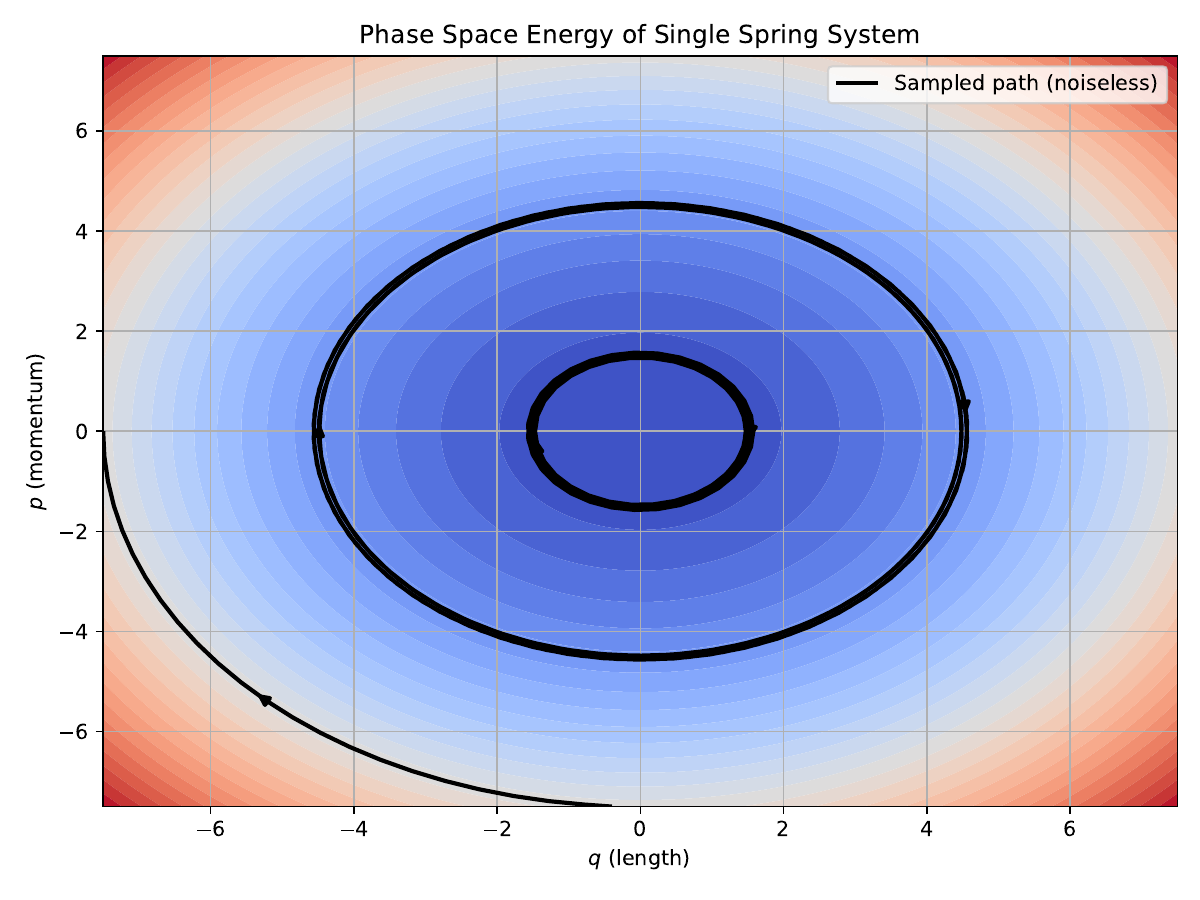}
        \caption{Conservative Spring}
    \end{subfigure}
    \begin{subfigure}[t]{0.25\linewidth}
        \includegraphics[width=\linewidth]{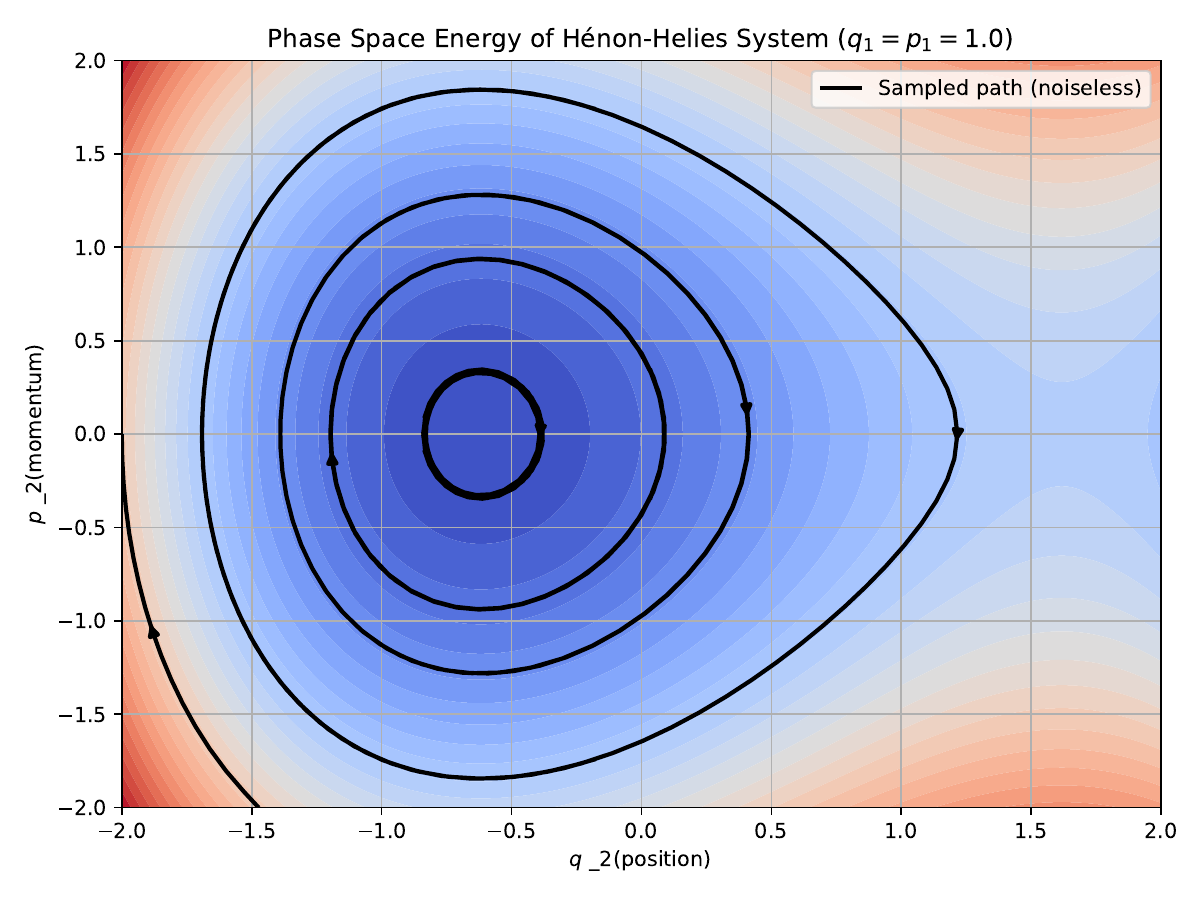}
        \caption{Henon-Heiles}
    \end{subfigure}
    \begin{subfigure}[t]{0.25\linewidth}
        \includegraphics[width=\linewidth]{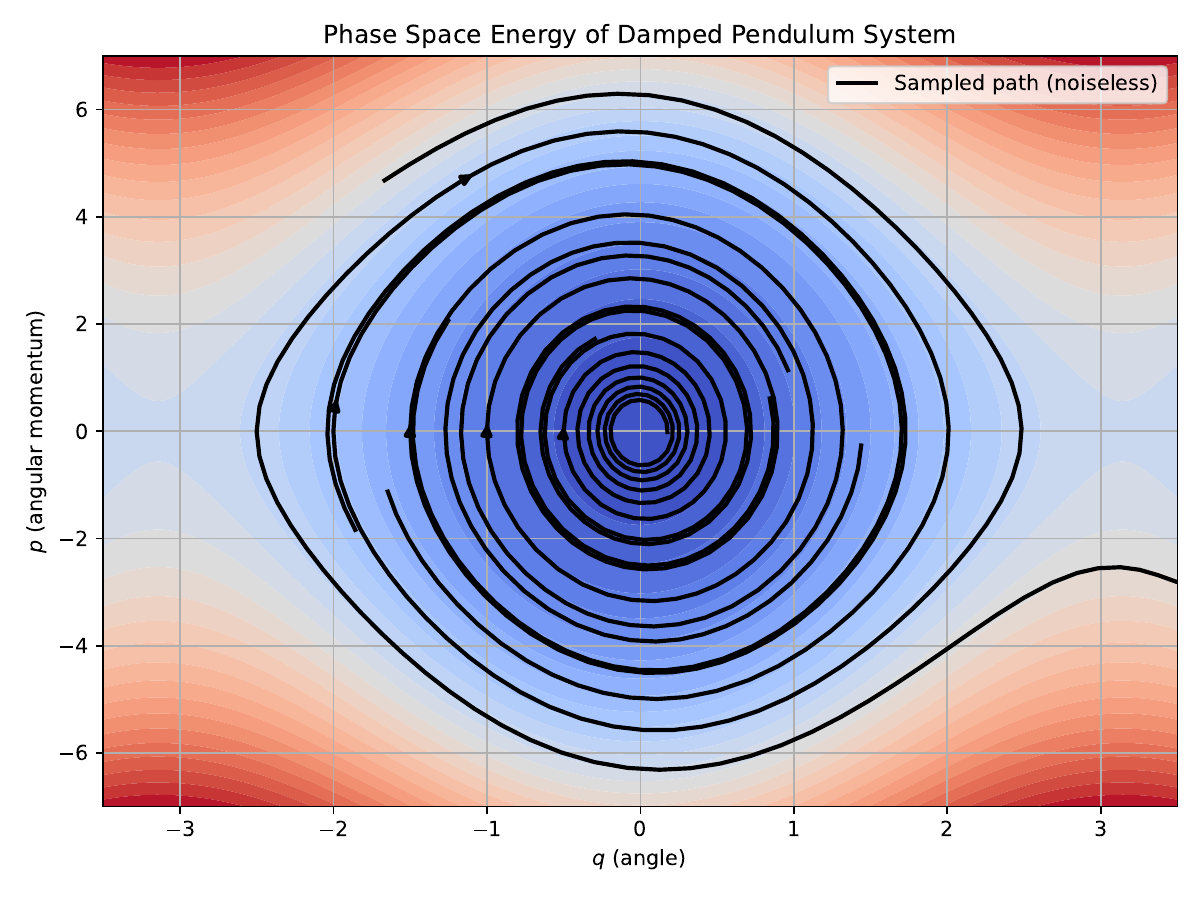}
        \caption{Damped Pendulum}
    \end{subfigure}
    \begin{subfigure}[t]{0.25\linewidth}
        \includegraphics[width=\linewidth]{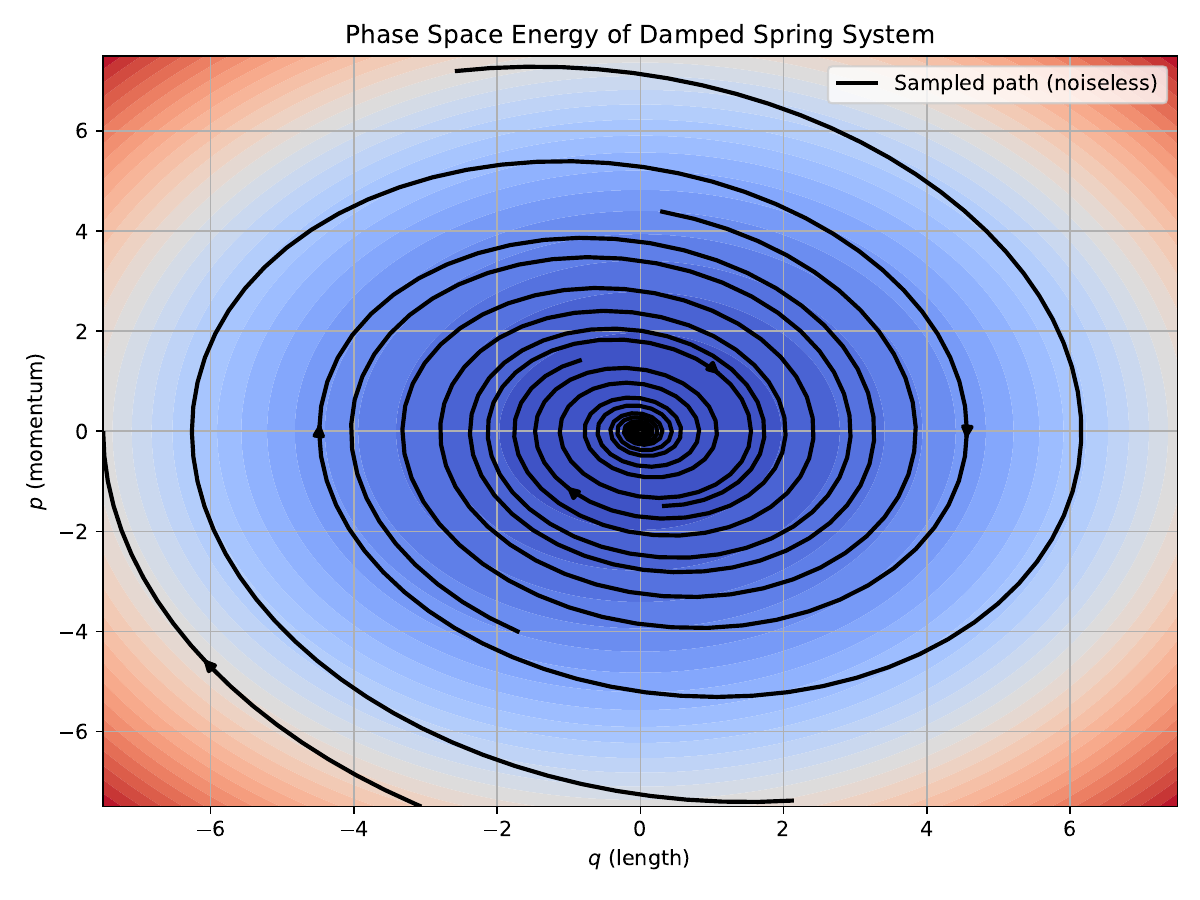}
        \caption{Damped Spring}
    \end{subfigure}
    \begin{subfigure}[t]{0.25\linewidth}
        \includegraphics[width=\linewidth]{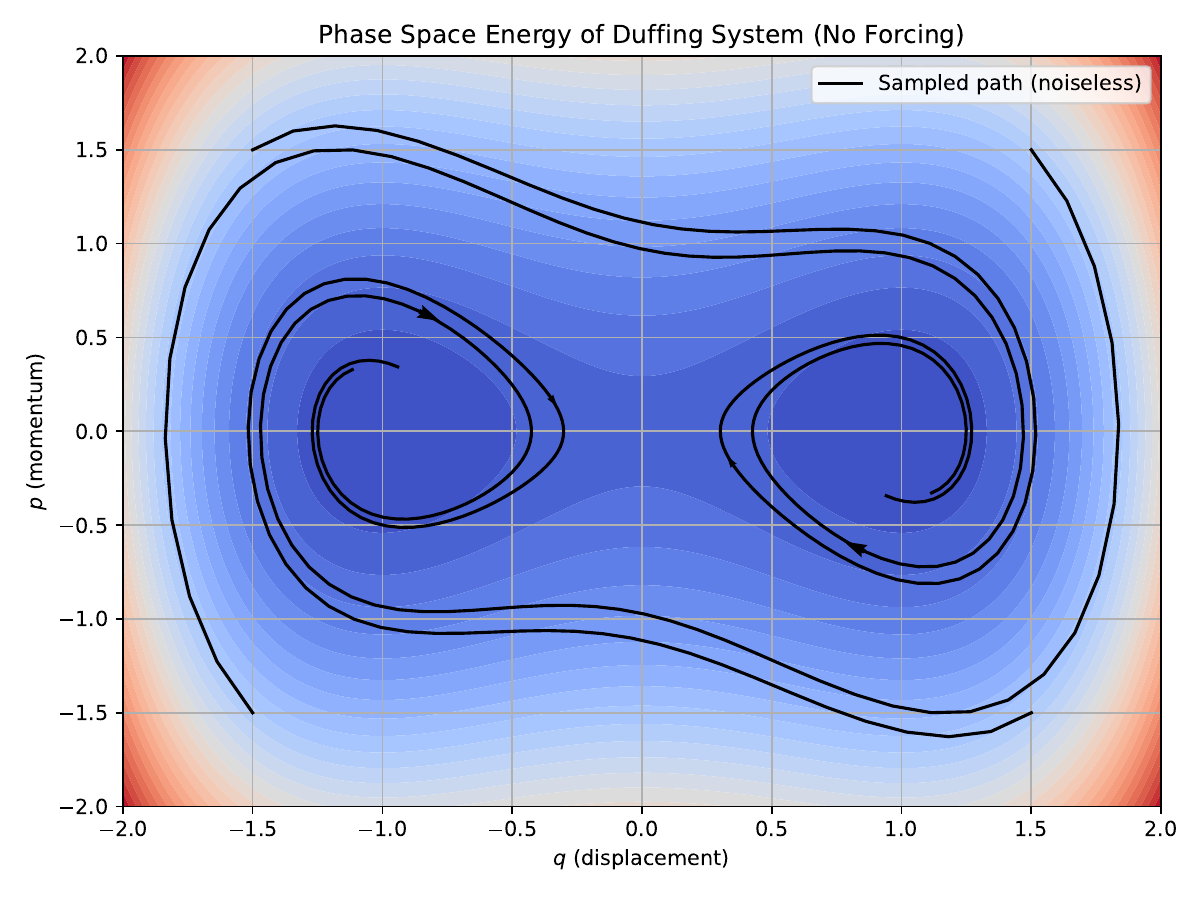}
        \caption{Unforced Duffing}
    \end{subfigure}
    \begin{subfigure}[t]{0.25\linewidth}
        \includegraphics[width=\linewidth]{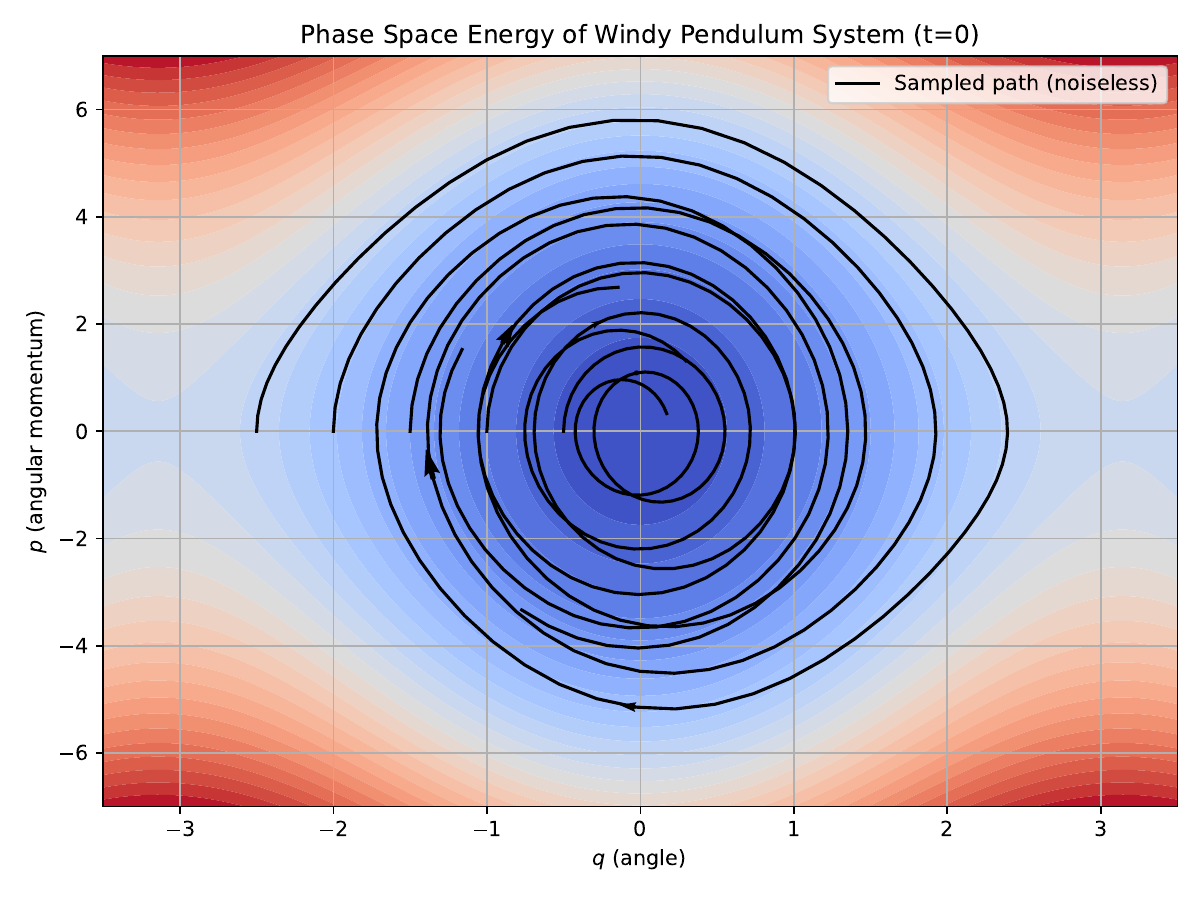}
        \caption{Windy Pendulum}
    \end{subfigure}
    \begin{subfigure}[t]{0.25\linewidth}
        \includegraphics[width=\linewidth]{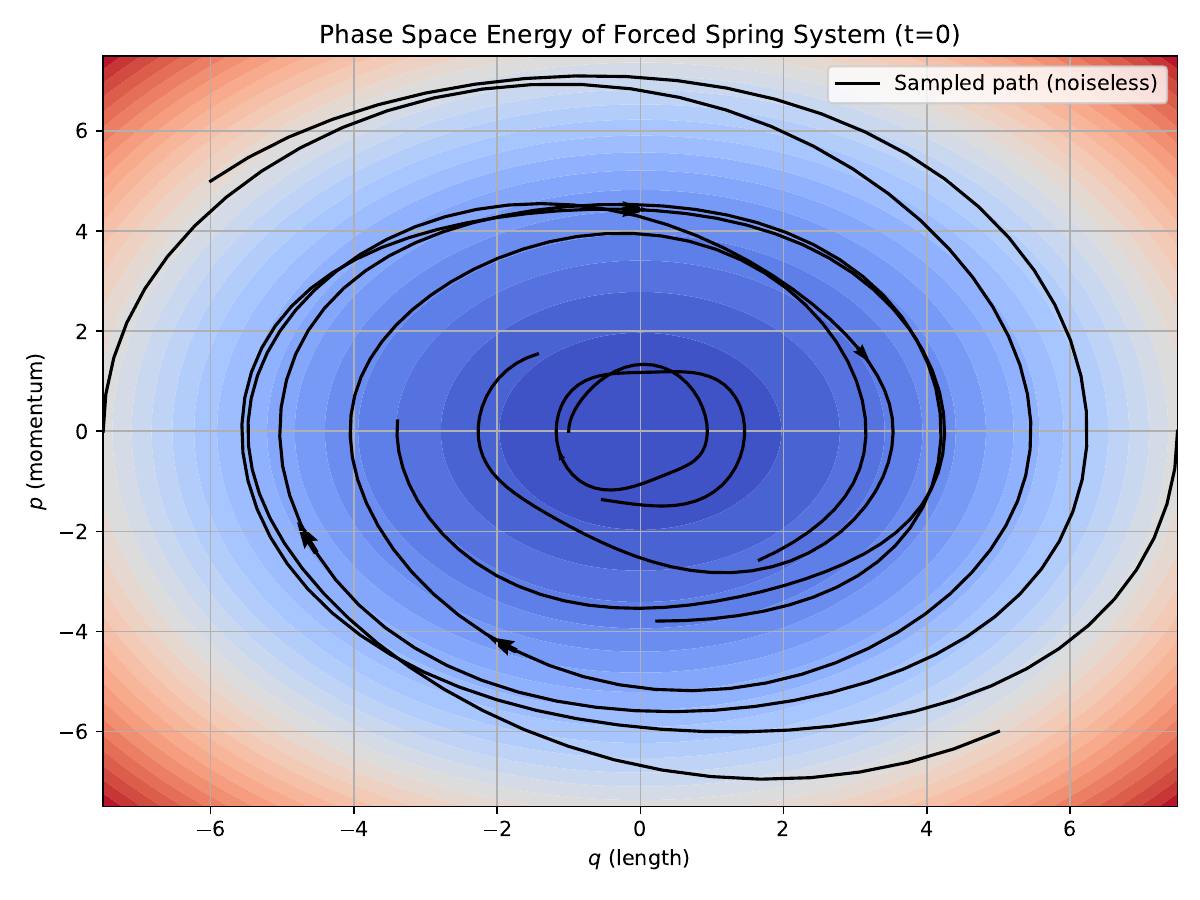}
        \caption{Forced Spring}
    \end{subfigure}
    \begin{subfigure}[t]{0.25\linewidth}
        \includegraphics[width=\linewidth]{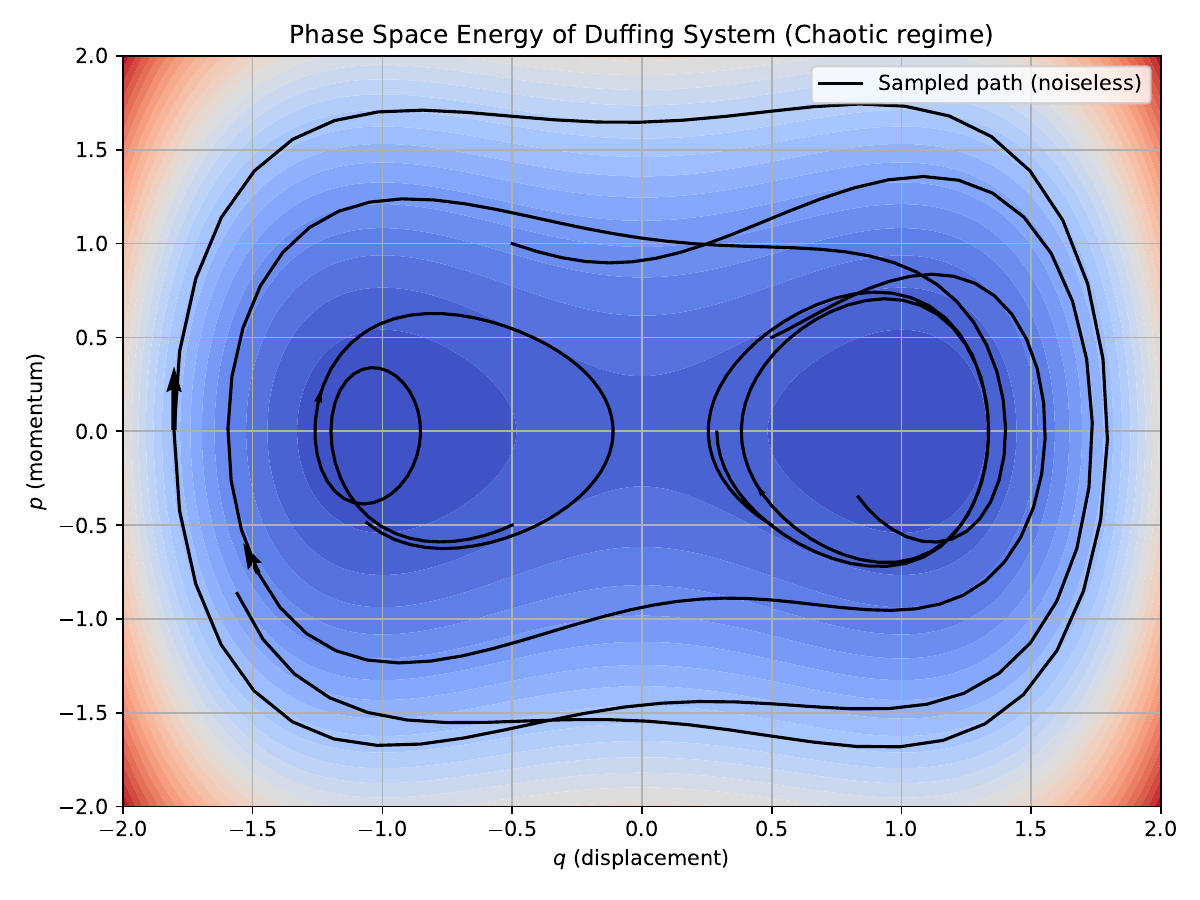}
        \caption{Forced Duffing}
    \end{subfigure}
    \caption{Phase map diagrams with sampled trajectories for various conservative (top row), dissipative (middle row), and port-Hamiltonian (bottom row) systems. In these examples, the heat map denotes the Hamiltonian energy, and the dynamics of systems are plotted for various initial conditions. Column 1 shows pendulum systems, column 2 shows spring systems, and column 3 shows the Henon-Heiles system for the conservative case and Duffing systems for the dissipative and port cases.}
    \label{fig:phase-maps}
\end{figure}

Hamiltonian systems have a number of associated symmetries and conservation laws, including conservation of energy and conservation of volume \cite{meiss2007hamsys}. However, these may only apply in the case of conservative dynamics. For dissipative dynamics, the energy and volume are decreasing in time, and for port-Hamiltonian systems, they are dependent on the external forcing. Thus, different training procedures are needed for the different classes of dynamics. 

\subsection{Example Hamiltonian Systems}
\label{sec:background:examples}
We list out $9$ Hamiltonian systems in Table \ref{tab:hamiltonian:example}. They are grouped based on the three classes described in the previous section. The phase space dynamics of these systems are visualized in Figure \ref{fig:phase-maps}.

\label{sec:appendix:hamiltonian:systems}
\begin{table}[ht]
    \centering
    \begin{tabular}{c|c|c|c}
        \hline
        \textbf{System Name} & \textbf{Hamiltonian} & \textbf{Euler-Lagrangian Equation} & \textbf{(q,p)-Dynamics (type)} \\
        \hline
        Single Pendulum & P &  $l\ddot{q} + g\sin q=0$ &  $J\nabla \mathcal{H}$ \\
        Simple Spring & S & $m\ddot{q} +kq = 0$  & $J\nabla \mathcal{H}$ \\
        Henon-Heiles & HH & \(
    \begin{cases}
        \ddot{q}_1 + q_1 +2q_1q_2=0 \\
        \ddot{q}_2 + q_2 - (q_1^2-q_2^2)=0
    \end{cases}
        \) &  $J\nabla \mathcal{H}$ \\
        \hdashline
        Damped Pendulum & DP &  $l\ddot{q} + \gamma\dot{q} + g \sin q=0$ &  $\left( J + D \right) \nabla \mathcal{H}$ \\
        Damped Spring & DS & $m\ddot{q} +\gamma\dot{q}+kq = 0$  & $(J + D)\nabla \mathcal{H}$ \\
        Unforced Duffing Equation & UD & $\ddot{q}+\gamma \dot{q} + \alpha q +\beta q^3 = 0$ & $\left( J + D \right) \nabla \mathcal{H}$ \\
        \hdashline
        Windy Pendulum & WP & $l\ddot{q} + \gamma\dot{q} + g \sin q= F(t)$ &  $\left( J + D \right) \nabla \mathcal{H} + F(t)$ \\
        Forced Spring & FS & $m\ddot{q} +\gamma\dot{q}+kq = F(t)$ & $(J + D)\nabla \mathcal{H} + F(t)$ \\
        Duffing Equation & DE & $\ddot{q}+\gamma \dot{q} + \alpha q +\beta q^3 = F(t)$ & $\left( J + D \right) \nabla \mathcal{H} + F(t)$ \\
        \hline
    \end{tabular}
    \caption{Different (generalized) Hamiltonian systems with their corresponding Euler-Lagrangian equations are grouped into conservative, dissipative, and port-Hamiltonian classes, each having a form of $(q,p)$ dynamics.}
    \label{tab:hamiltonian:example}
\end{table}

The Hamiltonian energy functions are given below. Note that for the different types of pendulum, spring, and Duffing systems, they share same physical model, and therefore their Hamiltonians are identical, but the dynamics are different depending on dissipation/forcing.
\begin{align}
    \mathcal{H}(q,p) &= \frac{p^2}{2ml^2} - mgl \cos q \tag{P/DP/WP}\\
    \mathcal{H}(q,p) &= \frac{1}{2}kq^2 + \frac{p^2}{2m}\tag{S/DS/FS} \\
 \mathcal{H}(q_1,q_2,p_1,p_2) &= \frac{1}{2} (p_1^2 + p_2^2) + \frac{1}{2} (q_1^2 + q_2^2) + q_1^2 q_2 - \frac{1}{3} q_2^3 \tag{HH} \\
\mathcal{H}(q,p) &= \frac{p^2}{2m} + \frac{\alpha q^2}{2} + \frac{\beta q^4}{4} 
\tag{UD/DE}
\end{align}

The Lagrangian of each problem is:
\begin{align}
    \mathcal{L}(q, \dot{q}) &= \frac{1}{2} ml^2 \dot{q}^2 + mgl \cos q \tag{P/DP/WP} \\  
    \mathcal{L}(q, \dot{q}) &= \frac{1}{2} m \dot{q}^2 - \frac{1}{2} k q^2 \tag{S/DS/FS} \\  
    \mathcal{L}(q_1, q_2, \dot{q}_1, \dot{q}_2) &= \frac{1}{2} (\dot{q}_1^2 + \dot{q}_2^2) - \frac{1}{2} (q_1^2 + q_2^2) - q_1^2 q_2 + \frac{1}{3} q_2^3 \tag{HH} \\  
    \mathcal{L}(q, \dot{q}) &= \frac{1}{2} m \dot{q}^2 - \frac{\alpha q^2}{2} - \frac{\beta q^4}{4} \tag{UD/DE}  
\end{align}

The corresponding phase space dynamics are given accordingly, with time-dependent forces being defined:
\begin{align}
 \begin{bmatrix}
    \dot{q} \\ \dot{p}    
    \end{bmatrix}  &= \begin{bmatrix}
        0 &  1  \\
        -1 & 0
    \end{bmatrix} \begin{bmatrix}
        mgl \sin q \\
        p/ml^2
    \end{bmatrix}
    \tag{dqdp-P} \\
    \begin{bmatrix}
    \dot{q} \\ \dot{p}    
    \end{bmatrix} &= \begin{bmatrix}
        0 &  1  \\
        -1 & -\gamma
    \end{bmatrix} \begin{bmatrix}
        mgl \sin q \\
        p/ml^2
    \end{bmatrix}
    \tag{dqdp-DP} \\
    \begin{bmatrix}
    \dot{q} \\ \dot{p}    
    \end{bmatrix} &= \begin{bmatrix}
        0 &  1  \\
        -1 & -\gamma
    \end{bmatrix} \begin{bmatrix}
        mgl \sin q \\
        p/ml^2
    \end{bmatrix} + 
    \begin{bmatrix}
        0 \\  vt
    \end{bmatrix}
    \tag{dqdp-WP} \\
    \begin{bmatrix}
    \dot{q} \\ \dot{p}    
    \end{bmatrix} &= \begin{bmatrix}
        0 &  1  \\
        -1 & 0
    \end{bmatrix} \begin{bmatrix}
        kq \\
        p/m
    \end{bmatrix}
    \tag{dqdp-S}\\
    \begin{bmatrix}
    \dot{q} \\ \dot{p}    
    \end{bmatrix} &= \begin{bmatrix}
        0 &  1  \\
        -1 & -\gamma
    \end{bmatrix} \begin{bmatrix}
        kq \\
        p/m
    \end{bmatrix}
    \tag{dqdp-DS} \\
    \begin{bmatrix}
    \dot{q} \\ \dot{p}    
    \end{bmatrix} &= \begin{bmatrix}
        0 &  1  \\
        -1 & -\gamma
    \end{bmatrix} \begin{bmatrix}
        kq  \\
        p/m 
    \end{bmatrix} + \begin{bmatrix}
        0 \\ F_0\sin(\omega t)\sin(2\omega t)
    \end{bmatrix}
    \tag{dqdp-FS} \\
    \begin{bmatrix}
    \dot{q}_1  \\ \dot{q}_2 \\ \dot{p}_1 \\ \dot{p}_2    
    \end{bmatrix} &= \begin{bmatrix}
        0 &  I  \\
        -I & 0
    \end{bmatrix} \begin{bmatrix}
        q_1 + 2 q_1 q_2 \\
        q_2 + q_1^2 - q_2^2 \\
        p_1 \\
        p_2
    \end{bmatrix}
    \tag{dqdp-HH} \\
    \begin{bmatrix}
    \dot{q} \\ \dot{p}    
    \end{bmatrix} &= \begin{bmatrix}
        0 &  1  \\
        -1 & -\gamma
    \end{bmatrix} \begin{bmatrix}
        \alpha q + \beta q^3 \\
        p/m
    \end{bmatrix}
    \tag{dqdp-UD} \\
    \begin{bmatrix}
    \dot{q} \\ \dot{p}    
    \end{bmatrix} &= \begin{bmatrix}
        0 &  1  \\
        -1 & -\gamma
    \end{bmatrix} \begin{bmatrix}
        \alpha q + \beta q^3 \\
        p/m
    \end{bmatrix} + \begin{bmatrix}
        0 \\ F_{ext}
    \end{bmatrix}
    \tag{dqdp-DE} 
\end{align}

\subsection{Learning Generalized Hamiltonian Dynamics}
Hamiltonian neural network (HNN) \cite{greydanus2019hnn} is a method to learn Hamiltonian dynamics by parameterizing the Hamiltonian as a neural network and training it with a MSE loss function of the time-derivative data. HNN applies only for conservative systems. The work has been extended to dissipative systems with dissipative HNN (D-HNN) \cite{sosanya2022dhnn}, which learns a second neural network to represent the dissipative component of the dynamics. Port-HNN (P-HNN) \cite{desai2021port} parameterizes neural networks for the Hamiltonian, dissipation matrix, and external forcing function in order to learn dynamics in the Port-Hamiltonian framework \cite{schaft2014portham}. For training any of these models, data of not only the $q,p$ trajectories, but also $\dot q, \dot p$ trajectories is required. These methods also are not probabilistic methods, and thus they can suffer with noisy data sets.

\subsection{Probabilistic Learning from Noisy Data}
Gaussian processes (GPs) have been widely used for modeling of dynamical systems with uncertainty, particularly in the cases of sparse or noisy data \cite{heinonen2018odegp}. Gaussian processes have been used for learning dynamics for general ODEs \cite{bhouri2022gpnode}, and for Hamiltonian systems \cite{ross2023hgp}. 
To address computational limitations, Random Fourier Features (RFF) \cite{rahimi2007random} approximate GPs efficiently, with further improvements such as Sparse Spectrum GPs \cite{lazaro2010sparse} and Spherical Structured Features \cite{lyu2017spherical}. However, these methods often fail to capture the crucial symplectic structure for Hamiltonian systems. Methods such as Symplectic Gaussian Process Regression (SymGPR) \cite{rath21sympgpr}, Symplectic Spectrum Gaussian Processes (SSGP) \cite{tanaka2022ssgp}, and Hamiltonian Gaussian Processes (HGP) \cite{ross2023hgp} address this by incorporating symplecticity and GPs to model noisy dynamics, but lack explicit stability/conservation constraints. SymGPR uses Gaussian process regression, while SSGP and HGP are trained to maximize the evidence lower bound (ELBO) of a data set of trajectories. Gaussian Process Port-Hamiltonian Systems (GPPHS) \cite{Beckers2022gpport} also uses a GP approach and extends the application to learning of Port-Hamiltonian dynamics with dissipation and external forcing.

\section{Methods}
\label{sec:methods}

\subsection{Hamiltonian Surrogate Model}
\label{sec:hnet}
To learn a conservative, dissipative, and Port-Hamiltonian dynamics in a systematic scheme, we propose to decouple energy conservation term $J\nabla\mathcal{H}$, dissipative term $D \nabla \mathcal{H}$, and external forces $F(t)$, when applicable. Our scheme treats each module individually and enforces additional constraints to ensure that each term captures desired physical property from noisy observations. We start by stating our choice of stochastic Hamiltonian surrogate, which is a random Fourier feature (RFF) approximation of a Gaussian process (GP).

Our stochastic Hamiltonian follows a GP prior:
\begin{equation}
\begin{aligned}
    \mathcal{H}(\mathbf{q},\mathbf{p}) &\sim GP(0,K((\mathbf{q},\mathbf{p}),(\mathbf{q}',\mathbf{p}'))), \\
    \nabla \mathcal{H}(\mathbf{q},\mathbf{p}) &\sim GP(0,\tilde{K}), \quad \tilde{K} = \nabla^2K.
\end{aligned}
\end{equation}

The specific GP prior we choose is the spectral kernel with random Fourier features (RFF) \cite{tanaka2022ssgp}. 
\begin{equation}
    \mathcal H(\mathbf{q},\mathbf{p}) = \sum_{m=1}^M w_m^{\top}\phi_m(\mathbf{q},\mathbf{p}), ~~ \phi_m(\mathbf{x}) = \begin{bmatrix}
        \cos(\omega_m^{\top} \mathbf{x}) \\[1pt] \sin(\omega_m^{\top}\mathbf{x})
    \end{bmatrix}.
\end{equation}
The weights $w_m$ are sampled from normal distributions: 
\begin{equation}
\begin{aligned}    
    w_m &\sim \mathcal N(0, \sigma_0^2\mathbf{I}_2)\quad  &\mathrm{(prior)}, \\
    w_m &\sim \mathcal N(b, C)\quad  &\mathrm{(posterior)}.
\end{aligned}
\end{equation}
Here $\sigma_0^2, b, \sqrt{C}$ are parameters. Note that $\sqrt{C}$ is used instead of $C$ to enforce that $C$ is positive semi-definite. The frequencies are sampled $\omega_m \sim \mathcal N(0, \Lambda^{-1})$ where $\Lambda$ is a diagonal matrix as an additional parameter. The RFF basis representation approximates a GP prior with Gaussian kernel function. Note that one only needs to differentiate the RFF basis, which can be done analytically, to obtain the Hamiltonian gradient $\nabla H$ required for the vector field. The gradient of the Hamiltonian is given by:

\begin{equation}
\begin{aligned}
    \nabla \mathcal H(\mathbf{q},\mathbf{p}) &= \sum_{m=1}^M w_m^{\top}\nabla \phi_m(\mathbf{q},\mathbf{p}), \\
    \nabla \phi_m(\mathbf{x}) &= 
    \begin{bmatrix}
        - \sin(\omega_m^{\top}\mathbf{x}) \omega_m^{\top}\\ \cos(\omega_m^{\top}\mathbf{x}) \omega_m^{\top}
    \end{bmatrix}.
\end{aligned}
\end{equation}

\paragraph{Relation to SSGP}
Although the basic form of a symplectic RFF basis for a GP is adapted from original SSGP paper \cite{tanaka2022ssgp}, we enforce additional physically-aware constraints as additional loss terms. Our model does not confine ourselves into RFF basis representation when a port-Hamiltonian system is introduced, where in original SSGP the external forces are not considered as an explicit module to be learned.

\subsection{Parameterizations for Different Dynamics}
\label{sec:parameters}
Based on RFF representation of Hamiltonian energy function, we justify how to parameterize three different classes of Hamiltonian dynamics.

\paragraph{Conservative Hamiltonian Systems} 
For the conservative case \eqref{eq:hd:conservative}, the set of learnable parameters $\theta$ is composed of those for the GP-RFF surrogate and for data uncertainty. The surrogate is defined by the RFF prior variance $\sigma_0^2$, posterior parameters $b$ and $\sqrt{C}$, and length-scale matrix $\Lambda$. The uncertainty is modeled by the standard deviation of the initial condition, $a$, and the observation noise, $\sigma$. The complete parameter set is therefore:
\begin{equation}
    \theta = (\sigma_0, b, \sqrt{C}, \Lambda, a, \sigma).
\end{equation}

Then the parameterized dynamics are:
\begin{equation}
    \dot x = J \sum_{m=1}^M w_m^T \nabla \phi_m(x).
\end{equation}
The $q-p$ dynamics are:
\begin{equation}
\begin{aligned}
    \dot q &= \sum_{m=1}^M w^T_m \nabla_p \phi_m(q,p) \\
    \dot p &= -\sum_{m=1}^M w^T_m \nabla_q \phi_m(q,p)
\end{aligned}
\end{equation}

\paragraph{Dissipative Hamiltonian Systems}  For the case of dissipative Hamiltonian dynamics \eqref{eq:hd:dissipative}, additional parameters are needed for the dissipation term, specifically the matrix $D$. The dissipation matrix is not directly parameterized since it has a certain structure. We assume the structure of a diagonal matrix with the first half of the diagonal (affecting $\mathbf{q}$) being zero and the second half of the diagonal (affecting $\mathbf{p}$) being non-positive. Thus, the form is $D=diag(0,\cdots, 0, D_1, \cdots , D_d), D_i\leq 0$. To enforce non-positivity of $D_i$, we use a parameter $\eta\in \mathbb{R}^d$ and set $D_i = - \eta_i^2$. Thus, the full parameter array for learning dissipative Hamiltonian dynamics is:
\begin{equation}
    \theta = (\sigma_0, b,\sqrt{C}, \Lambda, a, \sigma, \eta).
\end{equation}

The parameterized dissipative Hamiltonian dynamics are:
\begin{equation}
    \dot x = (J+D) \sum_{m=1}^M w_m^T \nabla \phi_m(x).
\end{equation}
The $q-p$ dynamics are:
\begin{equation}
\begin{aligned}
    \dot q &= \sum_{m=1}^M w^T_m \nabla_p \phi_m(q,p) \\
    \dot p &= -\sum_{m=1}^M w^T_m (\nabla_q \phi_m(q,p) + \eta^2 \odot \nabla_p \phi_m(q,p))
\end{aligned}
\end{equation}

\paragraph{Port-Hamiltonian Systems} For port Hamiltonian dynamics \eqref{eq:hd:port} the external forcing term $F(t)$ must be parameterized. We assume forcing directly effects the momentum only and thus $F(t) = [0 \; F_\vartheta(t)]^T$, where the parameterized forcing function (for $p$) is $F_\vartheta (t)$ with parameters $\vartheta$. We set $F_\vartheta$ to be an MLP neural network. The full parameter array for learning port-Hamiltonian dynamics is:
\begin{equation}
    \theta = (\sigma_0, b,\sqrt{C}, \Lambda, a, \sigma, \eta, \vartheta).
\end{equation}

Then the port-Hamiltonian dynamics are as follows:
\begin{equation}
    \dot x = (J+D) \sum_{m=1}^M w_m^T \nabla \phi_m(x) +F_\vartheta(t).
\end{equation}

The parameterized $q-p$ dynamics are as follows:
\begin{equation}
\begin{aligned}
    \dot q &= \sum_{m=1}^M w^T_m \nabla_p \phi_m(q,p), \\
    \dot p &= -\sum_{m=1}^M w^T_m (\nabla_q \phi_m(q,p) + \eta^2 \odot \nabla_p \phi_m(q,p)) + f_\vartheta(t).
\end{aligned}
\end{equation}

\subsection{Probabilistic Objective and Regularizers}
Having defined the surrogate model, we now introduce the objective function and physics-informed regularizers used for training. The training objective combines (i) a variational data–likelihood term that follows directly from the probabilistic generative model, and (ii) soft physics constraints that bias the learned vector field toward the desired class of Hamiltonian dynamics.

\subsubsection{Generative Model and Variational Inference}
For each trajectory $i$ we draw an initial phase space state 
$
\mathbf{x}_{0}^{(i)} \sim p(\mathbf{x}_{0})=\mathcal{N}(\mathbf{0},a^{2}\mathbf{I}),
$
integrate the parameterized vector field to obtain dynamical state
$
\mathbf{x}_{ij}=\Phi_{\theta}(t_{ij};\mathbf{x}_{0}^{(i)}),
$
and obtain noisy observations
$
\mathbf{y}_{ij}\sim\mathcal{N}(\mathbf{x}_{ij},\sigma^{2}\mathbf{I}),\;
j=1,\dots,J_{i}.
$
Let $\mathbf{W}$ denote the collection of RFF weights. With the usual conditional independence assumptions the joint distribution factorizes as
\[
p(\mathcal{D},\mathbf{X},\mathbf{W})
=
\Bigl(\!\prod_{i,j} p(\mathbf{y}_{ij}\mid\mathbf{x}_{ij})\Bigr)
\Bigl(\!\prod_{i} p(\mathbf{x}_{0}^{(i)})\Bigr)
p(\mathbf{W}),
\]
where $\mathbf{X}$ stacks all phase space states along all trajectories.
We choose the mean–field variational family
\(
q(\mathbf{W},\mathbf{X}_{0})=q(\mathbf{W})\prod_{i}q(\mathbf{x}_{0}^{(i)})
\)
with
$
q(\mathbf{W})=\prod_{m}\mathcal{N}(\mathbf{w}_{m}\mid\mathbf{b}_{m},\mathbf{C}_{m})$ and
$
q(\mathbf{x}_{0}^{(i)})=\mathcal{N}(\boldsymbol{\mu}_{i},\mathrm{diag}\,\boldsymbol{\sigma}_{i}^{2}).
$
Standard manipulations of maximization of joint distribution log likelihood yield the evidence lower bound (ELBO) loss
\begin{equation}
\label{eq:elbo}
\begin{aligned}
    \mathcal{L}_{\mathrm{ELBO}} ={} & \sum_{i=1}^{I}\sum_{j=1}^{J_{i}}
    \mathbb{E}_{q}\!\left[
    \log\mathcal{N}\!\bigl(\mathbf{y}_{ij}\mid\mathbf{x}_{ij},\sigma^{2}\mathbf{I}\bigr)
    \right] \\
    & - \mathrm{KL}\!\bigl(q(\mathbf{W})\|p(\mathbf{W})\bigr)  - \sum_{i=1}^{I}\mathrm{KL}\!\bigl(q(\mathbf{x}_{0}^{(i)})\|p(\mathbf{x}_{0})\bigr).
\end{aligned}
\end{equation}
In practice, the expectation in the ELBO is estimated via Monte Carlo: we sample $\mathbf{w}\sim q(\mathbf{w})$ and $\boldsymbol{\omega}\sim q(\boldsymbol{\omega})$, draw each initial state $\mathbf{x}_{i0}\sim q(\mathbf{x}_{i0})$, propagate it deterministically through $\Phi_{\theta}$ using a differentiable ODE solver \cite{chen2018neuralode}, and average the resulting log‑likelihoods of $\{\mathbf{y}_{ij}\}$.

\subsubsection{ELBO Derivation}

Here we derive the ELBO, which follows that given in \cite{tanaka2022ssgp}. All random variables are clearly identified, and the deterministic nature of the ODE flow is handled correctly, yielding an evidence lower bound (ELBO) containing only those terms that truly require approximation. 

Let $\mathcal{D}=\{\mathbf{y}_{ij}\}_{i=1{:}I,\,j=1{:}J_i}$ be the noisy observations sampled at times $\{t_{ij}\}$.  
The model is specified by  
\begin{enumerate}
    \item Latent RFF weights:
          $\displaystyle \mathbf{W}\sim \mathcal{N}(\mathbf{0},\sigma_0^{2}\mathbf{I})$;
    \item Latent initial states:
          $\displaystyle \mathbf{x}_{i0}\sim \mathcal{N}(\mathbf{0},a^{2}\mathbf{I})\quad(i=1,\dots,I)$;
    \item Deterministic trajectory evolution:
          \[
          \mathbf{x}_{ij}
          =\Phi_{\boldsymbol{\theta}}\!\left(t_{ij};\,\mathbf{x}_{i0},\mathbf{W}\right),
          \]
          where $\Phi_{\boldsymbol{\theta}}$ is the flow generated by integrating the parameterized vector field;
    \item Noisy observations:
          $\displaystyle
          \mathbf{y}_{ij}\mid\mathbf{x}_{ij}
          \sim\mathcal{N}(\mathbf{x}_{ij},\sigma^{2}\mathbf{I}).
          $
\end{enumerate}
Because every state after $t=0$ is a deterministic function of $(\mathbf{W},\mathbf{x}_{i0})$, the joint density factorizes as
\begin{equation}
p(\mathcal{D},\mathbf{X}_{0},\mathbf{W})
= \Bigl[\prod_{i,j} p(\mathbf{y}_{ij}\mid\mathbf{x}_{ij})\Bigr]
  \Bigl[\prod_{i} p(\mathbf{x}_{i0})\Bigr]\,
  p(\mathbf{W}),
\end{equation}
with no separate probability measure required for the intermediate states $\mathbf{x}_{ij}$.

We adopt a mean-field Gaussian family for the latent variables:
\begin{equation}
q(\mathbf{W},\mathbf{X}_{0})
   = q(\mathbf{W})\prod_{i=1}^{I} q(\mathbf{x}_{i0}),
\end{equation}
with
\begin{equation}
q(\mathbf{W})=\prod_{m=1}^{M}\mathcal{N}(\mathbf{w}_{m}\mid\mathbf{b}_{m},\mathbf{C}_{m}),\qquad
q(\mathbf{x}_{i0})=\mathcal{N}(\boldsymbol{\mu}_{i},\operatorname{diag}\boldsymbol{\sigma}_{i}^{2}).
\end{equation}
No variational factor is needed for the deterministic states $\{\mathbf{x}_{ij}\}_{j\ge 1}$.

Applying Jensen’s inequality to $\log p(\mathcal{D})$ and using the above factorizations gives
\begin{align}
\mathcal{L}_{\mathrm{ELBO}}
   &= \sum_{i=1}^{I}\sum_{j=1}^{J_i}
      \mathbb{E}_{q(\mathbf{W})q(\mathbf{x}_{i0})}
      \Bigl[\log \mathcal{N}\!\bigl(\mathbf{y}_{ij}\,\bigm|\,
            \Phi_{\boldsymbol{\theta}}(t_{ij};\mathbf{x}_{i0},\mathbf{W}),
            \sigma^{2}\mathbf{I}\bigr)\Bigr] \nonumber\\
   &\quad - \operatorname{KL}\!\bigl(q(\mathbf{W})\,\|\,\mathcal{N}(\mathbf{0},\sigma_{0}^{2}\mathbf{I})\bigr)
         - \sum_{i=1}^{I}\operatorname{KL}\!\bigl(q(\mathbf{x}_{i0})\,\|\,\mathcal{N}(\mathbf{0},a^{2}\mathbf{I})\bigr).
\label{eq:elbo_final_app}
\end{align}
Only three terms appear: a data-fit term and two KL regularizers -- one for the RFF weights and one for the initial conditions.  No KL term for the intermediate states occurs, as they are deterministic.

\subsubsection{Conservation Law Regularizers}
\label{sec:loss}
Conservation laws that apply to all conservative Hamiltonian systems include the conservation of energy and the conservation of volume (Liouville's Theorem) \cite{meiss2007hamsys}. Note that in the dissipative and port Hamiltonian cases, we evaluate these losses on only the conservative portion of the dynamics, that is using the vector field $J \nabla \mathcal H$.

\textbf{Conservation of Energy:} The law of conservation of energy states that the total energy of the system is conserved. In a Hamiltonian system, the Hamiltonian $\mathcal H(\mathbf q, \mathbf p)$ is itself equal to the total energy of the system. Thus, the conservation law is:
\begin{equation}
    \mathcal{H}(\mathbf{q}(t), \mathbf{p}(t)) = \mathcal{H}(\mathbf{q}(0), \mathbf{p}(0)),
    \label{eq:cons-energy}
\end{equation}
which can be softly enforced with the following additional loss term:
\begin{equation}
    \mathcal{L}_{\mathrm{Energy}} = \mathbb{E}_{\mathbf{q}, \mathbf{p}, t} \left[ \mathcal{H}(\rho_t (\mathbf{q}, \mathbf{p})) - \mathcal{H}(\mathbf{q}, \mathbf{p}) \right]^2.
\end{equation}
The Hamiltonian flow map \(\rho_t(\cdot)=\Phi_{\boldsymbol{\theta}}(t;\cdot,\mathbf{w},\boldsymbol{\omega})\) is the mapping from phase space at the initial time to a future time, defined by integrating along the Hamiltonian vector field. Note that this conservation law constraint term can be estimated by evaluating on the trajectories which were used to compute the ELBO loss in the previous section.

\textbf{Conservation of Volume:} The law of conservation of volume states that given a set of initial conditions, after evolving the set using Hamiltonian dynamics up to time $t$, the volume of the set at time $t$ is equal to the volume of the initial set. Mathematically,
\begin{equation}
\int_{\rho_t(A)} \mathbf{1}d\mathbf{q} \, d\mathbf{p} = \int_A \mathbf{1}d\mathbf{q} \, d\mathbf{p},
\label{eq:cons-vol}
\end{equation}
where \(A\) is a set of initial conditions in \(\mathbf{q}\)-\(\mathbf{p}\) space. This can be evaluated by sampling volumes within phase space and comparing the portion of points which lie with the volume at various time steps. The loss term is defined as:
\begin{equation}
    \mathcal{L}_{\mathrm{Vol}} = \frac{1}{N} \Bigg[ \sum_{n=1}^{N} \Bigl(\mathbf{1}_{A}(\mathbf{q}_n,\mathbf{p}_n)
    \!-\!\mathbf{1}_{A}(\rho_{t}(\mathbf{q}_n,\mathbf{p}_n))\Bigr)  \Bigg]^2,
\end{equation}
where $\mathbf{1}_A$ refers the characteristic function of a sampled ``volume" $A$. In implementation we sample a rectangular domain $A$ in each epoch and the loss is computed under different sampled time stamps. 

\subsubsection{Lyapunov Stability Regularizer}
In the context of ordinary differential equations (ODEs), stability refers to the behavior of solutions as time progresses. For an ODE $x' = f(x)$, asymptotic stability implies that the solution converges to an equilibrium $\lim_{t \rightarrow \infty} x(t) = x^*$. Exponential stability ensures that $\|x(t)-x^*\| \leq C e^{-\alpha t} \|x(0)-x^*\|$ for some constants $C$ and $\alpha$. Lyapunov stability conditions can be used to verify asymptotic or exponential stability. For an ODE $x' = f(x)$, a system is \textbf{Lyapunov stable} if there exists a Lyapunov function $V(x)$ such that $V(x^*) = 0$, $V(x) > 0$ for all $x \neq x^*$ and $\frac{d}{dt} V(x(t)) \leq 0$.
It is \textbf{asymptotically stable} additionally if $\frac{d}{dt} V(x(t)) < 0$.
It is \textbf{$\alpha$ exponentially stable} if $\frac{d}{dt} V(x(t)) \leq -\alpha V(x(t))$ and $c_1 \|x-x^*\| \leq V(x) \leq c_2 \|x-x^*\|$. Prior works \cite{manek2019stabledeepdynamics,rodriguez2022lyanet} have considered the use of Lyapunov stability conditions in learning dynamical systems, but our work applies this to the problem of probabilistic learning of Hamiltonian dynamics.

Given the nature of Hamiltonian systems, the Hamiltonian $H$ itself can serve as the Lyapunov function, $V = H$. Hamiltonian systems should be Lyapunov stable, so we set $\alpha=0$.
Additionally, the positive-definiteness of the Lyapunov function (i.e. the Hamiltonian) must be enforced. Considering all of these constraints, the Lyapunov loss function is:
\begin{equation}
\begin{aligned}
\mathcal{L}_{\mathrm{Lyap}} &= \int_0^{\top}\lambda_{1,1}\mathrm{ReLU}\left( \frac{d}{dt}H(x(t)) \right)dt \\
&+ \int_0^T\lambda_{1,2} \mathrm{ReLU}(-H(x(t))) dt.
\end{aligned}
\end{equation}
\begin{figure}[ht]
    \centering    \includegraphics[width=0.6\linewidth]{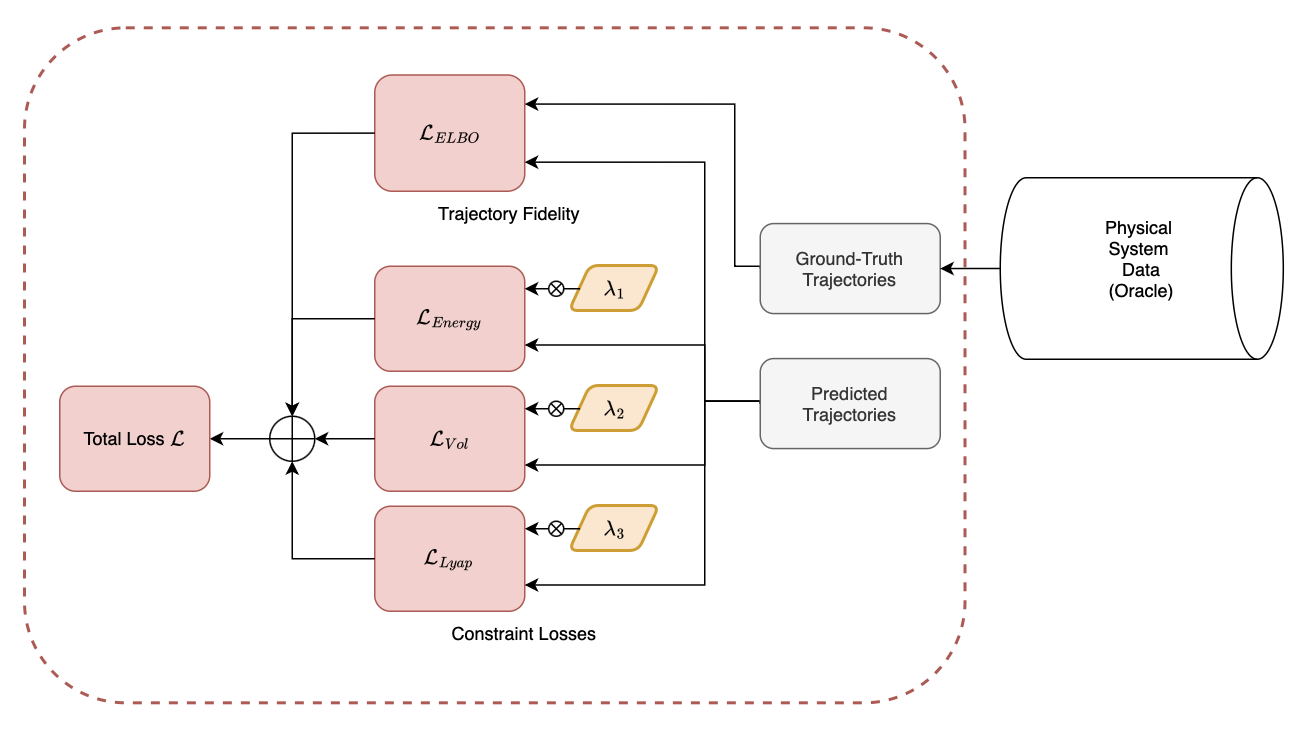}
    \caption{The breakdown of our loss function into the ELBO, energy and volume conservation, and Lyapunov stability terms. See Section \ref{sec:loss} for details.}
    \label{fig:loss-breakdown}
\end{figure}

\subsection{Changes in Conservation Laws for Dissipative and Port-Hamiltonian Systems}
\begin{table}[]
    \centering
    \begin{tabular}{c|c|c|c}
        \hline
        Class & Physical Properties & Form of Laws & Model Parameterization ($\theta$) \\
        \hline
        Conservative & Conservation of Energy/Volume & Eq. \ref{eq:cons-energy}, Eq. \ref{eq:cons-vol} & $\sigma_0, b,\sqrt{C}, \Lambda, a, \sigma$  \\
        Dissipative & Dissipation of Energy/Volume & Eq. \ref{eq:diss-energy}, Eq. \ref{eq:diss-vol} & $\sigma_0, b,\sqrt{C}, \Lambda, a, \sigma, \eta$  \\
        Port-Hamiltonian & Dissipation/External Input of Energy/Volume & Eq. \ref{eq:port-energy}, Eq. \ref{eq:diss-vol} & $\sigma_0, b,\sqrt{C}, \Lambda, a, \sigma, \eta, \vartheta$ \\
        \hline
    \end{tabular}
    \caption{Breakdown of the physical constraints, their numerical forms, and the model parameterization for three classes: conservative, dissipative, and port-Hamiltonian dynamics}
    \label{tab:classes}
\end{table}

The ELBO loss described for the conservative case remains the same for the dissipative and port-Hamiltonian cases, as does the Lyapunov stability loss. However, the conservation law loss must be modified since in dissipative systems energy and volume are not longer constant in time. Since the model separates the conservative, dissipative, and forcing terms of the dynamics, the conservative laws can be enforced by integrating the conservative part only, which is what our implementation does. Below we examine how energy and volume are affected by dissipative and external forcing terms. \\
For energy, consider the time derivative of the Hamiltonian:
\begin{equation}
\begin{aligned}
    \frac{d}{dt} \mathcal H(q(t), p(t)) &= \nabla \mathcal H^T (J+D) \nabla \mathcal H \\
    &= \cancel{\nabla \mathcal H^T J \nabla \mathcal H} + \nabla \mathcal H^T D \nabla \mathcal H \\
    &= - \sum_{i=1}^d \left( \eta_i \frac{\partial \mathcal H}{\partial p_i} \right)^2
\end{aligned}
\end{equation}
Then the dissipation of energy can be written as
\begin{equation}
    \frac{d \mathcal H}{dt} + \sum_{i=1}^d \left( \eta_i \frac{\partial \mathcal H}{\partial p_i} \right)^2 = 0.
    \label{eq:diss-energy}
\end{equation}
For port-Hamiltonian systems, the energy evolution can be written as
\begin{equation}
    \frac{d \mathcal H}{dt} + \sum_{i=1}^d \left( \eta_i \frac{\partial \mathcal H}{\partial p_i} \right)^2 - \nabla H^T F(t) = 0.
    \label{eq:port-energy}
\end{equation}
For volume, we consider the divergence of the vector field:
\begin{equation}
\begin{aligned}
    \nabla \cdot \mathbf f &= \nabla \cdot ((J+D)\nabla \mathcal H) \\
    &= \cancel{\nabla \cdot J \nabla \mathcal H} + \nabla \cdot D \nabla \mathcal H \\
    &= - \sum_{i=1}^d \eta_i^2 \frac{\partial^2 \mathcal H}{\partial p_i^2}.
\end{aligned}
\end{equation}
Then the dissipation of volume can be written as:
\begin{equation}
    \nabla \cdot \mathbf f + \sum_{i=1}^d \eta_i^2 \frac{\partial^2 \mathcal H}{\partial p_i^2} = 0.
    \label{eq:diss-vol}
\end{equation}
The port-Hamiltonian version of this is the same since the external forcing term $F(t)$ is only time-dependent and thus its divergence is zero.

\subsection{Comparison to Onsager-Machlup Functional}
The Onsager-Machlup function $OM$ \cite{durr1978onsager} \cite{zhang2024onsager} is an analog of the Lagrangian for SDEs. It has the property that a minimizer $z$ of the Onsager-Machlup functional, $\int_0^T OM(z,\dot z)dt$, is the most probable path for the associated SDE. The Onsager-Machlup function for an SDE of the form:
\begin{align}
    d\mathbf x_t = \mathbf f(\mathbf x)dt + d\mathbf W_t
\end{align}
is 
\begin{equation}
    OM(\mathbf z, \dot{\mathbf z}) = \frac{1}{2} \left\| \dot{\mathbf{z}} - \mathbf f(\mathbf z) \right\|^2 + \frac{1}{2} (\nabla \cdot \mathbf f(\mathbf z)).
\end{equation}
For a Hamiltonian system with dissipation and noise of the form:
\begin{align*}
    d\mathbf x_t = [(J+D) \nabla \mathcal H(\mathbf x) + F(t)]dt + d\mathbf W_t
\end{align*}
the Onsager-Machlup function is:
\begin{align}
    OM(\mathbf z, \dot{\mathbf z}) &= \frac{1}{2} \left\| \dot{\mathbf{z}} - [(J+D)\nabla \mathcal H(\mathbf z) + F(t)] \right\|^2 + \frac{1}{2} (\nabla \cdot [(J+D)\nabla \mathcal H(\mathbf z) + F(t)]) \\
    &= \frac{1}{2} \left\| \dot{\mathbf{z}} - (J+D)\nabla \mathcal H(\mathbf z) - F(t)\right\|^2 + \frac{1}{2} (\nabla \cdot D\nabla \mathcal H(\mathbf z))
\end{align}
with the antisymmetric matrix term disappearing. Thus, if there is no dissipation, then the Onsager-Machlup function reduces to an $L^2$ error. 
Other work has used the Onsager-Machlup functional in machine learning. In \cite{li2021mlonsager}, it is used to derive a Hamiltonian system for the most probable path between the initial and terminal conditions. In \cite{haas2014entropyonsager} it is used to evaluate the entropy of trajectories.
Note that the first term is simply the $L^2$ loss of the predicted trajectories. The second term (the entropy term) is the divergence of the vector field. This is equivalent to conservation of volume based on Liouville's theorem, which states that volume-conserving vector fields are divergence-free \cite{meiss2007hamsys}.

\subsection{Numerical Recipe of Regularized Loss Term}

\paragraph{Conservation of Energy}
To compute the ELBO loss, we already generate trajectories for different initial conditions. We can evaluate to Hamiltonian along these trajectories and check the difference from initial time. Thus, the term may be estimated as follows:
\begin{equation}
     \mathcal{L}_{Energy} \approx \sum_{i=1}^I\sum_{j=1}^J\sum_{k=1}^{N_x} \frac{1}{IJ_i N_x} \left( \mathcal{H}^{\theta} \left( \mathbf{x}_{ij}^{(k)} \right) - \mathcal{H}^{\theta}\left( \mathbf{x}_{i0}^{(k)} \right) \right)^2.
\end{equation}

Here $N_x$ refers number of GP process realization of $\mathcal{H}^{\theta}$. Note that these conservation law constraint terms can be estimated by evaluating on the trajectories which were generated to compute the ELBO loss in the previous section.

\paragraph{Conservation of Volume}
The loss term is
\begin{equation}
    \mathcal{L}_{\mathrm{Vol}} = \frac{1}{N} \Bigg[ \sum_{n=1}^{N} \Bigl(\mathbf{1}_{A}(\mathbf{q}_n,\mathbf{p}_n)
    \!-\!\mathbf{1}_{A}(\rho_{t}(\mathbf{q}_n,\mathbf{p}_n))\Bigr)  \Bigg]^2,
\end{equation}
where $\mathbf{1}_A$ refers the characteristic function of a sampled ``volume" $A$. In implementation we sample a rectangular domain $A$ in each epoch and the loss is computed under different sampled time stamps. 
\paragraph{Lyapunov Stability}
    In order to estimate this, the loss can be computed along the same trajectories as were generated for the computation of the ELBO loss and also used for the conservation of energy and volume. This estimation is the following:
    \begin{align}
        \mathcal{L}_{Lyap} &\approx  \sum_{i=1}^I\sum_{j=1}^J\sum_{k=1}^{N_x} \bigg[ \lambda_{1,1} ReLU \left( \frac{\mathcal{H}^{\theta} \left( \mathbf x_{i,j+1}^{(k)} \right) - \mathcal{H}^{\theta}  \left( \mathbf x_{i,j}^{(k)} \right)}{\Delta t} \right)  + \lambda_{1,2} ReLU \left( -\mathcal{H}^{\theta}  \left( \mathbf x_{i,j}^{(k)} \right) \right) \bigg]
    \end{align}
In practice, since the first term is already satisfied by enforcing conservation of energy we omit it in the code implementation.

\subsubsection{Dependence of Loss Terms on Parameter Groups}
We have several groups of parameters as well as a multi-term loss function. The parameters required in the computation of each loss function are as follows:
\begin{itemize}
    \item ELBO: $\sigma_0, b,\sqrt{C}, \Lambda, a, \sigma, \eta, \vartheta$ (all)
    \begin{itemize}
        \item KL-divergence (initial condition): $a$
        \item KL-divergence (RFF weights): $b, \sqrt{C}, \sigma_0$
        \item Negative log-liklihood: $b,\sqrt{C}, \Lambda, \eta, \vartheta$
    \end{itemize}
    \item Conservation of Energy: $b,\sqrt{C}, \Lambda$
    \item Conservation of Volume: $b,\sqrt{C}, \Lambda$
    \item Lyapunov Stability: $b,\sqrt{C}, \Lambda, \eta, \vartheta$
\end{itemize}
\subsection{Optimization and Training}
\label{sec:query}
\subsubsection{Soft‑Constrained Objective}
\label{sec:soft}
The ideal learning problem is a likelihood–maximization subject to exact
physics constraints.  Writing it in minimization form (we minimize
the negative ELBO) gives the constrained program:
\begin{equation}
\begin{aligned}
\underset{\theta}{\operatorname{argmin}}\;&
      \mathcal L_{\mathrm{ELBO}}(\theta) \\[2pt]
\text{s.t.}\;&(\dot{\mathbf{q}}_i, \dot{\mathbf{p}}_i)=(J+D)\nabla\mathcal H(\mathbf{q}_i,\mathbf{p}_i)+F(t), \\[-1pt]
      &(\mathbf{q}_{i0},\mathbf{p}_{i0})\sim\mathcal N(\mathbf y_{i0},a^{2}\mathbf I), \\[2pt]
      &\bigl(\mathcal H\circ\rho_t\bigr)(\mathbf{q}_{i0},\mathbf{p}_{i0})=\mathcal H(\mathbf{q}_{i0},\mathbf{p}_{i0}), \\[-1pt]
      &\int_{\rho_t(A)} \mathbf{1}d\mathbf{q}d\mathbf{p} = \int_{A}\mathbf{1}d\mathbf{q}d\mathbf{p}, \\[-1pt]
      &\tfrac{d}{dt}\mathcal H(\mathbf{q}_{ij},\mathbf{p}_{ij})\le
        -\alpha\,\mathcal H(\mathbf{q}_{ij},\mathbf{p}_{ij}),\\
    &\mathcal H(\mathbf{q}_{ij},\mathbf{p}_{ij})\ge0,\\
      &\forall\,i=1{:}I,\;j=0{:}J_i.
\end{aligned}
\label{eq:constraint:opt}
\end{equation}

We relax these hard constraints with weighted penalty terms, producing the
soft objective:
\begin{equation}
\begin{aligned}
\mathcal{L}\bigl(\theta,\boldsymbol{\lambda}\bigr)
    &= \mathcal{L}_{\mathrm{ELBO}}(\theta)
       + \lambda_{1}\,\mathcal{L}_{\mathrm{Lyap}}(\theta) \\
    &\quad + \lambda_{2}\,\mathcal{L}_{\mathrm{Energy}}(\theta)
       + \lambda_{3}\,\mathcal{L}_{\mathrm{Vol}}(\theta).
\end{aligned}
\end{equation}

The vector \(\boldsymbol\lambda\) is adapted during training using the
gradient descent / ascent (GDA) scheme described in the next
paragraph.  When all \(\lambda_{k}=0\) the method reduces to pure
variational inference; when they are large the optimization approaches a
hard‑constrained formulation.

\subsubsection{Balancing the Multi-Term Loss}
\label{sec:hyperparam}
The loss consists of various terms, which must be carefully balanced in order to prevent under- or over-constrained optimization:
Hyperparameters $\boldsymbol{\lambda}$ can be balanced by solving a max-min optimization problem for both the parameters and hyperparameters:
\begin{equation}
    \min_{{\theta}} \max_{\boldsymbol{\lambda}} \mathcal{L}( \theta, \boldsymbol \lambda).
\end{equation}
Both can be updated using gradient-based optimization methods (SGD, Adam, etc.), but for the hyperparameters we use ascent rather than descent since it is a maximization problem. This is a variant of the Gradient Descent-Ascent (GDA) \cite{lin2020gda} \cite{zhang2020sgda}:
\begin{align}
    \theta_{k+1} &= \theta_k - g_1 \left( \frac{\partial \mathcal{L}}{\partial \theta} \right), \\  
    \boldsymbol\lambda_{k+1} &= \boldsymbol\lambda_k + g_2 \left( \frac{\partial \mathcal{L}}{\partial \boldsymbol\lambda} \right), 
\end{align}
where $g_1,g_2$ depend on the optimizer used. Note that the gradient of the loss with respect to the hyperparameters can be directly computed since the loss is linear with respect to the hyperparameters. In our experiments, we compare the use of no balancing (all lambdas equal to one) with GDA-based balancing. We test two variants of GDA. The first uses Adam \cite{kingma2014adam} for the parameter update and gradient ascent for the hyperparameters. The second uses Adam for both updates. Additional, we consider MTAdam \cite{malkiel2020mtadam} as an alternative way of balancing multiple loss terms. Rather than solving a min-max problem with updates to $\lambda$, MTAdam groups the parameters and within each group normalizes the gradients of the individual loss terms. Jacobian descent \cite{quinton2024jacobiandescent} is another method, which computes the Jacobian of the loss vector with respect to the parameter vector, and then applies one of several aggregators to the Jacobian. Aggregators such as UPGrad are meant to construct the update such that each individual loss is improved, rather than one loss being improved at the expense of another. See Table \ref{tab:noisy-loss-comp}.

\subsubsection{Training Algorithm}
The complete training loop is summarized in Algorithm \ref{alg:main}. We learn the model parameters by querying an oracle for noisy trajectory data and updating the parameters based on the gradient of the total balanced loss function.
\begin{algorithm}[H]
\caption{Learning Generalized Hamiltonian Dynamics}
\label{alg:main}
\begin{algorithmic}[1]
\Require Oracle $\mathcal{O}$, epochs $K$, batch size $B$, horizon $T$, learning rate $\alpha$
\State Initialize parameters $\theta_0$ \Comment{Sections \ref{sec:hnet} \& \ref{sec:parameters}}
\For{$k = 1$ to $K$}
    \State Query $\{\mathbf{y}_t^{(b)} = (\mathbf{q}_t^{(b)}, \mathbf{p}_t^{(b)})\}_{b=1,t=1}^{B,T}$ from $\mathcal{O}$
    \For{$b = 1$ to $B$}
        \State Initialize $\mathbf{x}_0^{(b)} = (\mathbf{q}_0^{(b)}, \mathbf{p}_0^{(b)})$.
        \State Sample RFF weights $w \sim \mathcal{N}(b, \sqrt{C})$.
        \State Sample RFF frequencies $\omega \sim \mathcal{N}(0, \Lambda^{-1})$.
        \State Construct the Hamiltonian $\mathcal H$, its gradient  $\nabla \mathcal H$, and the vector field$f$.  \Comment{Sections \ref{sec:hnet} \& \ref{sec:parameters}}
        \State Integration step: $\{ \mathbf{x}_{t}^{(b)} \}_{t=1}^{\top}\gets \mathrm{odeint}(\mathbf{x}_0^{(b)}, f, T \})$.
    \EndFor
    \State Compute total loss $\mathcal{L}(\theta_k, \boldsymbol{\lambda}_k)$. \Comment{Section \ref{sec:loss}}
    \State Compute loss gradient $\nabla_\theta \mathcal{L}(\theta_k, \boldsymbol{\lambda}_k)$. 
    \State Parameter update: $\theta_{k+1} \gets \text{Adam}(\theta_k, \nabla_\theta \mathcal{L}, \alpha)$
    \State Hyperparameter update:  
    $\boldsymbol{\lambda}_{k+1} \gets \boldsymbol{\lambda}_k + \alpha \nabla_{\boldsymbol{\lambda}}\mathcal{L}$ \Comment{Section \ref{sec:hyperparam}}
\EndFor
\State \Return Optimized parameters $\theta_K$
\end{algorithmic}
\end{algorithm}
\begin{table}[!htbp]
    \centering
    \begin{tabular}{c|c|c|c|c|c|c|c}
        \hline
        \textbf{Dataset} & \textbf{Type} & \textbf{Trajs.} & \textbf{Steps} & \textbf{Noise} & \textbf{Damping} & \textbf{Forcing} & \textbf{Other Params} \\
        \hline
        Single Pendulum & Cons.     & 100 & 100 & 0.1 & N/A & N/A & $m = 1 , g = 9.81 , \ell = 1$ \\
        Damped Pendulum & Diss.       & 100 & 100 & 0.1 & 0.1 & N/A & $m = 1 , g = 9.81 , \ell = 1$  \\
        Windy Pendulum  & Port  & 100 & 100 & 0.01 & 0.1 & 0.1t & $m = 1 , g = 9.81 , \ell = 1$ \\
        Mass-Spring & Cons. & 100 & 100 & 0.1 & N/A & N/A & $m = 1, k = 1$ \\
        Damped Spring & Diss. & 100 & 100 & 0.1 & 0.1 & N/A & $m = 1, k = 1$ \\
        Forced Spring & Port & 100 & 100 & 0.1 & 0.1 & $0.1 \sin(t) \sin(2t)$ & $m = 1, k = 1$ \\
        Henon-Heiles & Cons. & 100 & 100 & 0.1 & N/A & N/A & $m = 1, \alpha = 1$ \\
        Unforced Duffing & Diss. & 100 & 100 & 0.1  & 0.3 & N/A & $\alpha = -1, \beta = 1$ \\
        Forced Duffing & Port & 100 & 100 & 0.1 & 0.1 & 0.39 & $\alpha = -1, \beta = 1$ \\
        \hline
    \end{tabular}
    \caption{Parameters for the setup of the various Hamiltonian dynamics trajectory datasets.}
    \label{tab:data-config}
\end{table}

\subsection{A Priori Knowledge of Noise Levels}
We consider the case where the level of observation noise in the dynamics is known. In this case, both the parameterization and the loss function are slightly modified. The parameterization is reduced because the parameters $(\sigma, a)$ used for modeling the noise level are no longer needed. Thus, the parameterizations for the three classes of dynamics are as follows:

\begin{itemize}
    \item Conservative: $\sigma_0, b,\sqrt{C}, \Lambda.$ 
    \item Dissipative: $\sigma_0, b,\sqrt{C}, \Lambda, \eta.$
    \item Port-Hamiltonian: $\sigma_0, b,\sqrt{C}, \Lambda, \eta, \vartheta.$
\end{itemize}

For the loss function, the KL term on the initial conditions is unnecessary since $a$ is known. The negative log-likelihood term is modified by removing the term dependent only on $\sigma$. For classes, where $\sigma$ is known to be zero, the negative log-likelihood is not well defined, so we use MSE as a replacement.

\begin{table}[h]
    \centering
    \begin{tabular}{c|c|c|c|c|c}
        \hline
        Class & System & (D,P)HNN & SSGP & Ours (Equal) & Ours (GDA) \\
        \hline
        Cons. & Single Pendulum & $0.5922 \pm 1.4887$ & $1.0186 \pm 0.8825$ & \textbf{0.1404} $\pm$ 0.2166 & \textit{0.2834} $\pm$ 0.3872 \\
        Diss. & Damped Spring & $0.0347 \pm 0.0640$ &  0.0275 $\pm$ 0.0135 &  \textbf{0.0269} $\pm$ 0.0127 &  \textit{0.0273} $\pm$ 0.0130  \\
        Diss. & Damped Pendulum & $0.3312 \pm 0.4097$ &  $0.1377 \pm 0.1917$ & \textbf{0.0902} $\pm$ 0.1270 &  \textit{0.0929} $\pm$ 0.1021  \\
        Diss. & Unforced Duffing & $0.1790 \pm 0.5278$ & $0.0904 \pm 0.3435$ & \textit{0.0879} $\pm$ 0.2996 & \textbf{0.0876} $\pm$ 0.2914 \\
        Port & Forced Spring & $0.0923 \pm 0.1362$ & \textbf{0.0293} $\pm$ 0.0102 & \textit{0.0311} $\pm$ 0.0126 & $0.0319 \pm 0.0129$ \\
        Port & Forced Duffing & $0.4127 \pm 0.7770$ & $0.4920 \pm 1.0299$ & \textbf{0.2742} $\pm$ 0.2930 & \textit{0.2763} $\pm$ 0.2602 \\
        \hline
    \end{tabular}
    \caption{Evaluation on testing datasets of 6 systems of 3 Hamiltonian classes. We compare HNN variants, SSGP, and our methods. We use HNN \cite{greydanus2019hnn}, DHNN \cite{sosanya2022dhnn}, or PHNN \cite{desai2021port} depending on the associated class of system. For our method, we compare equally weighted loss terms to GDA-balanced terms. We compare using the metric of the MSE loss of the $q,p$ testing trajectories. In each system besides forced spring, the MSE loss is lower for our method, with the equal weighting typically performing better.}
    \label{tab:loss-comp}
\end{table}

\begin{table}[h]
    \centering
    {
    \footnotesize
    \begin{tabular}{lcccccc}
    \hline
    \textbf{Noise} & SSGP & Ours (Equal) & Ours (GDA) & Ours (MTAdam) & Ours (JD) & Ours (Noise Prior) \\
    \hline
    0   & 0.0477 $\pm$ 0.0717 & \textit{0.0365} $\pm$ 0.0536 & \textbf{0.0316} $\pm$ 0.0447 & 0.4440 $\pm$ 0.4672 & 0.0377 $\pm$ 0.0510 & 0.0876 $\pm$ 0.1157 \\
    0.01  & 0.0405 $\pm$ 0.0531 & 0.0470 $\pm$ 0.0606 & \textit{0.0363} $\pm$ 0.0491 & 0.4377 $\pm$ 0.4635 & 0.0458 $\pm$ 0.0593 & \textbf{0.0273} $\pm$ 0.0332 \\
    0.05  & 0.0664 $\pm$ 0.0815 & 0.0927 $\pm$ 0.1102 & \textit{0.0456} $\pm$ 0.0636 & 0.4440 $\pm$ 0.4672 & 0.0957 $\pm$ 0.1140 & \textbf{0.0312} $\pm$ 0.0465 \\
    0.1   & 0.0897 $\pm$ 0.1240 & 0.1097 $\pm$ 0.1511 & 0.1119 $\pm$ 0.1540 & 0.4584 $\pm$ 0.4642 & \textit{0.0734} $\pm$ 0.1015 & \textbf{0.0559} $\pm$ 0.0693 \\
    0.2   & 0.1521 $\pm$ 0.1844 & \textbf{0.1284} $\pm$ 0.1267 & 0.1376 $\pm$ 0.1354 & 0.5071 $\pm$ 0.4716 & \textit{0.1286} $\pm$ 0.1268 & 0.1633 $\pm$ 0.2005 \\
    \hline
    \end{tabular}
    }
    \caption{Comparison of various noise levels evaluated on the testing dataset for the damped pendulum system. We test with uncorrupted data and datasets with Gaussian noise of standard deviation $0.01, 0.05, 0.1,$ and $0.2$. We compare the base SSGP model \cite{tanaka2022ssgp} to our method with various ways of weighting the multi-term loss function, which are equal summation, gradient descent-ascent (GDA), multi-term Adam (MTAdam) \cite{malkiel2020mtadam}, Jacobian descent (JD) with a UPGrad aggregator \cite{quinton2024jacobiandescent}, and GDA with prior knowledge of noise The metric is the MSE loss on the $q,p$ testing trajectories. Here, by applying additional regularization with proper balancing to enforce physical constraints and stability, we achieve superior performance in each case of noise level. In addition, in most cases having a prior knowledge of the noise level further improves the performance.}
    \label{tab:noisy-loss-comp}
\end{table}

\begin{table}[hb]
    \centering
    {
    \scriptsize
    \begin{tabular}{lccccccc}
        \hline
        \textbf{Noise} & SSGP & Ours (Equal) & Ours (GDA) & Ours (MTAdam) & Ours (JD) & Ours (JD2) & Ours (Noise Prior) \\
        \hline
        0    & \textbf{0.0028} $\pm$ 0.0017 & 0.0041 $\pm$ 0.0015 & 0.0044 $\pm$ 0.0016 & 0.2116 $\pm$ 0.2044 & \textit{0.0041} $\pm$ 0.0015 & \textit{0.0036} $\pm$ 0.0019 & 0.0148 $\pm$ 0.0110 \\
        0.01 & \textit{0.0039} $\pm$ 0.0019 & 0.0044 $\pm$ 0.0023 & 0.0046 $\pm$ 0.0024 & 0.2011 $\pm$ 0.1669 & \textit{0.0044} $\pm$ 0.0023 & \textbf{0.0035} $\pm$ 0.0020 & 0.0072 $\pm$ 0.0054 \\
        0.05 & 0.0108 $\pm$ 0.0063 & \textbf{0.0083} $\pm$ 0.0046 & \textit{0.0087} $\pm$ 0.0048 & 0.2880 $\pm$ 0.3215 & \textbf{0.0083} $\pm$ 0.0047 & 0.0077 $\pm$ 0.0052 & 0.0191 $\pm$ 0.0126 \\
        0.1  & \textbf{0.0242} $\pm$ 0.0126 & \textit{0.0307} $\pm$ 0.0161 & 0.0311 $\pm$ 0.0164 & 0.2470 $\pm$ 0.2517 & 0.0307 $\pm$ 0.0161 & 0.0271 $\pm$ 0.0160 & \textit{0.0292} $\pm$ 0.0183 \\
        0.2  & 0.0933 $\pm$ 0.0524 & 0.0918 $\pm$ 0.0591 & \textit{0.0893} $\pm$ 0.0534 & 0.2736 $\pm$ 0.2337 & 0.0917 $\pm$ 0.0590 & 0.0889 $\pm$ 0.0489 & \textbf{0.0859} $\pm$ 0.0495 \\
        \hline
    \end{tabular}
    }
    \caption{Comparison of various noise levels evaluated on the testing dataset for the conservative spring system. We test with uncorrupted data and datasets with Gaussian noise of standard deviation $0.01, 0.05, 0.1,$ and $0.2$. We compare the base SSGP model \cite{tanaka2022ssgp} to our method with various ways of weighting the multi-term loss function, which are equal summation, gradient descent-ascent (GDA), multi-term Adam (MTAdam) \cite{malkiel2020mtadam}, Jacobian descent with a UPGrad aggregator \cite{quinton2024jacobiandescent} with the combined ELBO (JD) and separated ELBO (JD2), and GDA with prior knowledge of noise. The metric is the MSE loss on the $q,p$ testing trajectories. Here our method performs the best in the majority of cases, but the type of loss balancing that is optimal varies depending on the noise level.}
    \label{tab:noisy-loss-comp-cons}
\end{table}

\begin{table}[hb]
    \centering
    {
    \tiny
    \begin{tabular}{lcccccccc}
        \hline
        \textbf{Noise} & SSGP & Ours (Equal) & Ours (GDA) & Ours (MTAdam) & Ours (JD) & Ours (JD2) & Ours (Noise Prior) & Ours (Noise Prior 2) \\
        \hline
        0    & \textit{0.2419} $\pm$ 0.2798 & 0.2438 $\pm$ 0.2732 & 0.2471 $\pm$ 0.2814 & 0.2532 $\pm$ 0.1351 & 0.2629 $\pm$ 0.3136 & 0.2900 $\pm$ 0.4033 & 0.2847 $\pm$ 0.3471 & \textbf{0.0628} $\pm$ 0.1531 \\
        0.01 & \textbf{0.2377} $\pm$ 0.2776 & 0.2869 $\pm$ 0.4264 & \textit{0.2470} $\pm$ 0.2908 & 0.2419 $\pm$ 0.1236 & 0.2972 $\pm$ 0.4204 & 0.2865 $\pm$ 0.4033 & 0.3241 $\pm$ 0.6112 & \textbf{0.0474} $\pm$ 0.1132 \\
        0.05 & 0.2600 $\pm$ 0.3145 & 0.2816 $\pm$ 0.3729 & \textit{0.2565} $\pm$ 0.2940 & \textbf{0.2524} $\pm$ 0.1215 & 0.3127 $\pm$ 0.4624 & 0.2577 $\pm$ 0.3058 & 0.2603 $\pm$ 0.3035 & \textbf{0.1054} $\pm$ 0.1824 \\
        0.1  & 0.2530 $\pm$ 0.2869 & 0.2559 $\pm$ 0.2743 & \textit{0.2344} $\pm$ 0.2256 & 0.2502 $\pm$ 0.1280 & 0.2561 $\pm$ 0.2898 & 0.2445 $\pm$ 0.2661 & 0.2958 $\pm$ 0.5060 & \textbf{0.1857} $\pm$ 0.2483\\
        0.2  & 0.2924 $\pm$ 0.2902 & \textit{0.2739} $\pm$ 0.1703 & \textbf{0.2738} $\pm$ 0.1708 & 0.2921 $\pm$ 0.1371 & 0.2863 $\pm$ 0.2640 & 0.2951 $\pm$ 0.3025 & 0.2976 $\pm$ 0.2760 & 0.2906 $\pm$ 0.3187 \\
        \hline
    \end{tabular}
    }
    \caption{Comparison of various noise levels evaluated on the testing dataset for the chaotic duffing system. We test with uncorrupted data and datasets with Gaussian noise of standard deviation $0.01, 0.05, 0.1,$ and $0.2$. We compare the base SSGP model \cite{tanaka2022ssgp} to our method with various ways of weighting the multi-term loss function, which are equal summation, gradient descent-ascent (GDA), multi-term Adam (MTAdam) \cite{malkiel2020mtadam}, Jacobian descent with a UPGrad aggregator \cite{quinton2024jacobiandescent} with the combined ELBO (JD) and separated ELBO (JD2), and GDA with prior knowledge of noise (Noise Prior) and its modification learning only the mean path (Noise Prior 2). The metric is the MSE loss on the $q,p$ testing trajectories. Here the method using our multi-term loss function with the a prior noise level, typically performs the best, most likely because it accurately separates noise from the true dynamics while accounting for physical constraints.}
    \label{tab:noisy-loss-comp-port}
\end{table}

\begin{table}[ht]
    \centering  
    \begin{tabular}{lccccccccc}
    \toprule
    \textbf{Dataset} & \textbf{equal} & \textbf{equal\_E} & \textbf{equal\_L} & \textbf{equal\_V} & \textbf{gda} & \textbf{gda\_E} & \textbf{gda\_L} & \textbf{gda\_V} & \textbf{noreg} \\
    \midrule
    Single Pendulum  & 0.1111 & \textbf{0.0941} & \textbf{0.0941} & 0.1315 & 0.1634 & 0.5146 & 0.2812 & 0.1315 & 1.0222 \\
    Damped Pendulum  & 0.0714 & 0.1622 & 0.1629 & 0.0671 & \textbf{0.0613} & 0.1004 & 0.1347 & 0.0671 & 0.1733 \\
    Windy Pendulum & 0.0670 & 0.3019 & 0.1113 & 0.4420 & 0.2972 & \textbf{0.0303} & 0.0829 & 0.4420 & 0.0493 \\
    \bottomrule
    \end{tabular}
    \caption{Mean squared error (MSE) over test sets for single (top row), damped (middle row), and windy (bottom row) pendulum systems. We compare the use of each loss term individually and the combination of all terms, using either equal and GDA-balanced weighting, against unregularized ELBO. Bold indicates the lowest MSE in each row. E denotes only the energy regularizer is used, while V and L denote the volume and Lyapunov regularizers respectively. }
    \label{tab:ablation}
\end{table}
\section{Experiments}
\label{sec:experiments}

\subsection{Experiment Setups}
We evaluate on benchmarks which include 3 classes of Hamiltonian systems: standard conservative systems, dissipative systems, and port-Hamiltonian systems, which have both dissipation and external forcing. We test our model on systems of each type. By default we generate 100 noisy trajectories as training data. Each trajectory contains $100$ time stamp samples over domain $[0,10]$. The batch size is assumed to be $100$. The learning rate is set to be $10^{-3}$ throughout the experiments. The testing performance is reported over $100$ trajectories generated via the same scheme as training data. This ensures the initial conditions are sampled from same prior (but with different phase dynamics) and the trajectories are noisy. Additional details of the configurations of the datasets are given in Table \ref{tab:data-config}. Experiments were performed on NVIDIA Grace CPU with 116 GB RAM and NVIDIA H200 GPU with 96 GB VRAM, using Linux operating system and PyTorch \cite{paszke2019pytorch} machine learning library. We use torchdiffeq \cite{chen2018neuralode} library for the numerical integration function $\mathrm{odeint}$ in Algorithm \ref{alg:main}.

\begin{figure}[!htbp]
    \centering
    \captionsetup[subfigure]{labelformat=empty}
    \begin{subfigure}[t]{0.19\linewidth}
        \centering
        \includegraphics[width=\linewidth]{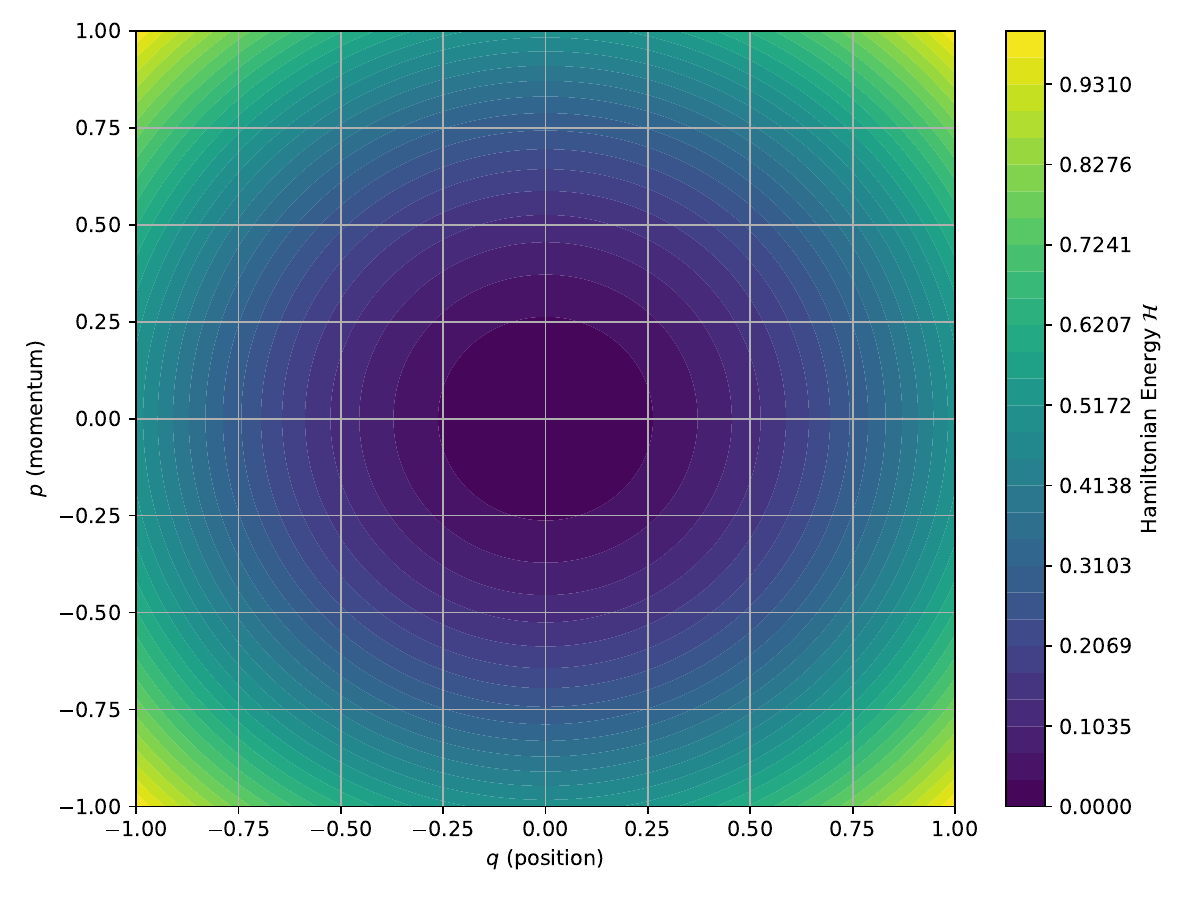}
        \caption{Ground truth}
    \end{subfigure}
    \begin{subfigure}[t]{0.19\linewidth}
        \centering
        \includegraphics[width=\linewidth]{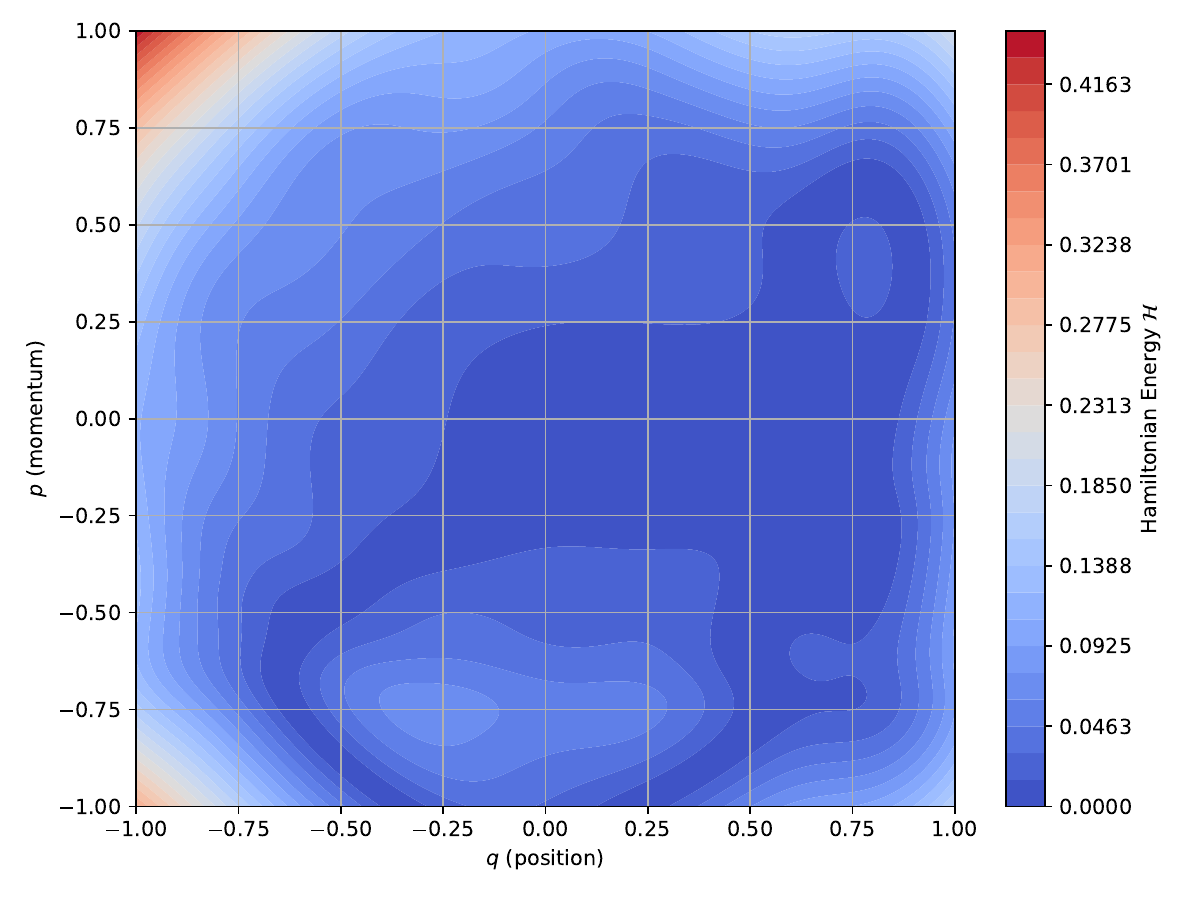}
        \caption{DHNN}
    \end{subfigure}
    \begin{subfigure}[t]{0.19\linewidth}
        \centering
        \includegraphics[width=\linewidth]{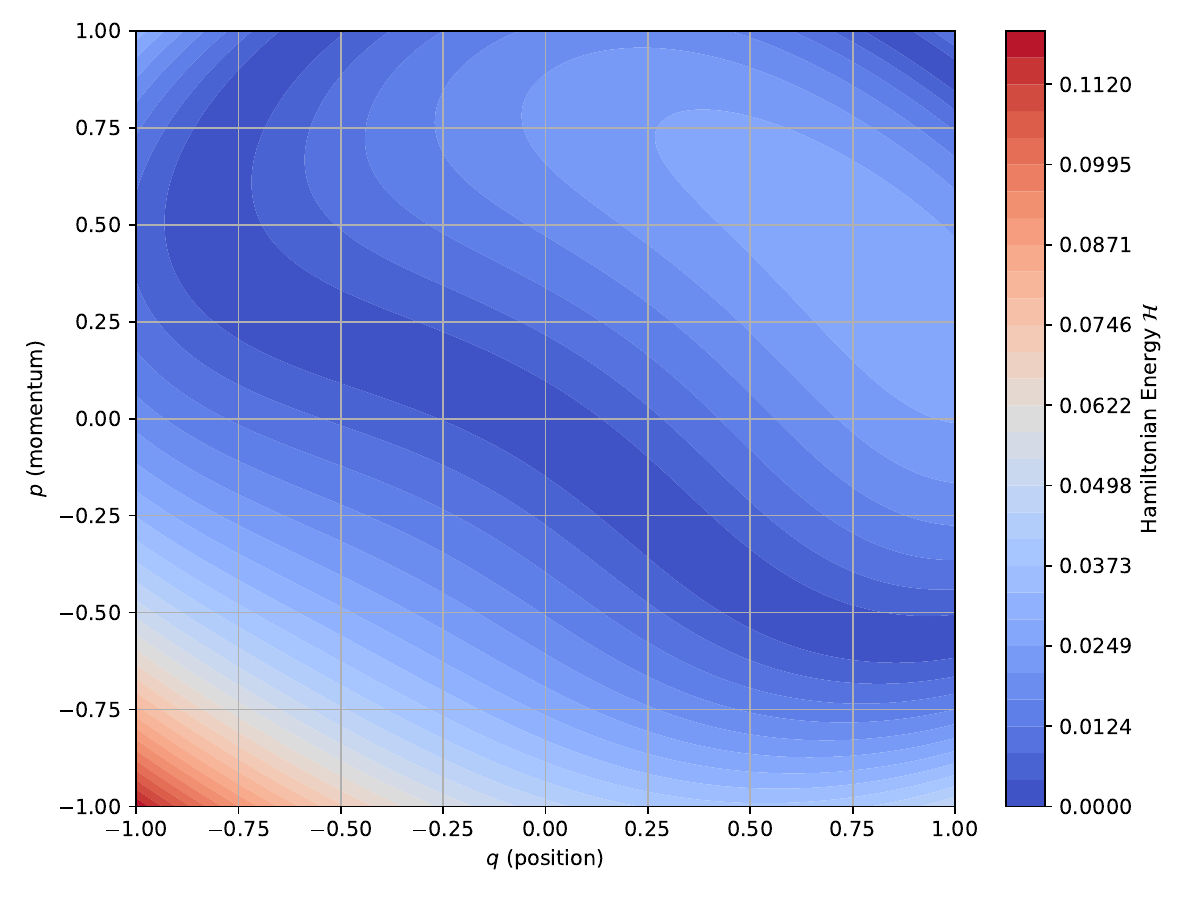}
        \caption{SSGP}
    \end{subfigure}
    \begin{subfigure}[t]{0.19\linewidth}
        \centering
        \includegraphics[width=\linewidth]{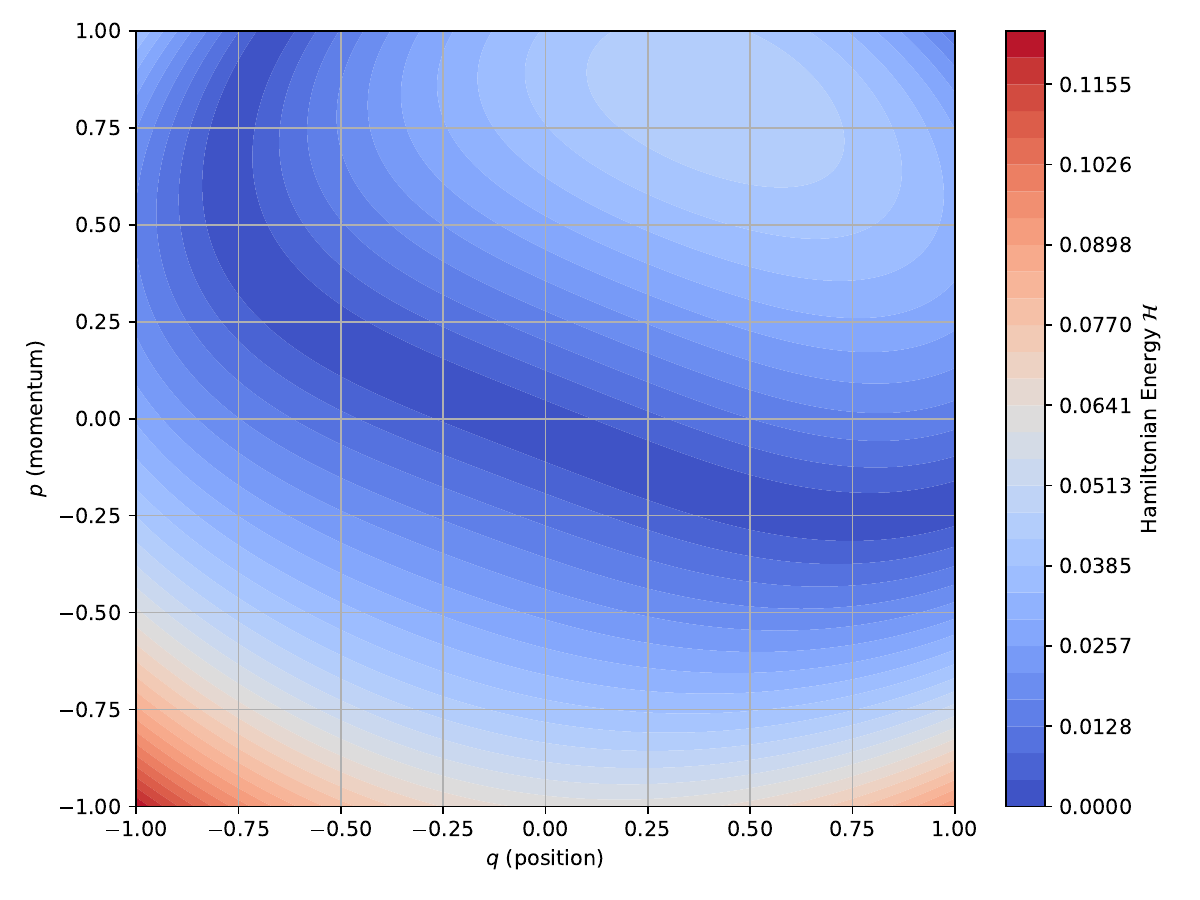}
        \caption{Ours (equal)}
    \end{subfigure}
    \begin{subfigure}[t]{0.19\linewidth}
        \centering
        \includegraphics[width=\linewidth]{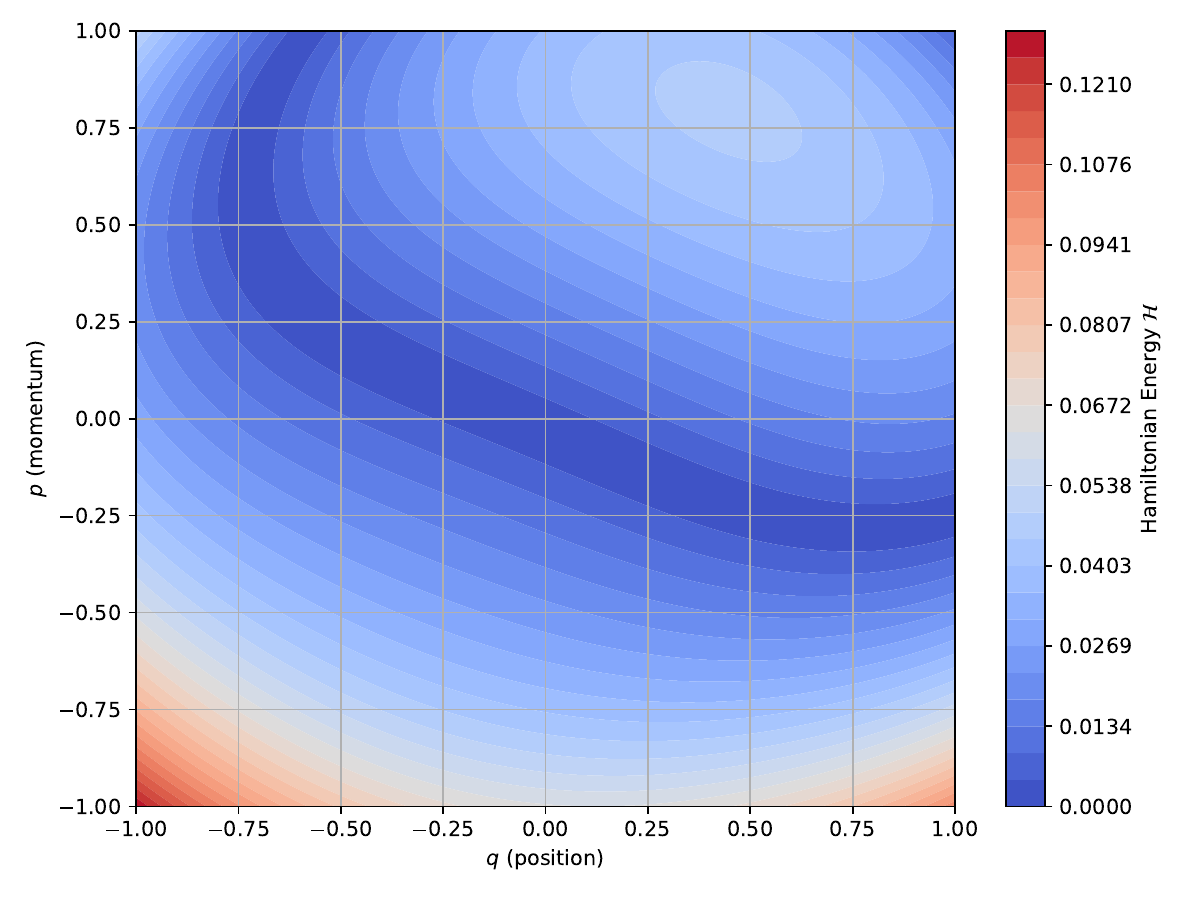}
        \caption{Ours (GDA)}
    \end{subfigure}
    \begin{subfigure}[t]{0.19\linewidth}
        \centering
        \includegraphics[width=\linewidth]{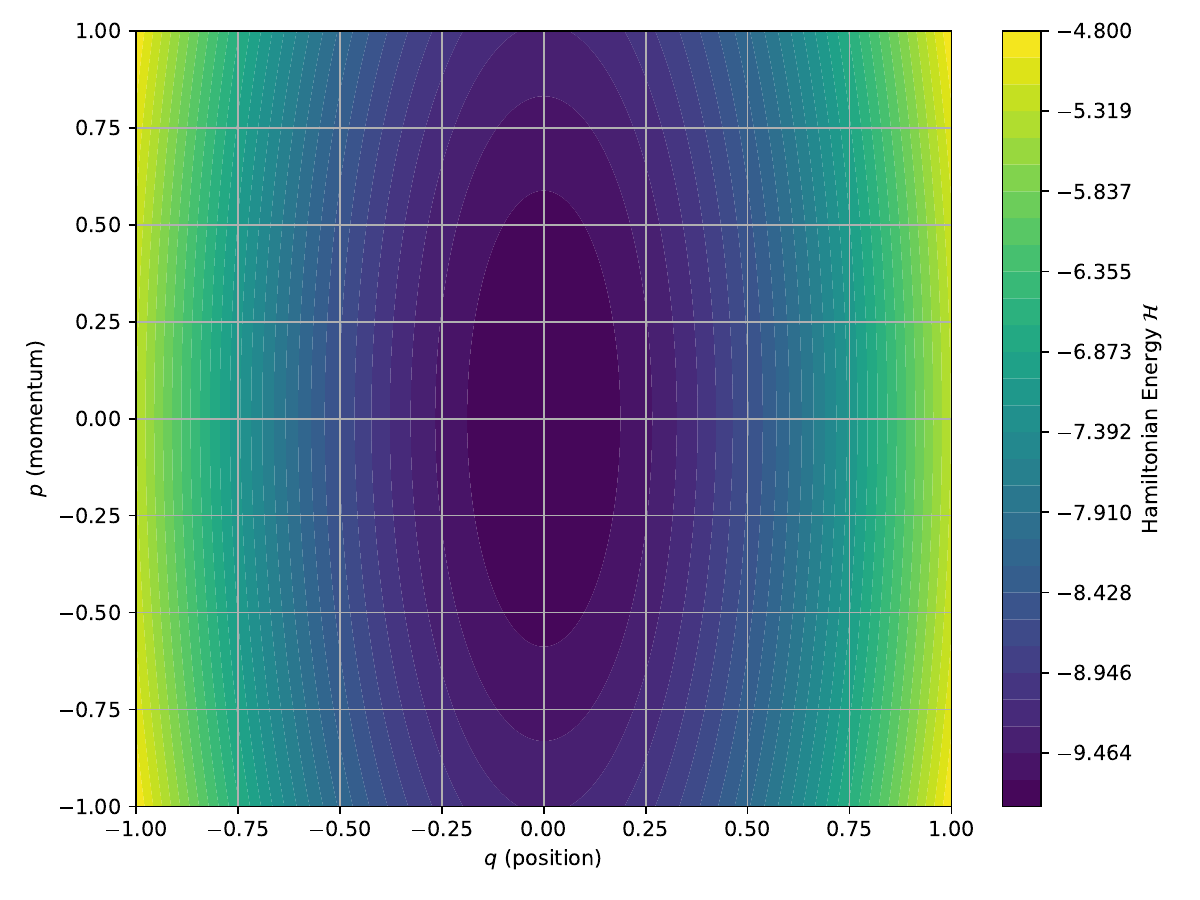}
        \caption{Ground truth}
    \end{subfigure}
    \begin{subfigure}[t]{0.19\linewidth}
        \centering
        \includegraphics[width=\linewidth]{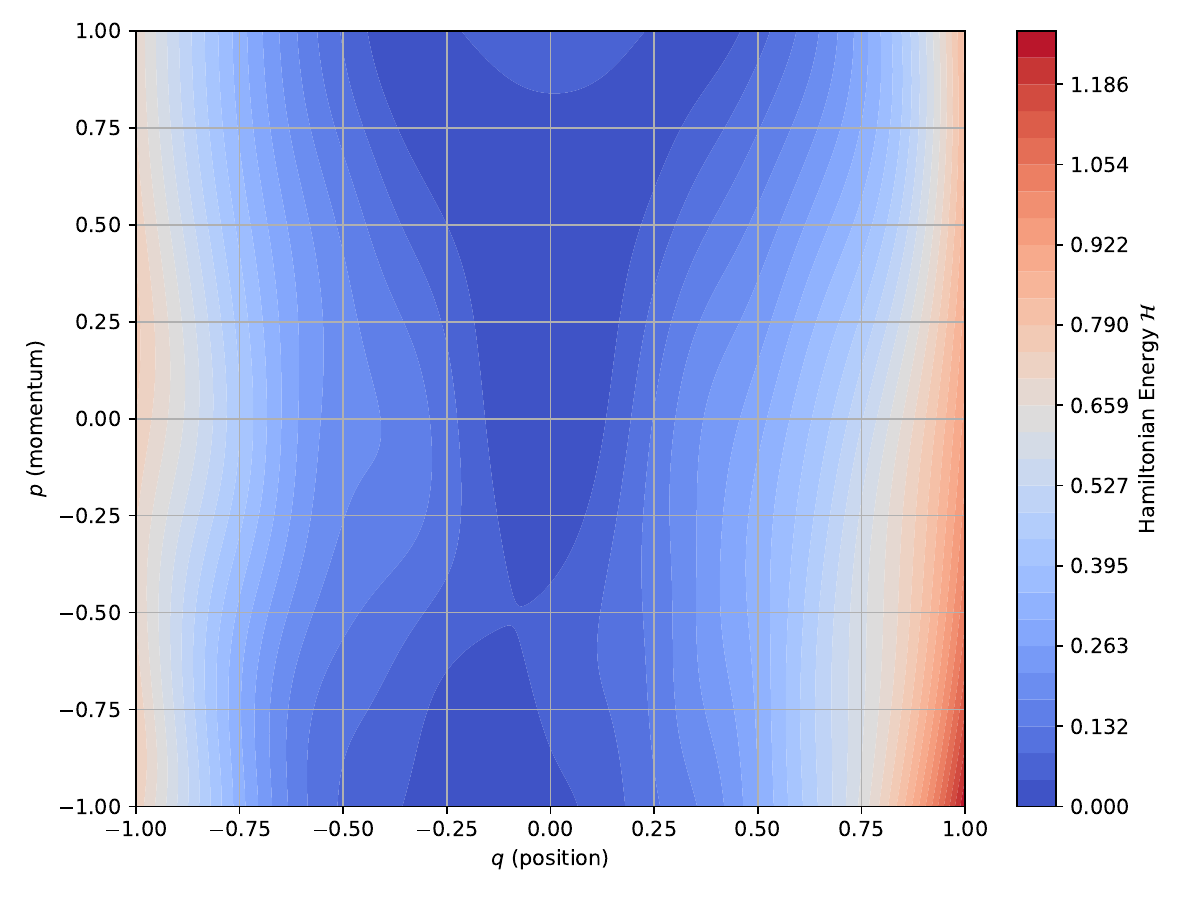}
        \caption{DHNN}
    \end{subfigure}
    \begin{subfigure}[t]{0.19\linewidth}
        \centering
        \includegraphics[width=\linewidth]{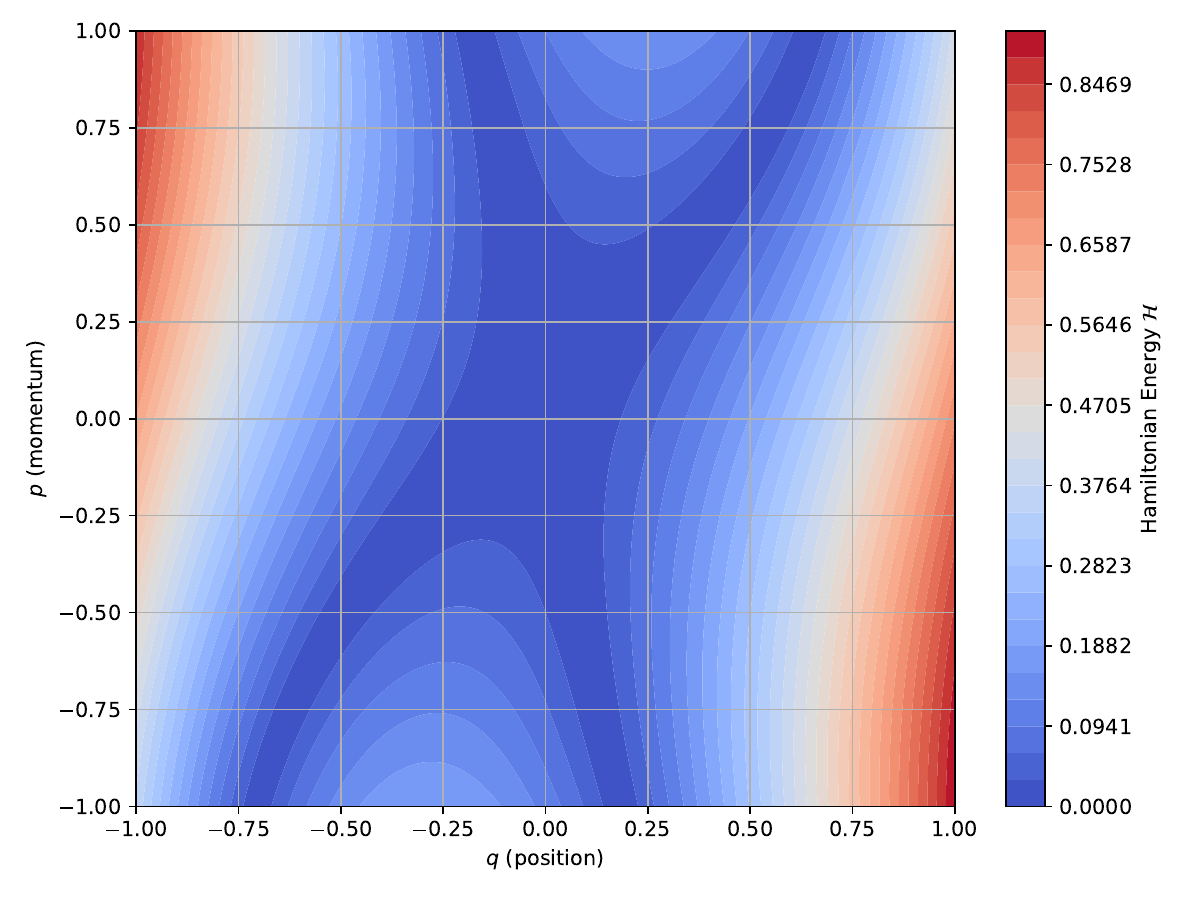}
        \caption{SSGP}
    \end{subfigure}
    \begin{subfigure}[t]{0.19\linewidth}
        \centering
        \includegraphics[width=\linewidth]{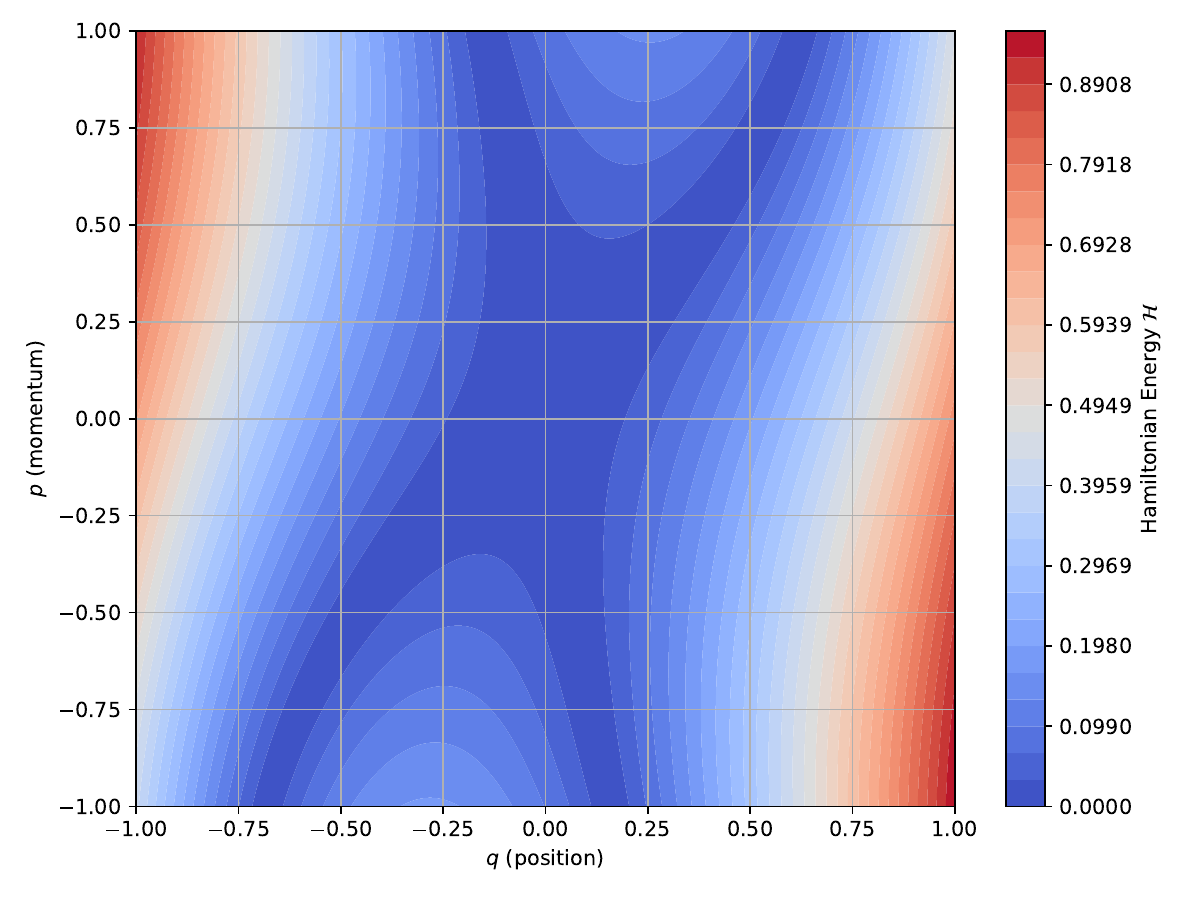}
        \caption{Ours (equal)}
    \end{subfigure}
    \begin{subfigure}[t]{0.19\linewidth}
        \centering
        \includegraphics[width=\linewidth]{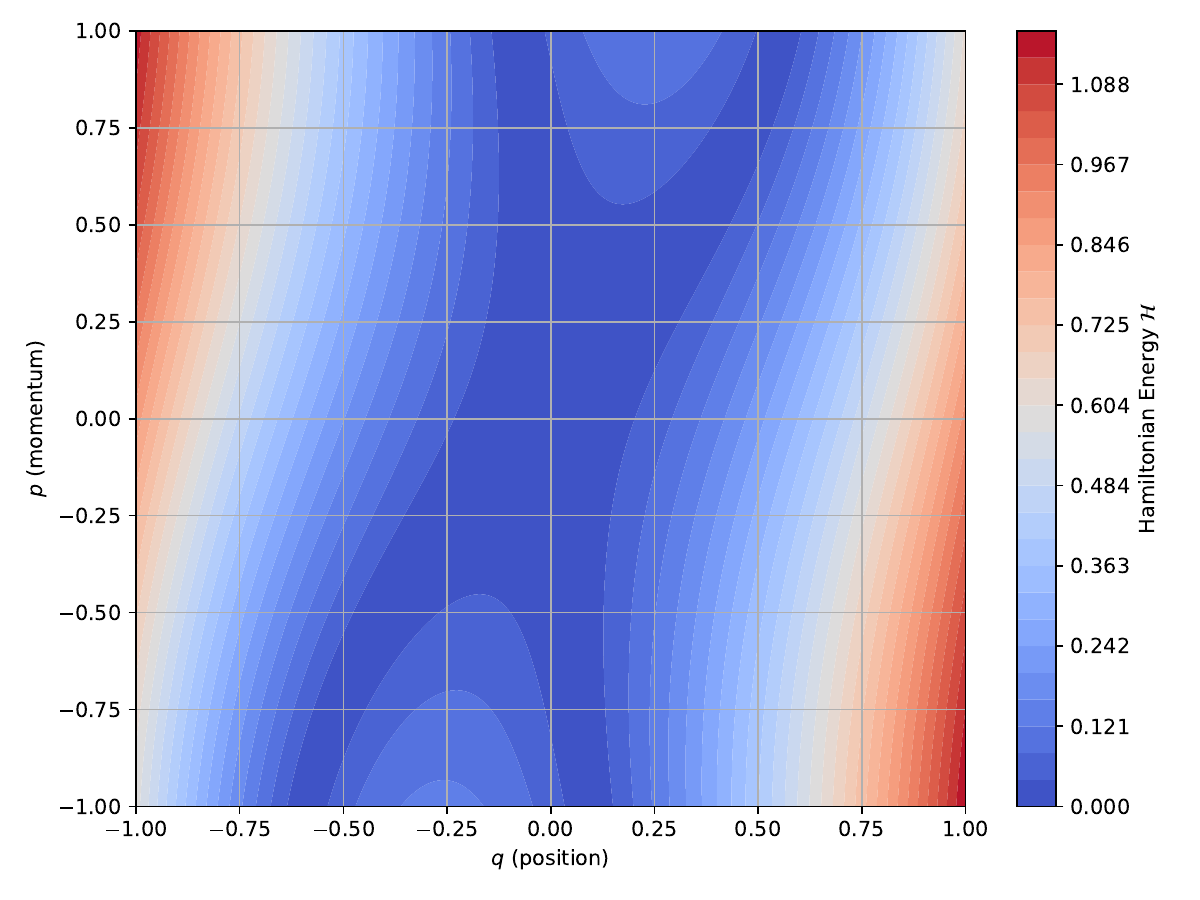}
        \caption{Ours (GDA)}
    \end{subfigure}
    \begin{subfigure}[t]{0.19\linewidth}
        \centering
        \includegraphics[width=\linewidth]{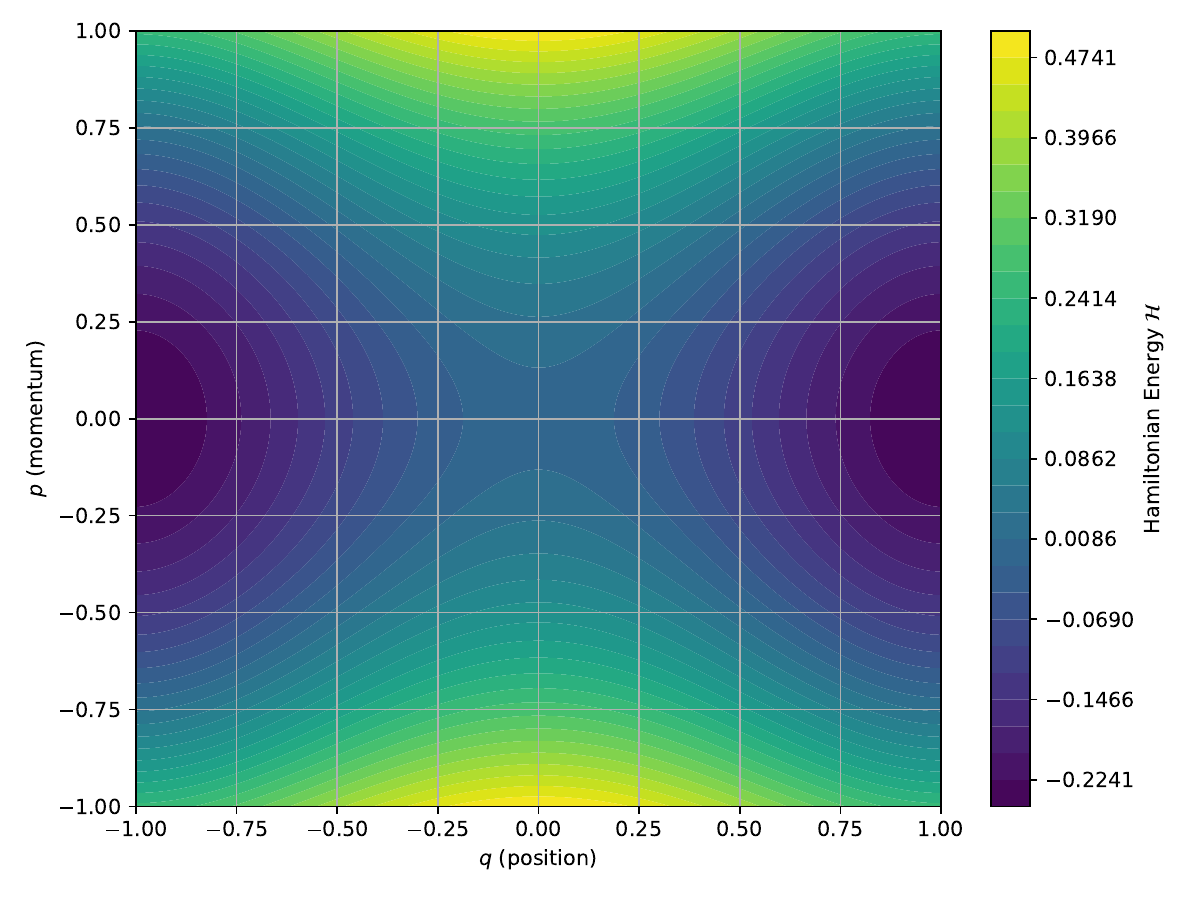}
        \caption{Ground truth}
    \end{subfigure}
    \begin{subfigure}[t]{0.19\linewidth}
        \centering
        \includegraphics[width=\linewidth]{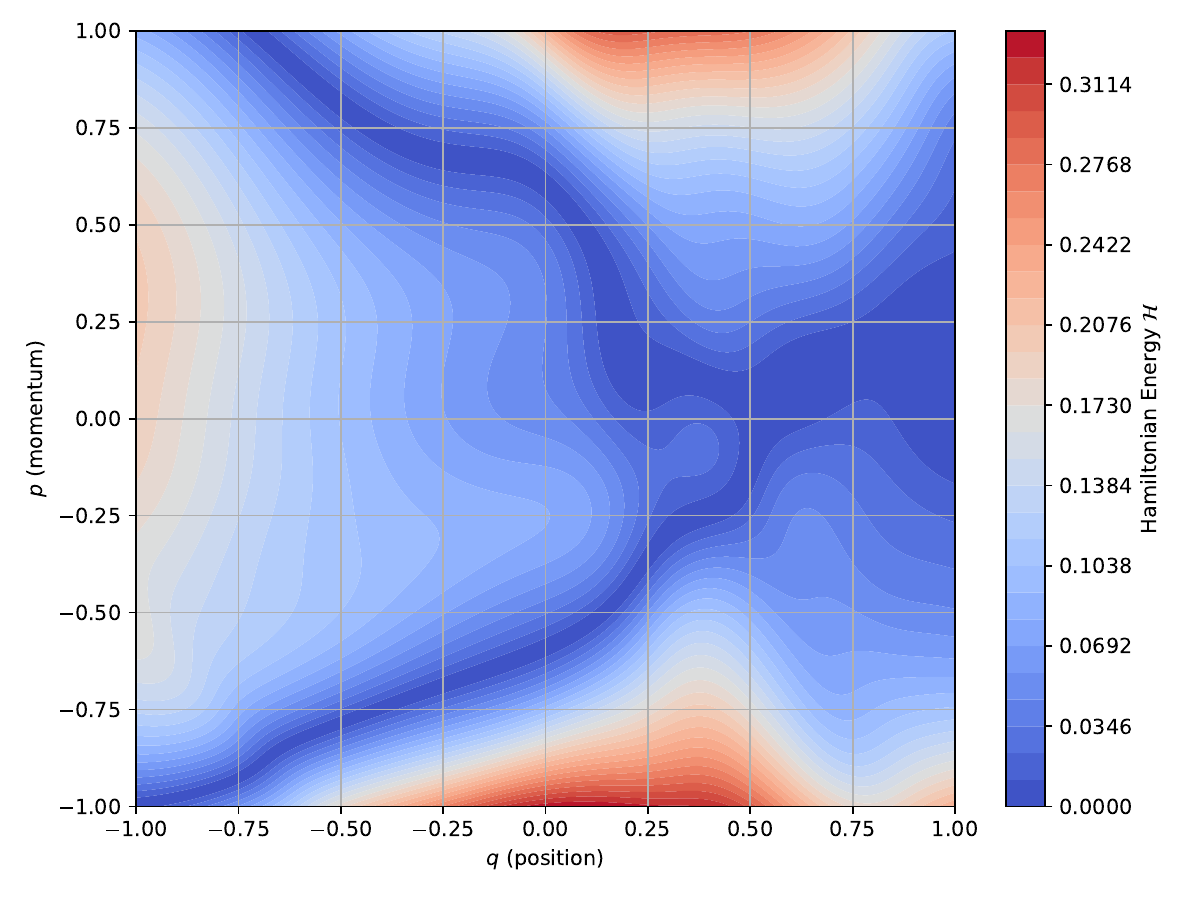}
        \caption{DHNN}
    \end{subfigure}
    \begin{subfigure}[t]{0.19\linewidth}
        \centering
        \includegraphics[width=\linewidth]{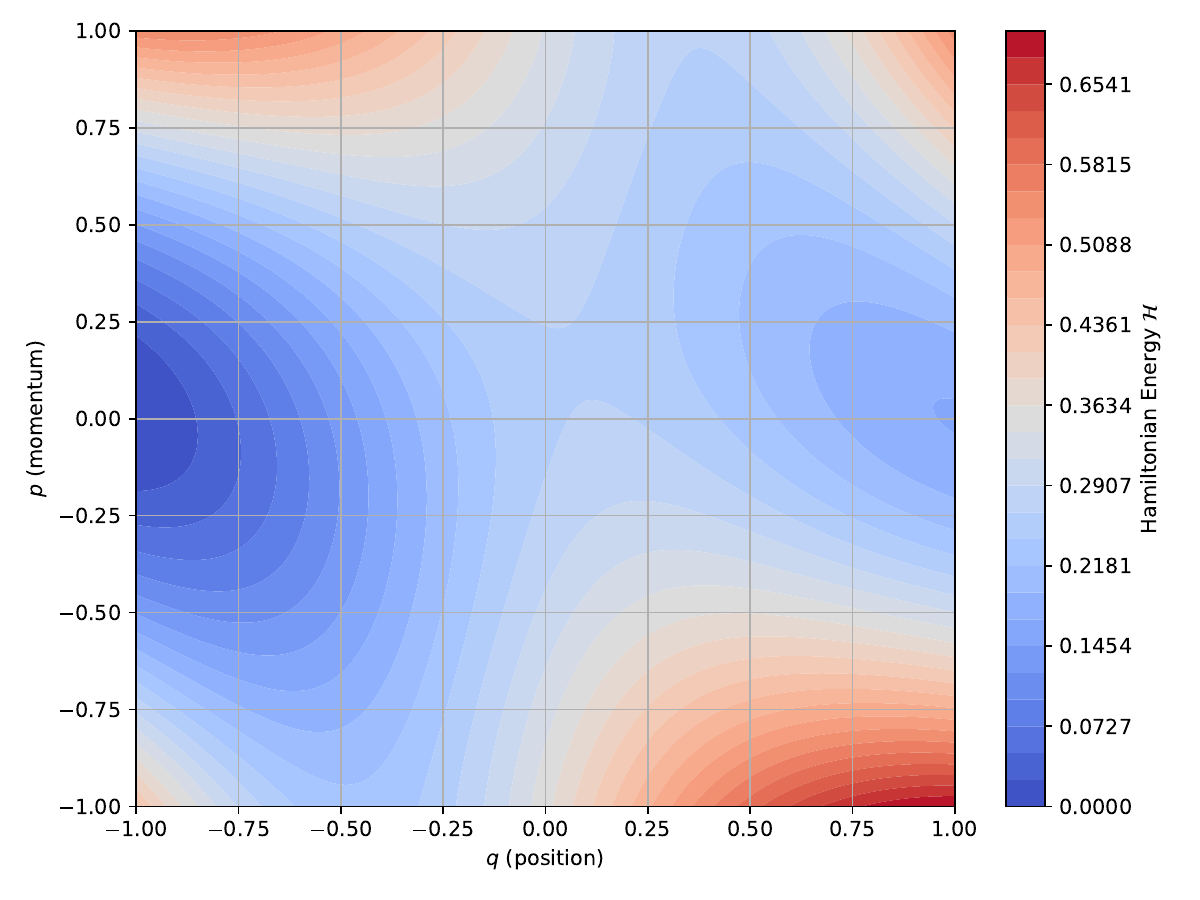}
        \caption{SSGP}
    \end{subfigure}
    \begin{subfigure}[t]{0.19\linewidth}
        \centering
        \includegraphics[width=\linewidth]{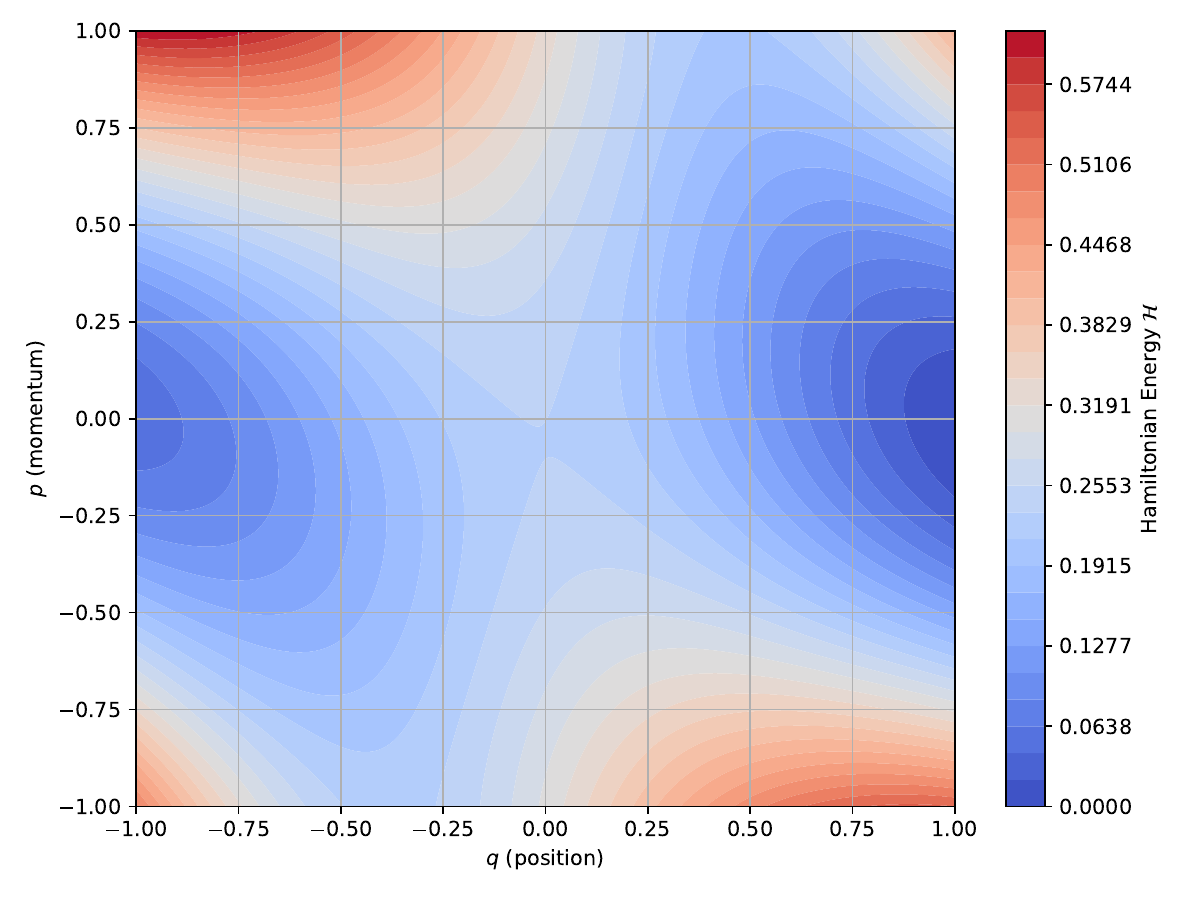}
        \caption{Ours (equal)}
    \end{subfigure}
    \begin{subfigure}[t]{0.19\linewidth}
        \centering
        \includegraphics[width=\linewidth]{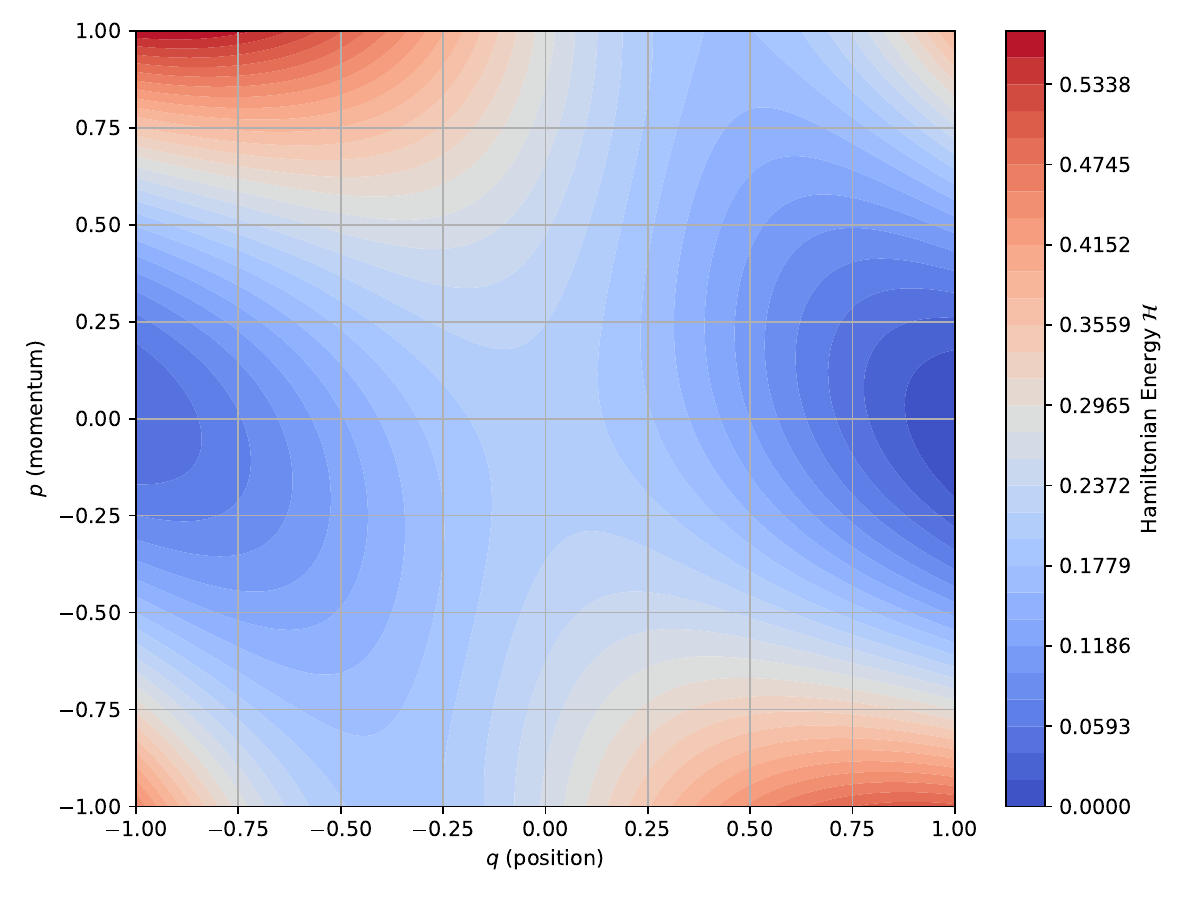}
        \caption{Ours (GDA)}
    \end{subfigure}
    \caption{The 1st column is the true phase maps with Hamiltonian energy contours three different dissipative Hamiltonian systems: damped spring (top row), damped pendulum (middle row), and unforced Duffing (bottom row). Columns 2 through 5 shows the error between the true Hamiltonian energy and learned Hamiltonian energy for both prior work and our methods. The prior work is shown in column 2 with DHNN \cite{sosanya2022dhnn} and column 3 with SSGP \cite{tanaka2022ssgp}. Our method is shown in column 4, using equally weighted loss terms, and column 5, using GDA-balanced loss terms.}
    \label{fig:phase-map-comp}
\end{figure}

\begin{figure}[!htbp]
    \centering
    \captionsetup[subfigure]{labelformat=empty}
    \begin{subfigure}[t]{0.19\linewidth}
        \centering
        \includegraphics[width=\linewidth]{img/phase_plot_comps/damped_spring_true.pdf}
        \caption{Ground truth}
    \end{subfigure}
    \begin{subfigure}[t]{0.19\linewidth}
        \centering
        \includegraphics[width=\linewidth]{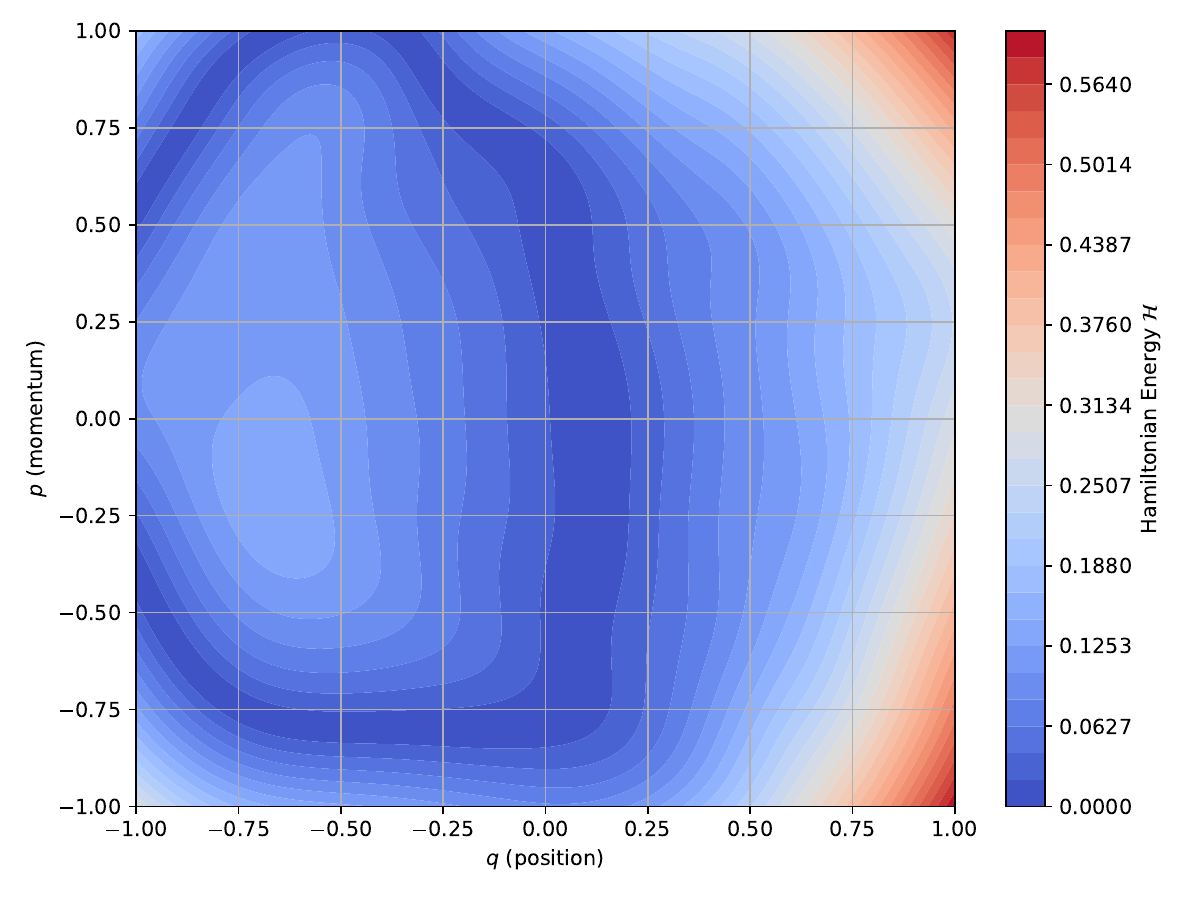}
        \caption{DHNN}
    \end{subfigure}
    \begin{subfigure}[t]{0.19\linewidth}
        \centering
        \includegraphics[width=\linewidth]{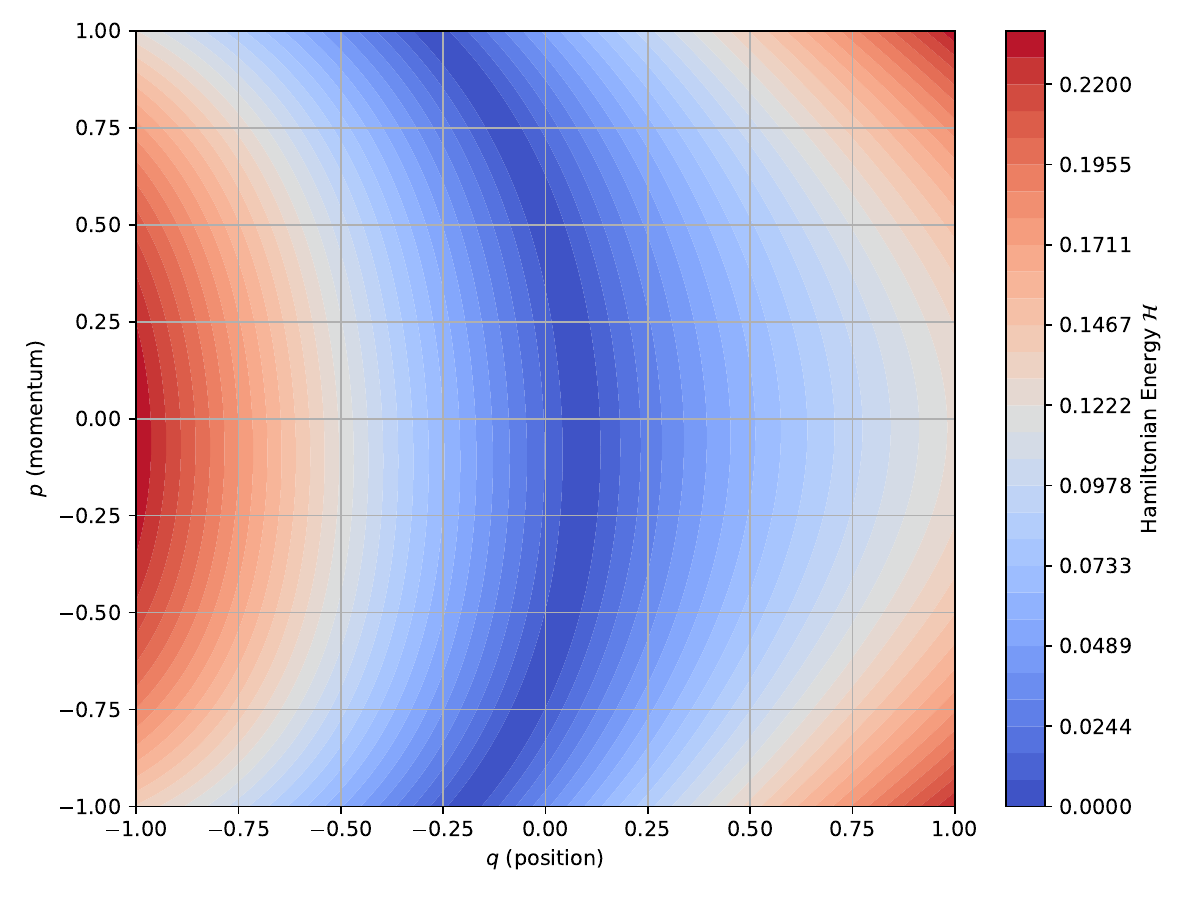}
        \caption{SSGP}
    \end{subfigure}
    \begin{subfigure}[t]{0.19\linewidth}
        \centering
        \includegraphics[width=\linewidth]{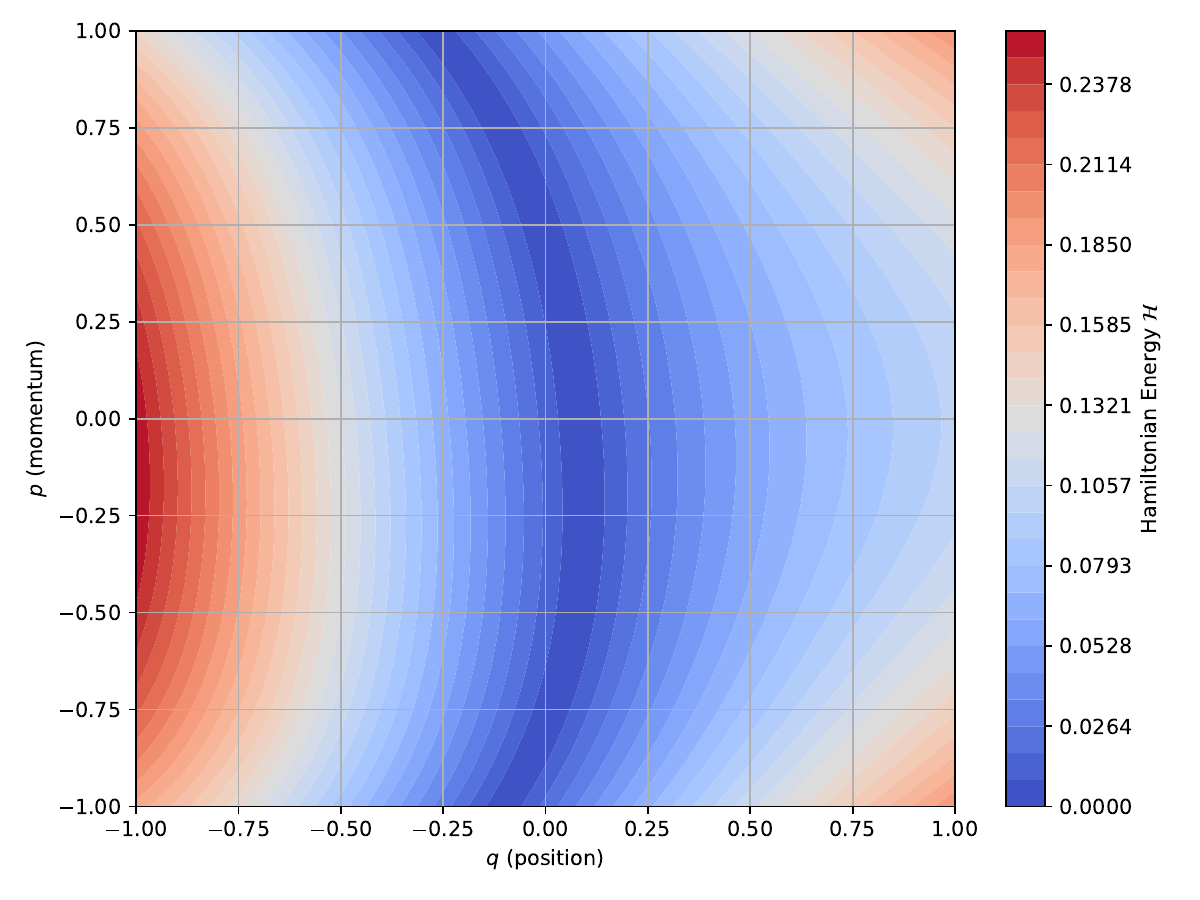}
        \caption{Ours (equal)}
    \end{subfigure}
    \begin{subfigure}[t]{0.19\linewidth}
        \centering
        \includegraphics[width=\linewidth]{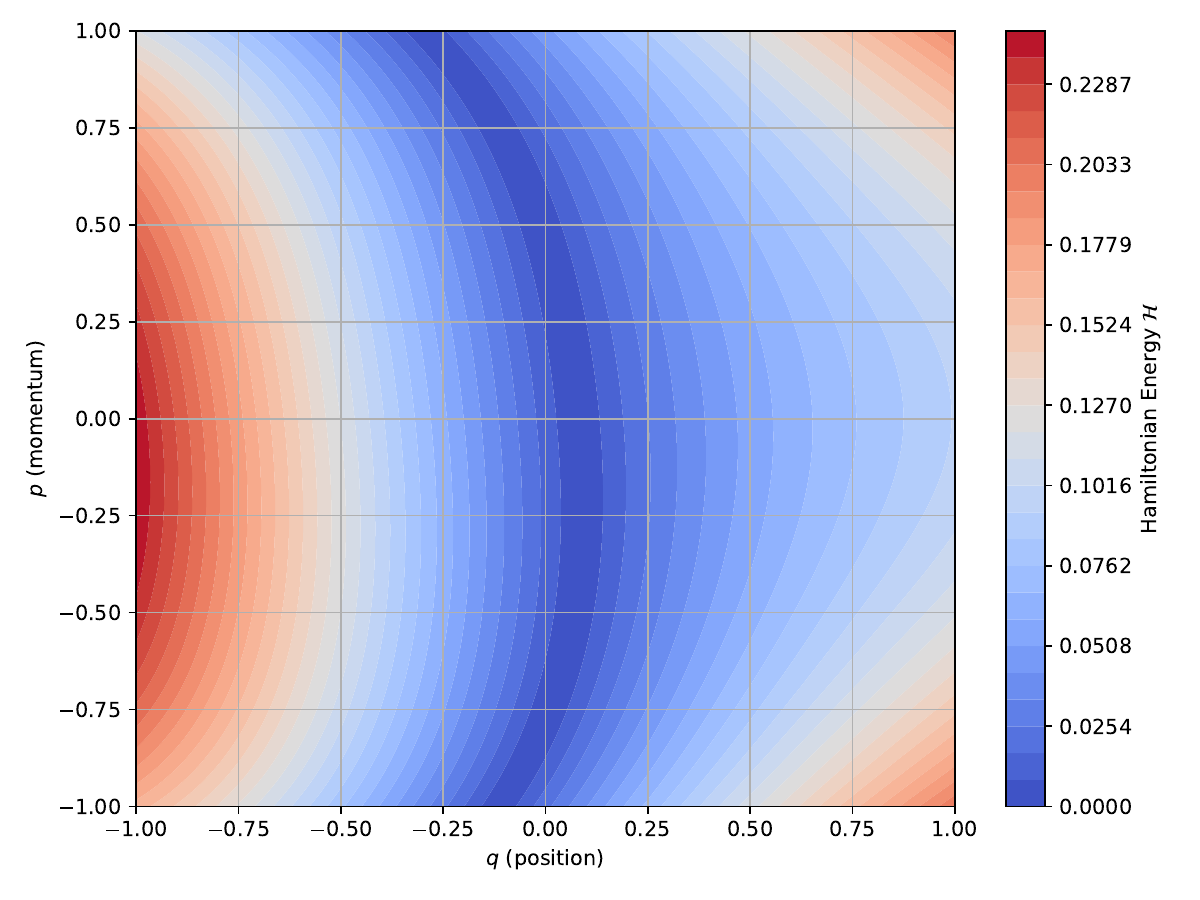}
        \caption{Ours (GDA)}
    \end{subfigure}
    \begin{subfigure}[t]{0.19\linewidth}
        \centering
        \includegraphics[width=\linewidth]{img/phase_plot_comps/damped_pendulum_true.pdf}
        \caption{Ground truth}
    \end{subfigure}
    \begin{subfigure}[t]{0.19\linewidth}
        \centering
        \includegraphics[width=\linewidth]{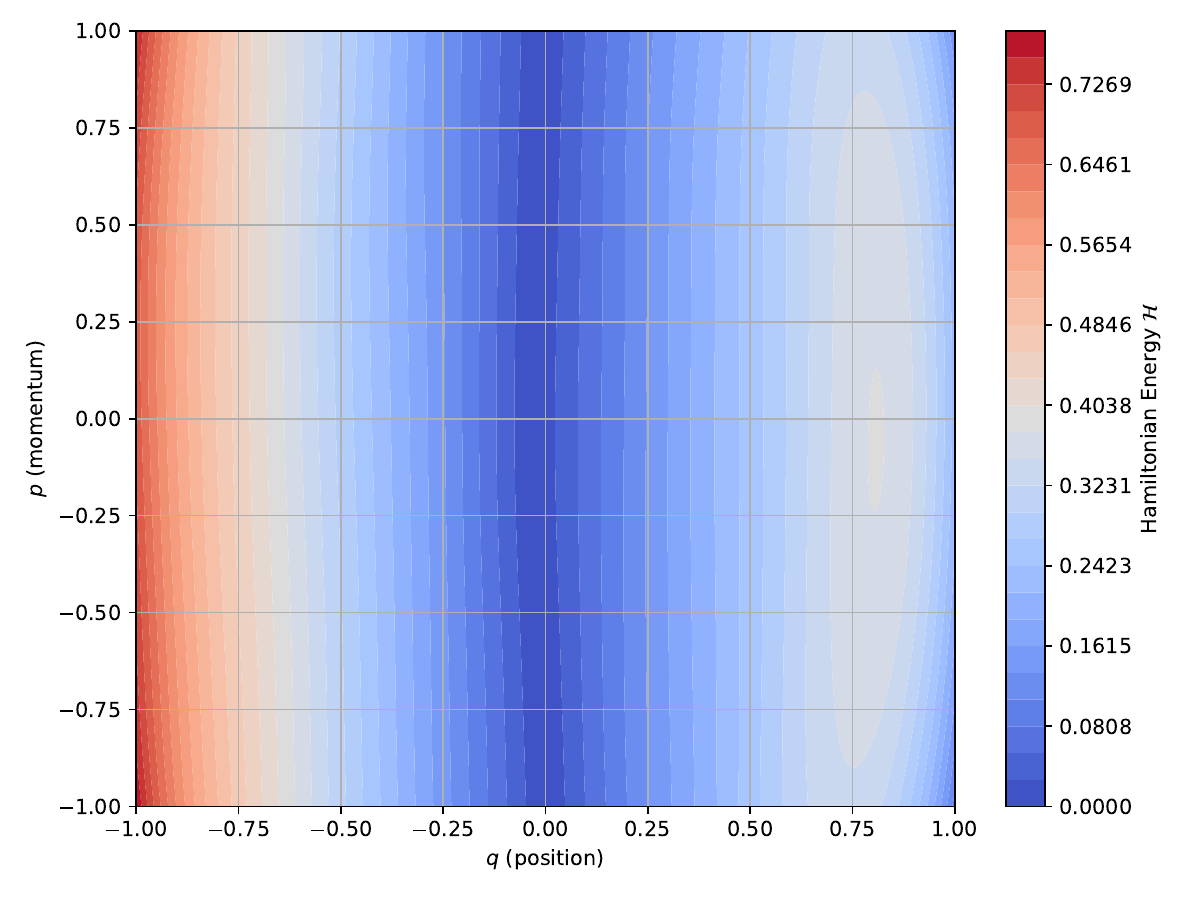}
        \caption{DHNN}
    \end{subfigure}
    \begin{subfigure}[t]{0.19\linewidth}
        \centering
        \includegraphics[width=\linewidth]{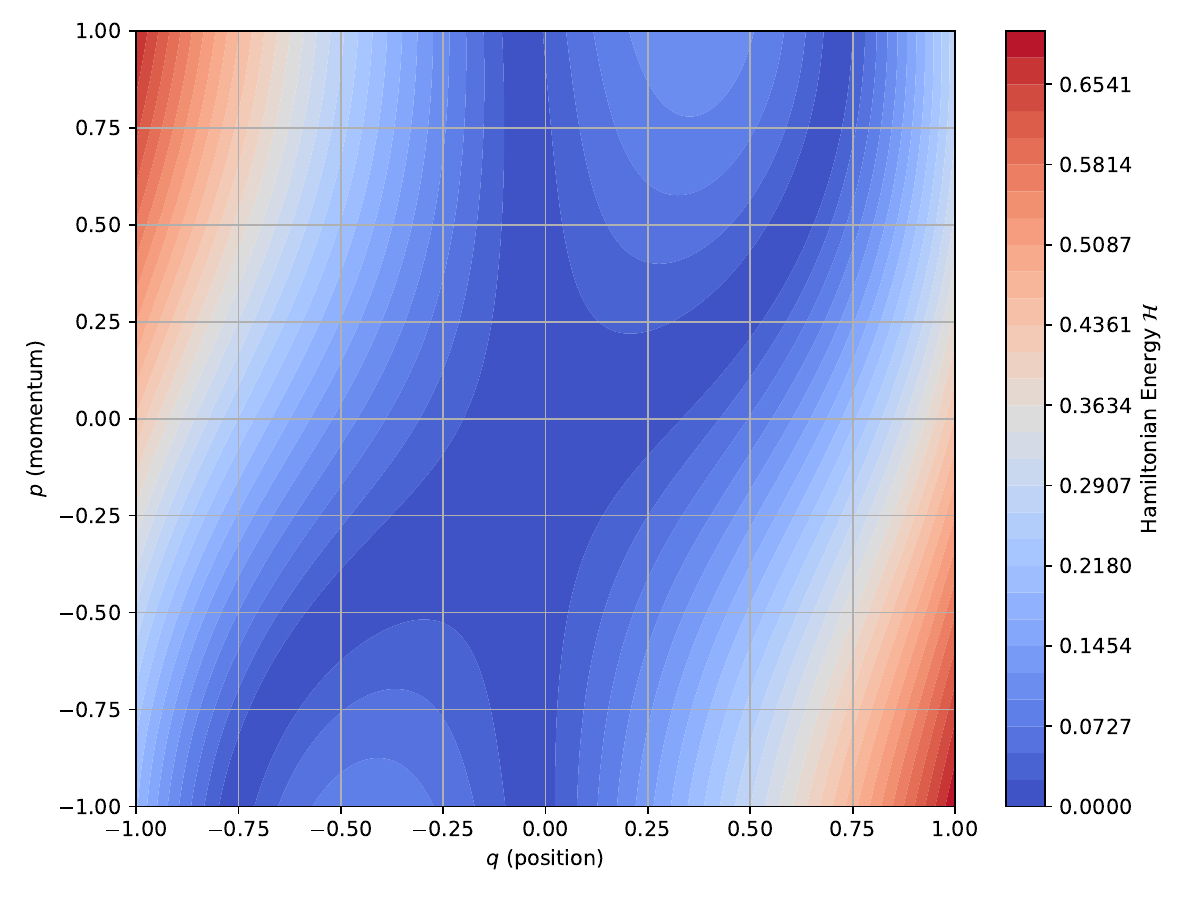}
        \caption{SSGP}
    \end{subfigure}
    \begin{subfigure}[t]{0.19\linewidth}
        \centering
        \includegraphics[width=\linewidth]{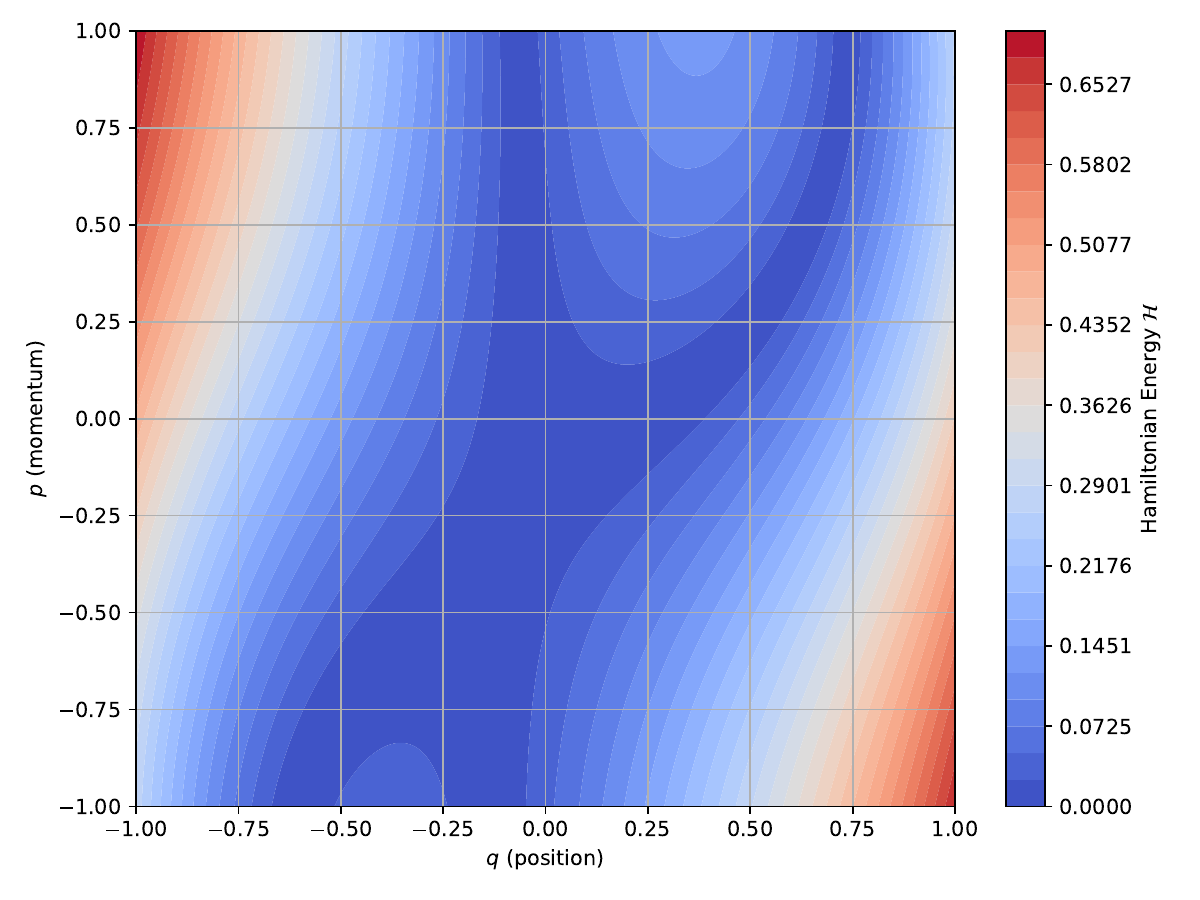}
        \caption{Ours (equal)}
    \end{subfigure}
    \begin{subfigure}[t]{0.19\linewidth}
        \centering
        \includegraphics[width=\linewidth]{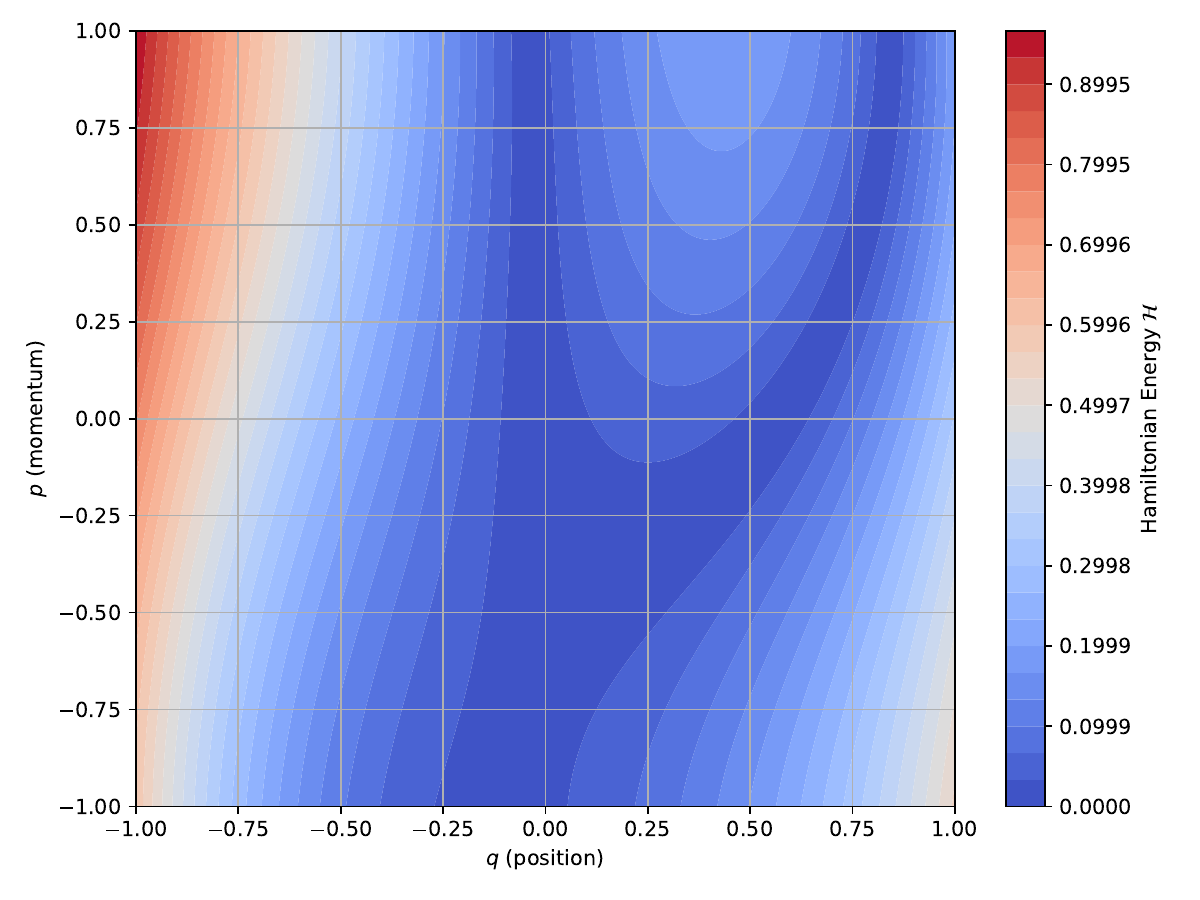}
        \caption{Ours (GDA)}
    \end{subfigure}
    \begin{subfigure}[t]{0.19\linewidth}
        \centering
        \includegraphics[width=\linewidth]{img/phase_plot_comps/unforced_duffing_true.pdf}
        \caption{Ground truth}
    \end{subfigure}
    \begin{subfigure}[t]{0.19\linewidth}
        \centering
        \includegraphics[width=\linewidth]{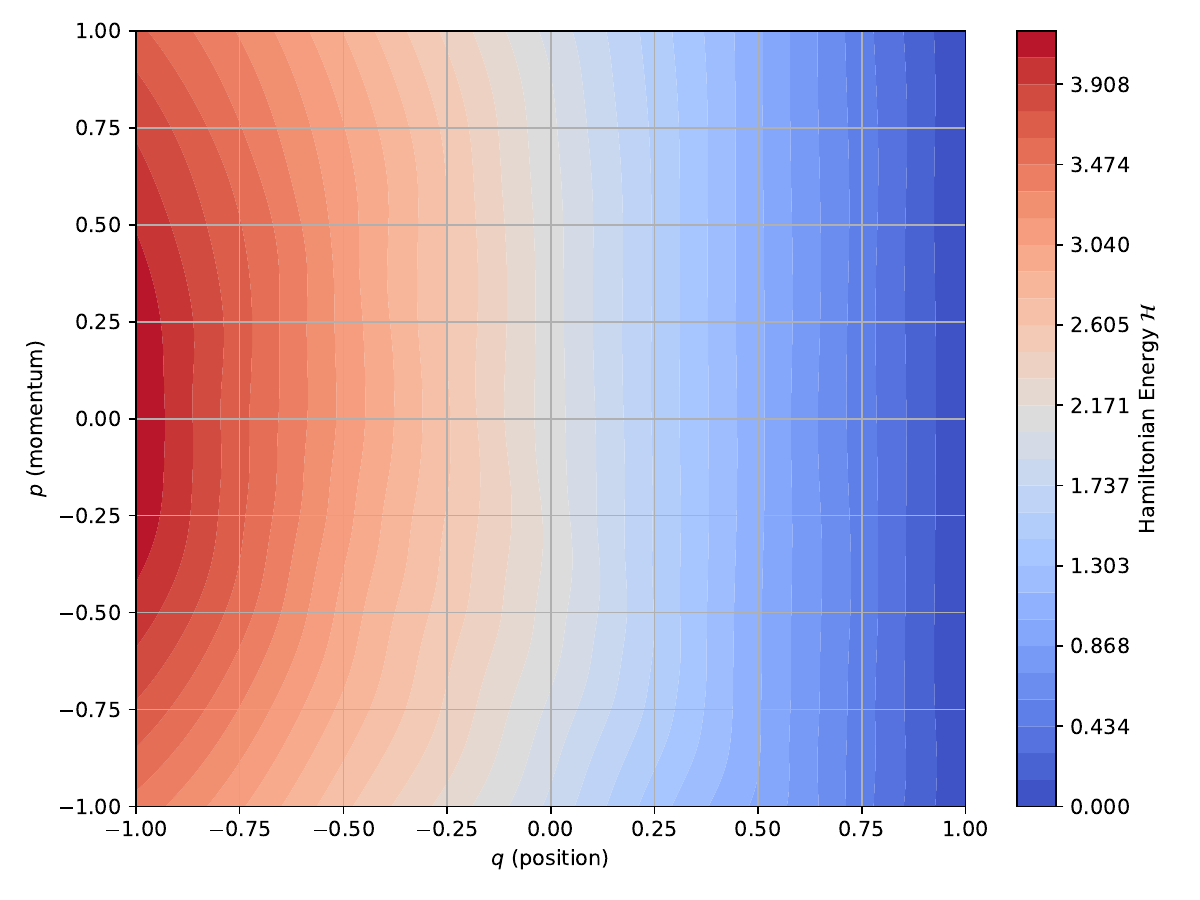}
        \caption{DHNN}
    \end{subfigure}
    \begin{subfigure}[t]{0.19\linewidth}
        \centering
        \includegraphics[width=\linewidth]{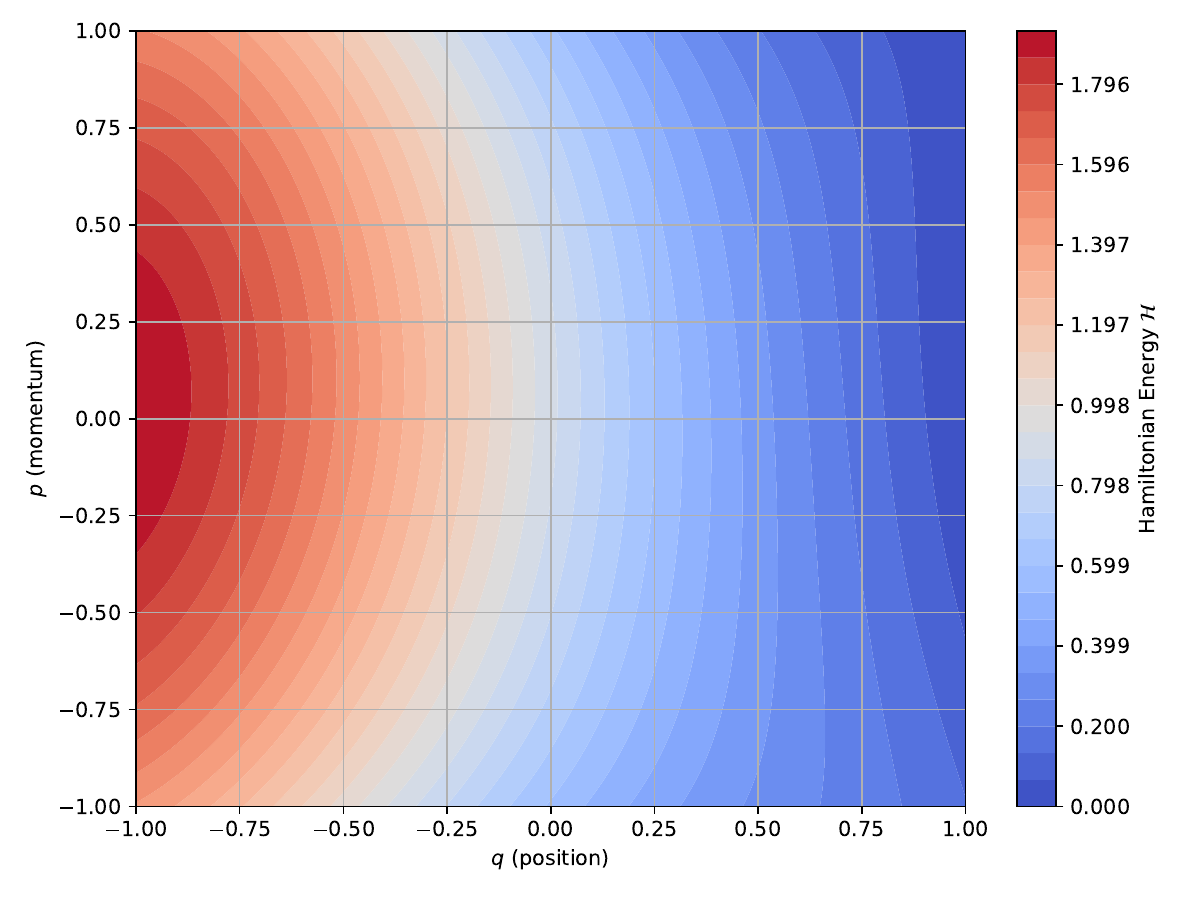}
        \caption{SSGP}
    \end{subfigure}
    \begin{subfigure}[t]{0.19\linewidth}
        \centering
        \includegraphics[width=\linewidth]{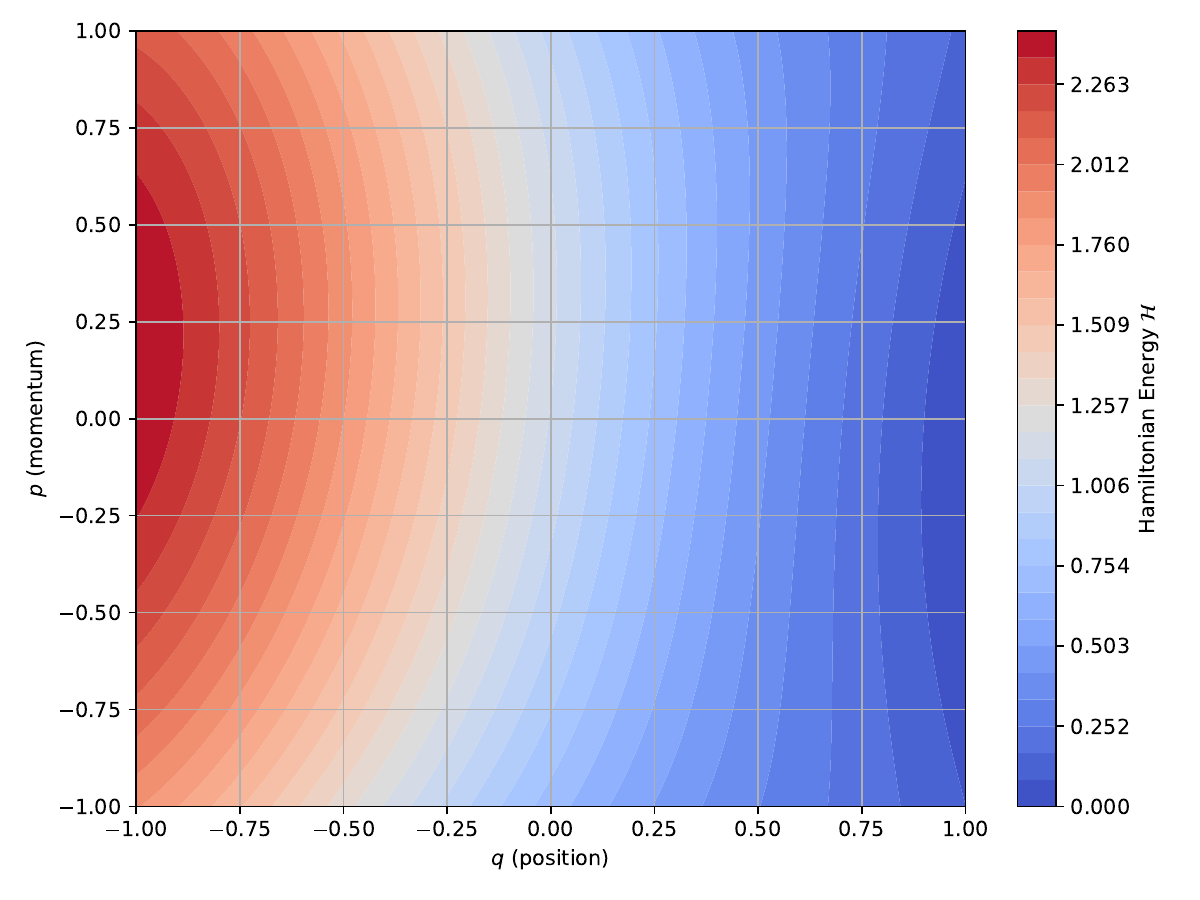}
        \caption{Ours (equal)}
    \end{subfigure}
    \begin{subfigure}[t]{0.19\linewidth}
        \centering
        \includegraphics[width=\linewidth]{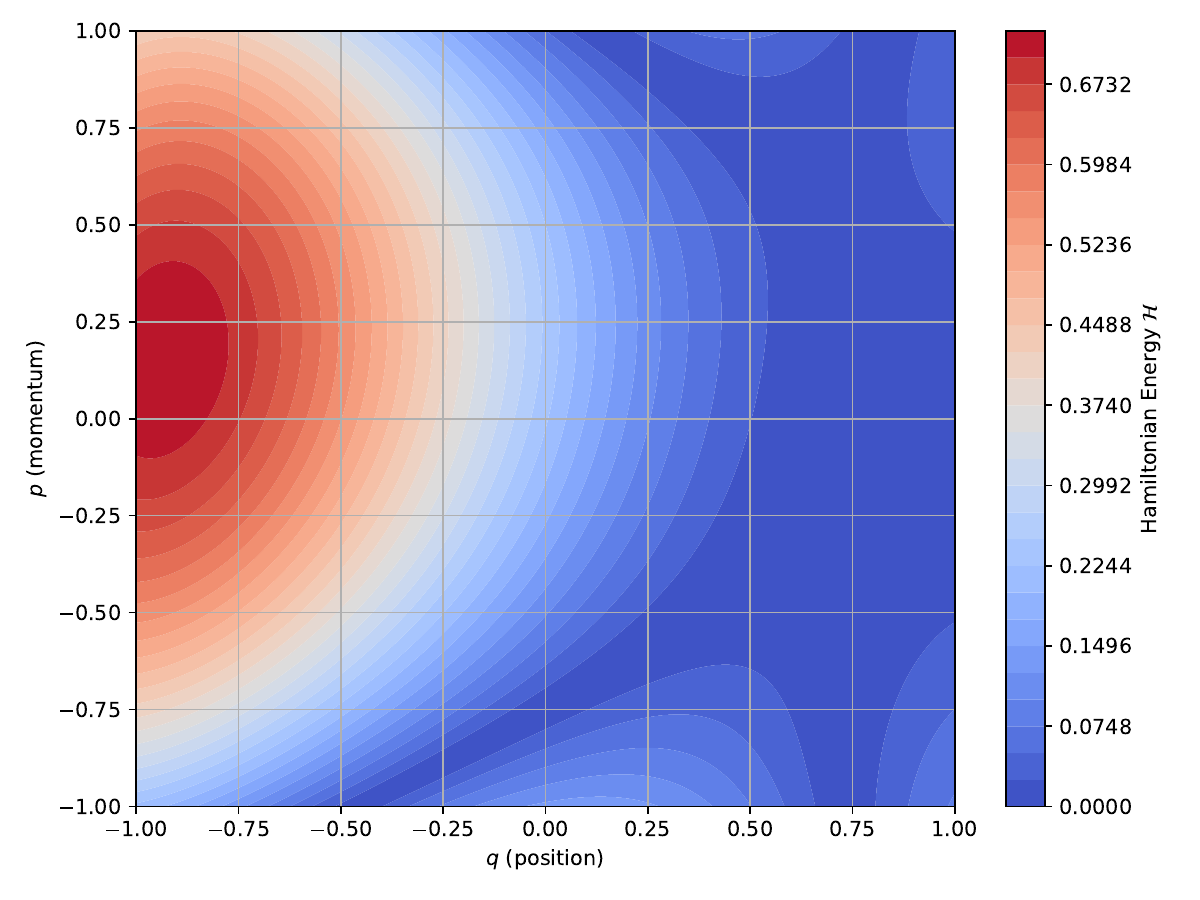}
        \caption{Ours (GDA)}
    \end{subfigure}
    \caption{The 1st column is the true phase maps with Hamiltonian energy contours three different port-Hamiltonian systems: forced spring (top row), windy pendulum (middle row), and forced Duffing (bottom row). Columns 2 through 5 shows the error between the true Hamiltonian energy and learned Hamiltonian energy for both prior work and our methods. The prior work is shown in column 2 with PHNN \cite{desai2021port} and column 3 with SSGP \cite{tanaka2022ssgp}. Our method is shown in column 4, using equally weighted loss terms, and column 5, using GDA-balanced loss terms.}
    \label{fig:phase-map-comp-port}
\end{figure}

\begin{figure}[ht]
      \centering
    \includegraphics[width=0.6\textwidth]{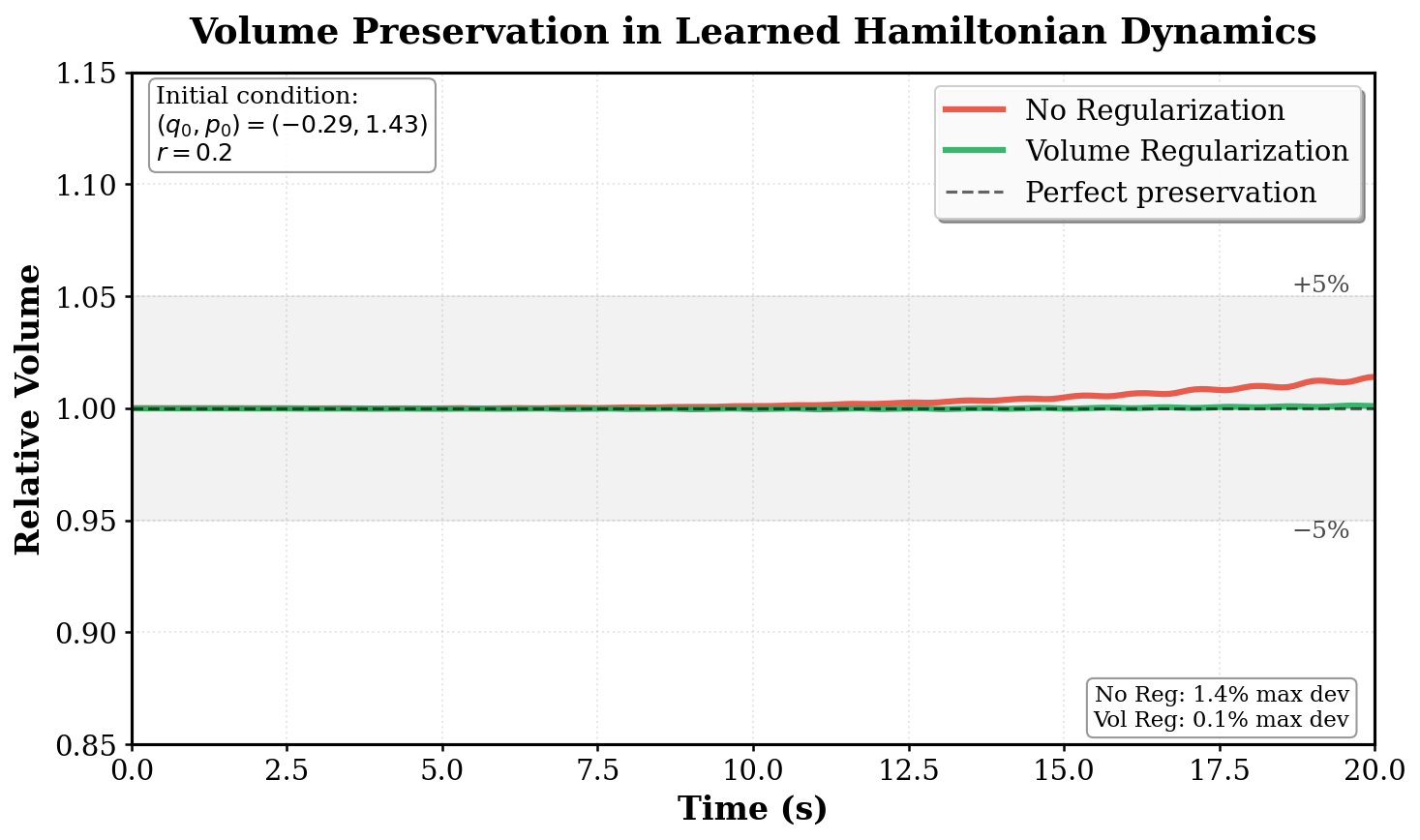}
      \caption{Volume preservation comparison between unregularized and volume-regularized models.
      Starting from initial condition $(q_0, p_0) = (2.0, -2.0)$ with a circular region of radius $r = 0.2$,
      we integrate the boundary forward in time using the learned dynamics with dissipation set to zero ($\eta = 0$).
      The unregularized model (red) exhibits significant volume contraction, while the volume-regularized model (green)
      maintains better preservation of phase space volume, demonstrating the effectiveness of the volume preservation
      regularizer $\mathcal{L}_{\text{vol}}$ in learning divergence-free dynamics. The dashed line indicates perfect
      volume preservation, and the shaded region represents $\pm 5\%$ deviation.}
      \label{fig:volume_preservation}
\end{figure}
\begin{figure}[ht]
    \centering
    \begin{subfigure}[t]{0.15\linewidth}
        \centering
        \includegraphics[width=\linewidth]{img/phase_plot_comps/damped_pendulum_true.pdf}
        \caption{True}
    \end{subfigure}
    \begin{subfigure}[t]{0.15\linewidth}
        \centering
        \includegraphics[width=\linewidth]{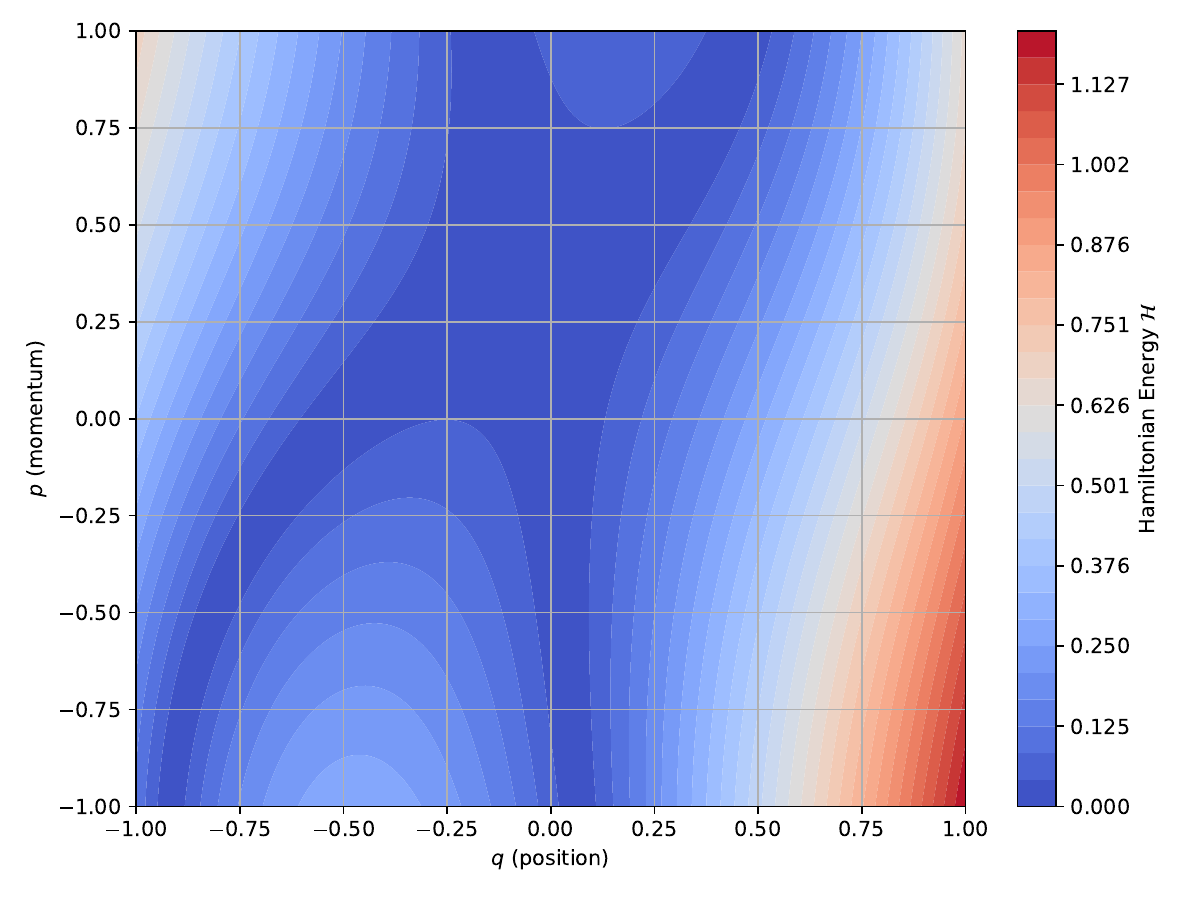}
        \caption{$\sigma=0$}
    \end{subfigure}
    \begin{subfigure}[t]{0.15\linewidth}
        \centering
        \includegraphics[width=\linewidth]{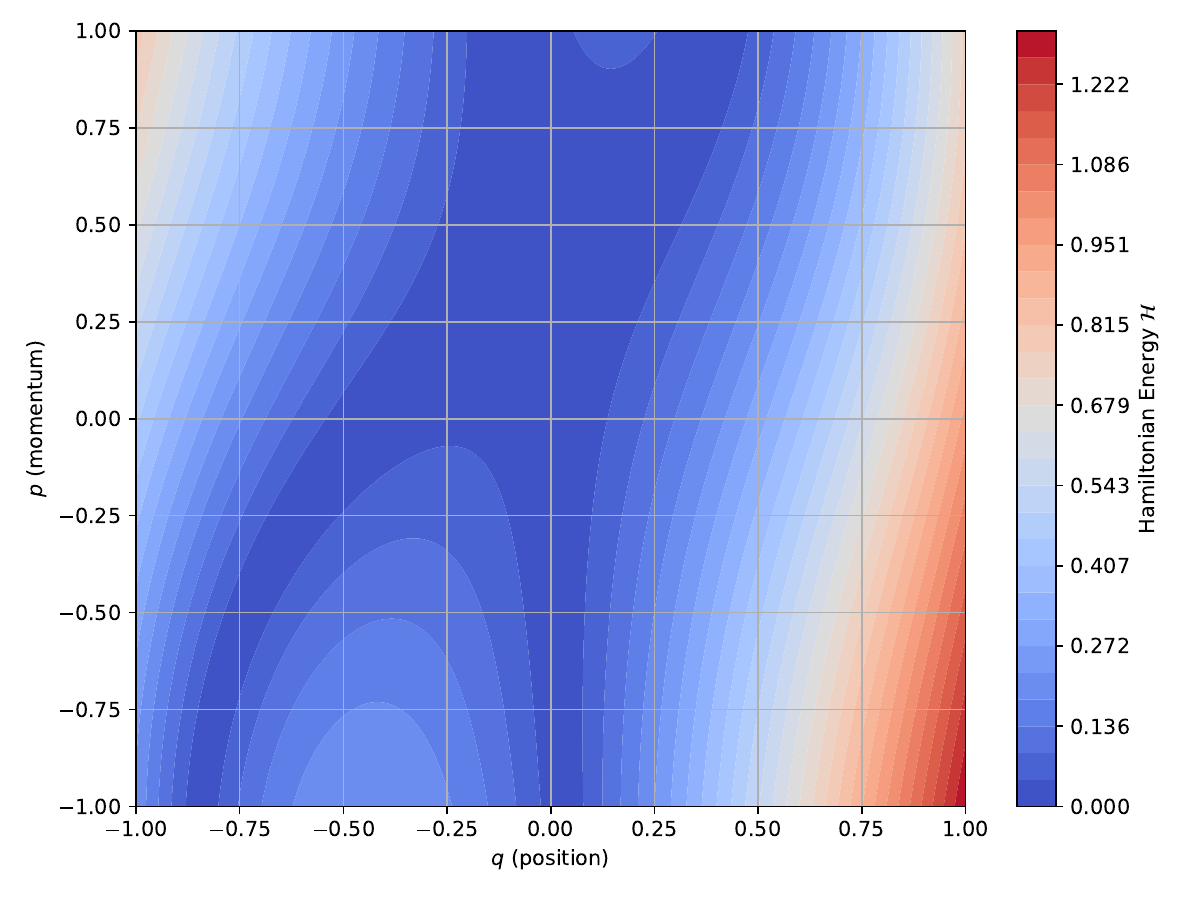}
        \caption{$\sigma=0.01$}
    \end{subfigure}
    \begin{subfigure}[t]{0.15\linewidth}
        \centering
        \includegraphics[width=\linewidth]{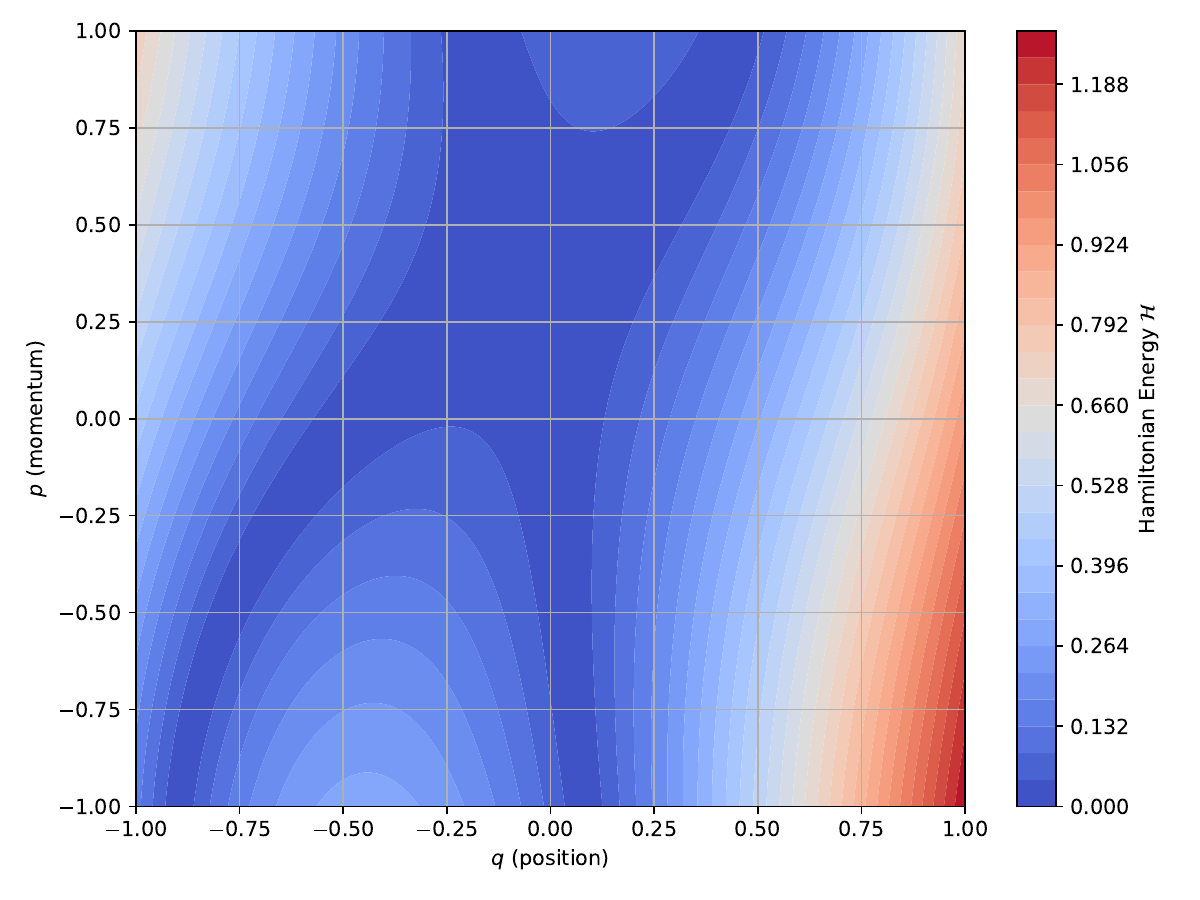}
        \caption{$\sigma=0.05$}
    \end{subfigure}
    \begin{subfigure}[t]{0.15\linewidth}
        \centering
        \includegraphics[width=\linewidth]{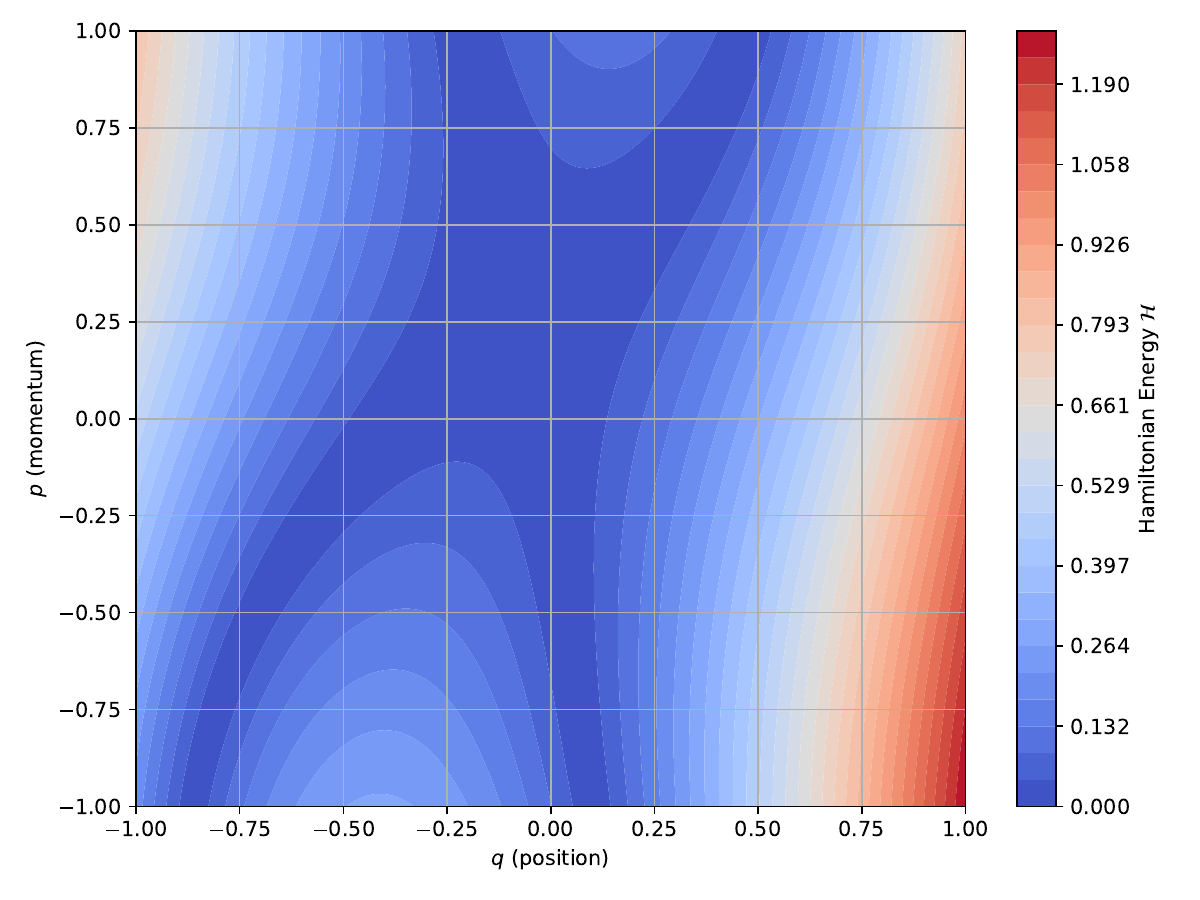}
        \caption{$\sigma=0.1$}
    \end{subfigure}
    \begin{subfigure}[t]{0.15\linewidth}
        \centering
        \includegraphics[width=\linewidth]{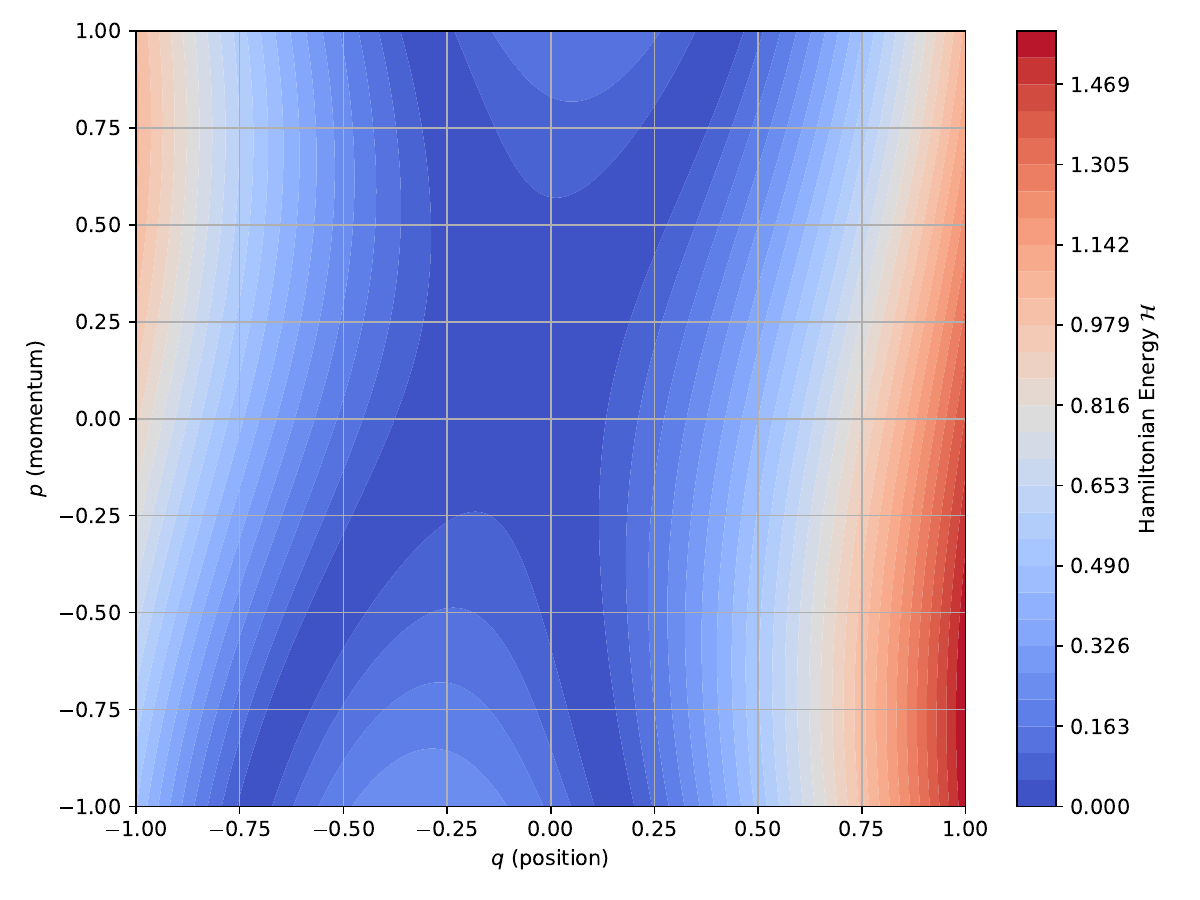}
        \caption{$\sigma=0.2$}
    \end{subfigure}
    \caption{Phase map comparison for the damped pendulum system with various noise levels. Subfigure (a) shows the ground truth Hamiltonian. Subfigure (b) show the error in learned (GDA) Hamiltonian from an uncorrupted dataset. Subfigures (c-f) show the error in the learned (GDA) Hamiltonian from datasets corrupted by Gaussian noise.}
    \label{fig:noisy-phase-maps}
\end{figure}

\subsection{Generalized Hamiltonian Dynamics' result}
First, we compare neural network methods (HNN \cite{greydanus2019hnn}, DHNN \cite{sosanya2022dhnn}, or P-HNN \cite{desai2021port}, depending on the class of dynamics) to SSGP \cite{tanaka2022ssgp} and to our method using regularized loss with conservation and stability corrections. Additionally, we also compare different methods of balancing the weights of the individual loss terms used in our method. We compare an equally weighted loss function to using GDA for updating $\lambda$ weights. We compare on various systems within the three classes of generalized Hamiltonian dynamics: conservative, dissipative, and Port-Hamiltonian. These systems were described in Section \ref{sec:background:examples} and the configurations are given in Table \ref{tab:data-config}. The results are listed in Table \ref{tab:loss-comp}, as well as visualized in Figure \ref{fig:phase-map-comp}. In Figure \ref{fig:phase-map-comp-port} we show additional comparisons of learned phase space maps for port-Hamiltonian systems. From Table \ref{tab:loss-comp}, and Figure \ref{fig:phase-map-comp}, it can be seen that our method can achieve improved performance across systems in each of the various classes of generalized Hamiltonian dynamics.

\paragraph{Validation of Conservative Property}
To demonstrate the effect of the conservation constraint regularizers, we display a volume calculation along time steps in Figure \ref{fig:volume_preservation}.
The unregularized SSGP model (red curve) does not preserve volume and $\nabla\cdot J\nabla\mathcal{H}$ term deviates starting from unseen integration times ($t\geq 10$), a clear indication that the learned vector field acquires a spurious non zero divergence. This behavior demonstrates that the regularizer effectively enforces a divergence‐free flow, preventing artificial volume loss and yielding a faithful symplectic approximation even without explicit dissipative terms.

\paragraph{Ablation Studies}
In Table \ref{tab:ablation}, we compare the use of different loss terms for single, damped, and windy pendulum systems. We compare the unregularized ELBO against applying each regularizer individually and combined. We compare both equal and GDA-balanced weighting of loss terms. In Tables \ref{tab:noisy-loss-comp}, \ref{tab:noisy-loss-comp-cons}, \ref{tab:noisy-loss-comp-port} and Figure \ref{fig:noisy-phase-maps}, we compare on datasets of various noise levels, ranging from noiseless to standard deviation $\sigma=0.2$. We compare for each of the three classes of Hamiltonian dynamics. In this study, we also include comparison with our model using a prior of the level of noise present in the data. 

\subsection{Choice of Hyperparameters}
For training each model, we use a learning rate of 1e-3 and 5000 epochs. Each epoch uses the full dataset. The SSGP \cite{tanaka2022ssgp} model uses 100 RFF basis functions. The HNN models use one hidden layer with 200 hidden nodes, with the HNN \cite{greydanus2019hnn} having 1 network $\mathcal H$, the DHNN \cite{sosanya2022dhnn} having 2 networks $\mathcal H, \mathcal D$, and the PHNN \cite{desai2021port} having 2 networks $\mathcal H, F$ and a dissipation matrix $D$. For port-Hamiltonian systems, the SSGP model is supplemented by an external forcing function $F(t)$, which is a neural network with one hidden layer of 100 hidden nodes. The configurations of the datasets are given in Table \ref{tab:data-config}. The configurations for the unforced and forced Duffing systems are based on \cite{desai2021port}. 

\section{Conclusion}
\label{sec:conclusion}
We propose a robust and generalizable learning framework for Hamiltonian systems, treating noisy observations as stochastic processes and calibrating trajectories via Gaussian processes. Stability is achieved through enforcing physically-informed constraints, and our numerical results highlight the model’s expressiveness and stability. Nonetheless, hard-constrained enforcement of conservation laws remains unresolved, and formal guarantees are an open question. Scaling to high-dimensional, real-world systems also remains a key challenge for future work.

\section*{Acknowledgment and Disclosure of Funding}
This research was supported in part by a grant from the Peter O’Donnell Foundation, the Michael J. Fox Foundation, Jim Holland-Backcountry Foundation to support AI in Parkinson, and in part from a grant from the Army Research Office accomplished under Cooperative Agreement Number
W911NF-19-2-0333.

\printbibliography

\end{document}